\documentclass[10pt,journal,compsoc]{IEEEtran}
\ifCLASSOPTIONcompsoc
  \usepackage[nocompress]{cite}
\else
  \usepackage{cite}
\fi
\ifCLASSINFOpdf
\else
\fi

\usepackage{amsmath,amsfonts}
\usepackage{amssymb}
\usepackage{mathtools}
\usepackage{amsthm}
\usepackage{array}
\usepackage[caption=false,font=small,labelfont=sf,textfont=sf]{subfig}
\usepackage{textcomp}
\usepackage{stfloats}
\usepackage{url}
\usepackage{verbatim}
\usepackage{graphicx}
\usepackage{cite}
\usepackage{hyperref}
\usepackage{cleveref}
\usepackage{booktabs}
\theoremstyle{plain}

\theoremstyle{definition}

\theoremstyle{remark}

\DeclarePairedDelimiterX{\SquareBrackets}[1]{[}{]}{#1}
\DeclarePairedDelimiterX{\RoundBrackets}[1]{(}{)}{#1}
\DeclarePairedDelimiterX{\DivergenceBrackets}[2]{[}{]}{#1\;\delimsize\|\;#2}

\NewDocumentCommand{\pr}{ O{p} r() }{
  \def\prArg{#2}\patchcmd{\prArg}{|}{\mid}{}{}#1\RoundBrackets{\prArg}}
\NewDocumentCommand{\p}{ r() }{\pr[p](#1)}
\NewDocumentCommand{\q}{ r() }{\pr[q](#1)}
\NewDocumentCommand{\prm}{ r() }{\pr[\mathrm{p}](#1)}
\NewDocumentCommand{\Normal}{ r() }{\pr[\operatorname{Normal}](#1)}
\NewDocumentCommand{\Cat}{ r() }{\pr[\operatorname{Cat}](#1)}
\NewDocumentCommand{\Beta}{ r() }{\pr[\operatorname{Beta}](#1)}
\NewDocumentCommand{\Bernoulli}{ r() }{\pr[\operatorname{Bernoulli}](#1)}
\NewDocumentCommand{\Dir}{ r() }{\pr[\operatorname{Dir}](#1)}

\crefname{algocf}{Algorithm}{Algorithms}

\usepackage[textsize=tiny]{todonotes}
\usepackage{threeparttablex}
\usepackage[linesnumbered,ruled,noend]{algorithm2e}
\usepackage{algpseudocode}
\usepackage{multirow}
\usepackage{colortbl}

\begin{document}
%
\title{Systematic Multi-Agent Vision-and-Language Navigation: Formulation, Benchmark,\\ and Method}
%
%
%
%

\author{Yunzhe Xu and Zhe Liu
\IEEEcompsocitemizethanks{\IEEEcompsocthanksitem The authors are with the School of Automation and Intelligent Sensing, Shanghai Jiao Tong University, Shanghai 200240, China (e-mail: xyz9911@sjtu.edu.cn; liuzhesjtu@sjtu.edu.cn).\protect\\
}
}

\IEEEtitleabstractindextext{%
\begin{abstract}
Vision-and-Language Navigation (VLN) has largely focused on a single agent following a single instruction, yet many real-world applications require teams of robots to tackle tasks beyond the capabilities of any individual agent. We present Systematic Multi-Agent Vision-and-Language Navigation, providing, to our knowledge, the first systematic formalization of multi-agent VLN as a constrained coordination problem: each mission consists of subtasks carrying dependency and resource constraints (presence locks and holding chains). A verified four-stage crafting pipeline instantiates the task as MAVLN, comprising 11{,}724 episodes across 145 scenes with teams of up to four agents under three instruction regimes, accompanied by tailored constraint-aware metrics. We further present TRISS, a coordination-ready navigation system coupling an LLM-based subtask scheduler, a shared topological memory that turns each agent's exploration into team knowledge, and a conflict-aware execution mechanism that realizes simultaneous intentions as collision-free routes. Extensive experiments establish TRISS as a comprehensive baseline and reveal substantial room for improvement across scheduling, planning, and execution, highlighting the challenges of coordinating under MAVLN task constraints. Project page: \href{https://xyz9911.github.io/mavln/}{https://xyz9911.github.io/mavln}.

\end{abstract}

}

\maketitle

\IEEEdisplaynontitleabstractindextext

%
\IEEEpeerreviewmaketitle

\begin{figure*}[t]
\centering
    \includegraphics[scale=0.57]{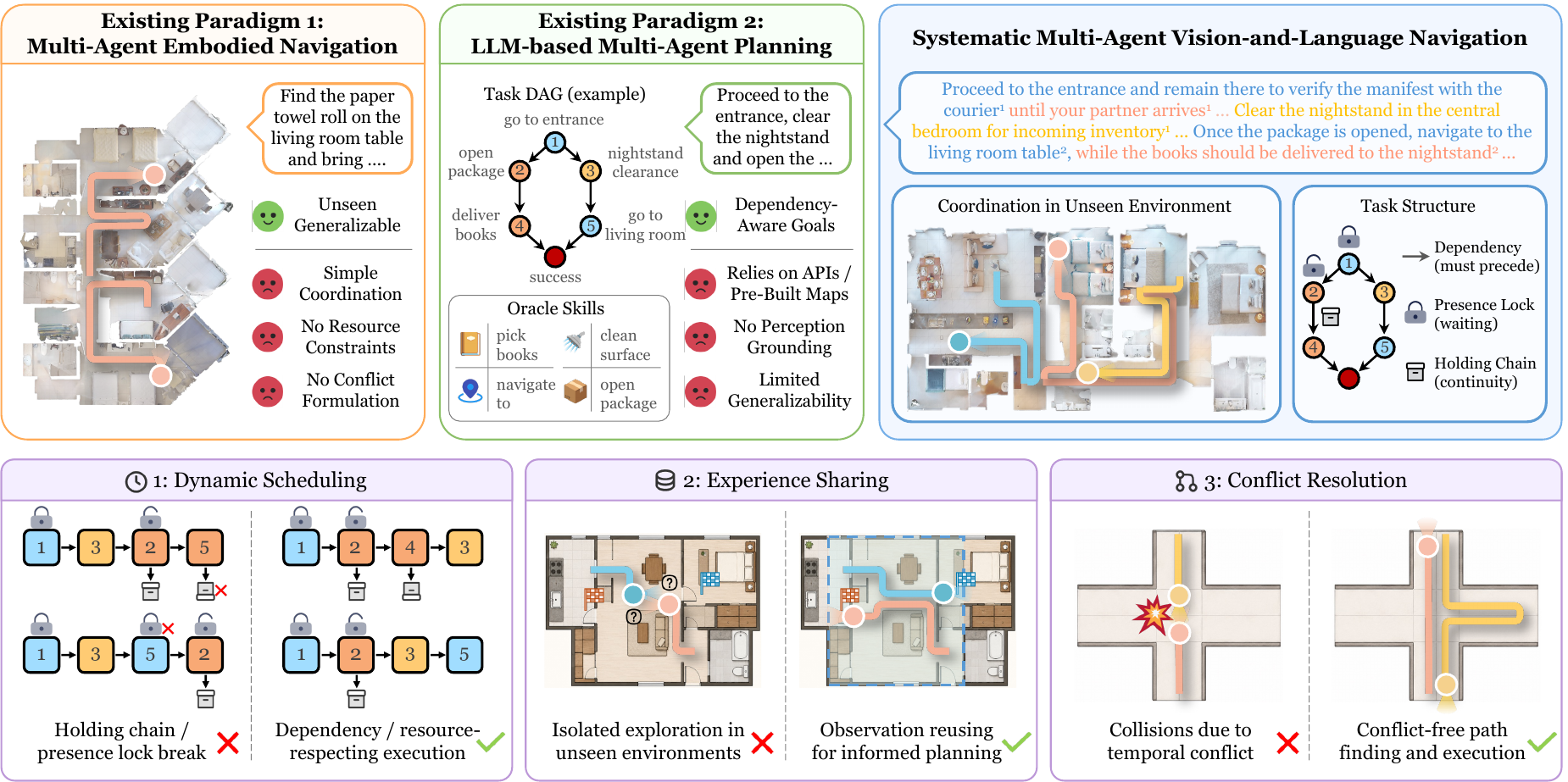}
    \caption{Overview of Systematic Multi-Agent Vision-and-Language Navigation. Existing settings either pursue simple goals with underused cross-agent experience and no measure of physical conflict, or plan rich task structures through atomic navigation APIs in known environments. Our task grounds a single mission into a team of VLN agents in an unseen environment. It requires dynamic scheduling under dependency and resource constraints, experience sharing to help teammates plan in unexplored spaces, and conflict resolution to turn simultaneous intentions into collision-free routes.}
    \label{fig:intro}
\end{figure*}

\section{Introduction}

\IEEEPARstart{V}{ision}-and-Language Navigation \cite{2018r2r}, a cornerstone challenge in Embodied AI, has witnessed rapid progress \cite{2022duet, 2023scalevln, 2024navid}. Under the original VLN setting, an agent follows an instruction and navigates within a pre-defined topological graph to reach a nearby goal. Though the paradigm's simplicity eases the integration of vision-and-language models, it limits practical relevance. Navigation in continuous environments \cite{2020vlnce} was therefore proposed to narrow this gap, and has been further extended to physically realistic settings \cite{2025vlnpe, 2025vlnverse}. Moreover, long-horizon navigation \cite{2025lhvln} and memory-persistent navigation \cite{2023ivln, 2026memoir} have been explored to carry out complex tasks comprising successive subtasks and to improve over time. Extending VLN to multi-agent scenarios is a natural next step toward general intelligence in the real world, as team collaboration reaches beyond single-agent capacity.

For embodied multi-agent navigation, early attempts include collaborative exploration \cite{2022maans}, scene understanding \cite{2020multiagentiqa}, manipulation \cite{2020furnmove}, and single-target \cite{2023conavgpt} or multi-target navigation \cite{2021collavn}. Importantly, CoNavBench \cite{2026conav} presents a major step toward multi-agent VLN, treating two agents as a coordinated team to exploit parallelism and save wall-clock time under relay-based coordination, and DeCoNav \cite{2026deconav} further propels this paradigm with dynamic task allocation and re-planning for real-time, adaptive coordination. Meanwhile, Co-VLN \cite{2026covln} reveals that peer observation sharing can significantly boost navigation performance, demonstrating the benefit of memory-wise coordination. In task-planning domains, LLMs are widely used to allocate subtasks among multiple agents \cite{2024smartllm} or to facilitate inter-agent communication for informed decisions \cite{2024coela}, while relying on oracle simulator functions \cite{2025emos} or pre-explored semantics \cite{2025roboosnext} for perception and low-level control.

Despite these advances, existing paradigms exhibit two critical limitations. First, current embodied multi-agent navigation tasks impose over-simplified constraints: the required coordination rarely goes beyond chain-style dependencies, limited environmental coverage curtails the experience reuse that benefits generalization, and inter-agent conflict is left unmeasured, undermining real-world feasibility. Second, planning-oriented systems are confined to known environments: although their task dependencies are commonly represented as directed acyclic graphs (DAGs), the reliance on pre-built reconstructions or simulator APIs sidesteps the challenge of generalizing to unseen environments. The two lines have thus matured on complementary halves of the same real-world problem, while their intersection remains unexplored.

To this end, we propose Systematic Multi-Agent Vision-and-Language Navigation, an embodied navigation task requiring sophisticated coordination in unseen environments. As shown in \Cref{fig:intro}, a team of agents must first identify the dependency and resource constraints implied by the instruction (e.g., the magenta agent's first subtask can only complete after the blue agent's counterpart, while the blue agent must hold its position until the magenta agent's completion releases it), and then execute the subtasks separately to fulfill the mission, benefiting from shared experiences and minimizing physical conflicts. We further propose a verified task crafting pipeline for multi-agent VLN, comprising grounded scene understanding for multi-granularity scene annotation, diverse mission synthesis driven by user intents and zone features, mission instantiation and schedule refinement for optimal task efficiency, and verified instruction rendering and filtering for fidelity control and reliable generation. The pipeline yields \textbf{MAVLN}, a Systematic Multi-Agent VLN benchmark featuring DAG-structured subtask dependencies, resource constraints and conflict awareness. The benchmark comprises 11{,}724 episodes with flexible team sizes and three instruction regimes, and is accompanied by tailored constraint-aware evaluation metrics that measure coordination effectiveness.

We identify three key challenges of the proposed task: dynamic subtask scheduling under dependency and resource constraints, simultaneous experience sharing, and inter-agent conflict resolution. To tackle these challenges, we present \textbf{T}opological Memo\textbf{r}y-based Collaborat\textbf{i}ve Navigation and \textbf{S}ubtask \textbf{S}cheduling (TRISS), a coordination-ready navigation system that pairs a compact LLM with an efficient planner model: 1) an LLM-based subtask scheduler that identifies task constraints and iteratively dispatches atomic instructions to downstream navigators; 2) a topological memory-based navigation planner that facilitates navigation planning with shared topological memory; and 3) a conflict-aware execution mechanism that couples optimal target allocation with conflict-free path finding while maintaining topological consistency.

The experiments validate TRISS's effectiveness and establish it as a comprehensive baseline. Nevertheless, substantial room for improvement remains across scheduling, planning, and execution, outlining concrete directions for future work. Our key contributions include:

\begin{itemize}
\item \textbf{Systematic Formulation of Multi-Agent VLN:} To the best of our knowledge, we provide the first systematic formalization of multi-agent VLN as a dependency- and resource-constrained coordination problem.

\item \textbf{Verified Multi-Agent Task Crafting:} We present an automated generation pipeline that yields reliable multi-agent tasks through grounded perception, mission synthesis, schedule refinement, and round-trip verification.

\item \textbf{Comprehensive Benchmark and Platform:} We release MAVLN, comprising 11{,}724 episodes under three instruction regimes across 145 scenes, together with a synchronized multi-agent VLN platform built on Habitat~3 and tailored constraint-aware evaluation metrics.

\item \textbf{Coordination-Ready Navigation System:} We present TRISS, a system coupling a compact LLM and an efficient planner through scheduling, shared memory, and conflict-aware execution. Experiments demonstrate its effectiveness while highlighting substantial headroom.
\end{itemize}
\section{Related Work}

\subsection{Vision-and-Language Navigation}
Traditional Vision-and-Language Navigation (VLN) \cite{2018r2r, 2020reverie, 2020vlnce} requires an agent to navigate to a destination based on an instruction, and methods have evolved through improvements including memory mechanisms \cite{2021hamt, 2022duet, 2024etpnav}, navigation planning \cite{2025sali, 2026memoir} and data augmentation \cite{2023scalevln, 2025navrag}. The focus has shifted to LLM integration, training large-scale navigation foundation models \cite{2024navid, 2025navfom} as VLN planners \cite{2024navila} to boost generalization capabilities in indoor \cite{2025janusvln, 2025streamvln} and outdoor environments \cite{2025flame}. LLMs' zero-shot capabilities \cite{2024navgpt, 2025cavln, 2025smartway} enable collaboration between LLMs and expert VLN models \cite{2023marchinchat, 2024llmcopliot}. However, traditional VLN cannot represent real-world complexity in addressing human needs, and this limitation has motivated multi-target navigation \cite{2020multion, 2026neuro}, long-horizon navigation \cite{2023ivln, 2026memoir, 2025lhvln} and physically realistic navigation \cite{2025vlnpe, 2025vlnverse}. Multi-agent VLN has also been explored, ranging from early attempts at multi-expert debates for decisions \cite{2024discuss} and collaborative aerial search \cite{2025aeroduo, 2026conavuav} to memory-wise coordination \cite{2026covln} and relay-style collaboration \cite{2026conav, 2026deconav}. In these tasks, agents either pursue independent goals or cooperate on loosely structured team goals. Moreover, the collaboration benefit is often confined to either makespan optimization \cite{2026conav} or shared memory \cite{2026covln}, leaving challenges including conflict resolution, experience reuse and scheduling either unaddressed or studied in isolation.

\subsection{Multi-Agent Embodied Navigation}
Embodied collaborative navigation has shown promising results in various scenarios including exploration \cite{2022maans}, scene understanding \cite{2020multiagentiqa}, instance localization \cite{2021collavn} and manipulation \cite{2020furnmove}. Agents commonly collaborate to find multiple targets \cite{2021collavn} or a shared single target \cite{2023conavgpt}. In particular, leveraging LLMs in collaborative navigation is becoming increasingly common, with protocols ranging from centralized coordination \cite{2023conavgpt} and communication-based coordination \cite{2024camon} to memory-based decentralized coordination \cite{2025mcoconav}. However, these tasks all focus on simple task goals, limiting their applicability to real-world scenarios. Task decomposition has been explored \cite{2022embodiedmultiagent}, but the dependencies are generally predefined and restricted to a fixed team size. Additionally, collision avoidance \cite{2015cbs, 2019primal} is seldom considered in embodied scenarios.

\subsection{LLM-based Multi-Agent Planning}
SMART-LLM \cite{2024smartllm} pioneered the application of LLMs to multi-agent task planning for robotic tasks. These planning styles range from leveraging linear programming \cite{2024lip}, PDDL \cite{2025lammap} and prompt optimization \cite{2026kppo} to enhancing system capacity in task allocation and dependency identification \cite{2026hierarchicalprompt}. CoELA \cite{2024coela} explores decentralized protocols in LLM-based agent coordination, and ProAgent \cite{2024proagent} further enables cooperation without explicit communication. EMOS \cite{2025emos} extends predefined agent roles to physical compositions by proposing a heterogeneous multi-agent collaboration system. Importantly, LLaMAR \cite{2024llamar} extends LLM-based multi-agent systems (MAS) to partially observable environments, but these MAS rely on either oracle perception to capture environment layouts or oracle skills to carry out low-level actions through simulator APIs, leaving the embodiment addressed only at the surface level. RoboOS-NeXT \cite{2025roboosnext} integrates planning with low-level execution, yet it relies heavily on spatial memory initialization, forfeiting its generalization to unseen environments.

\begin{figure*}[t]
    \centering
    \includegraphics[width=\textwidth]{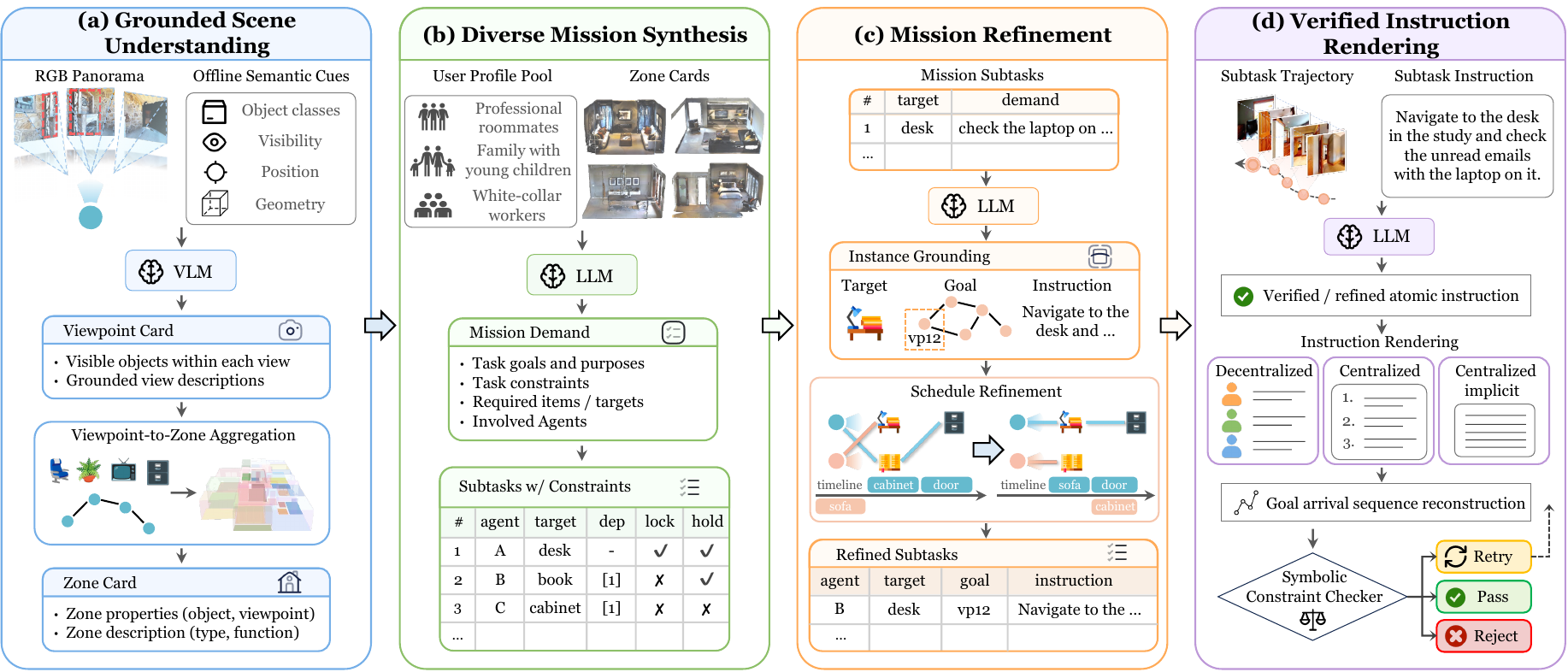}
    \caption{The crafting pipeline of MAVLN. \textbf{(a)} Descriptions are generated at viewpoint granularity and summarized into zone-level descriptions. \textbf{(b)} Mission scenarios are synthesized from a household profile and zone summaries, emitting constraint-annotated subtasks. \textbf{(c)} Subtask targets are grounded to viewpoints to generate atomic instructions, and the mission makespan is optimized. \textbf{(d)} A VLM verifies and refines each subtask instruction; consolidated instructions are rendered in three different styles. Crafted instructions pass a closed-loop filter comparing LLM-reconstructed arrival sequences against task constraints.}
    \label{fig:task}
\end{figure*}

\section{Task Formulation \& Benchmark}

\subsection{Systematic Multi-Agent VLN}
Vision-and-Language Navigation in Continuous Environments (VLN-CE) \cite{2020vlnce} requires an agent to follow an instruction $\ell$ and navigate using a set of low-level actions (move forward 0.25\,m, turn left or right 15 degrees, or stop). At each timestep $t$, the agent perceives a panoramic observation $\mathcal{O}_t = \{(o_i^{rgb}, o_i^{dep})\}_{i=1}^{12}$ comprising RGB and depth images, captured from 12 equally spaced horizontal heading directions.

Inheriting the perception and action spaces of VLN-CE, we formalize multi-agent VLN as follows. An episode places a team of agents $\mathcal{A}=\{a_1,\dots,a_M\}$ at distinct start poses and provides instructions $\mathcal{L}$ in one of three regimes (\Cref{sec:stats}). The task is a DAG over subtasks $\mathcal{T}=\{\tau_1,\dots,\tau_N\}$, where each subtask $\tau_k=(g_k, A_k, D_k, h_k, l_k, U_k)$ is visible only to the evaluator: $g_k$ is the goal grounded to one object instance, $A_k\subseteq\mathcal{A}$ the agents permitted to execute it, $D_k\subseteq\mathcal{T}$ its prerequisite subtasks, and $(h_k,l_k,U_k)$ its resource attributes. If $h_k$ holds, the executing agent carries an object after completing $\tau_k$ and may complete no subtask other than the consumer of $\tau_k$, by the same agent, forming a \textit{holding chain}; if $l_k$ holds, the agent acquires a \textit{presence lock}: the agent must maintain presence at $g_k$ until a teammate completes some $\tau_j$ with $\tau_k\in U_j$. Dependency and resource constraints are thus orthogonal: the former govern when subtasks may complete, the latter govern what an agent may do with itself while they are live. A subtask succeeds when a permitted agent declares arrival within the success radius of $g_k$ with all of $D_k$ completed and no holding or locking restriction violated; the mission succeeds if every subtask succeeds.

\subsection{Verified Task Crafting}
We leverage the Gemini-3 series as the backbone LLM across four curation stages (\Cref{fig:task}).

\noindent\textbf{Grounded Scene Understanding.} Our platform is built on Habitat 3 \cite{2024habitat} with the HM3D dataset \cite{2021hm3d}. We first render observations at each viewpoint based on the navigation graphs \cite{2023scalevln}, and gather surrounding objects through the semantic sensor. We then prompt a VLM with RGB panorama splits and object cues to obtain descriptions containing verified details of observable objects. Based on optimized zone partitions of HM3D scenes, we prompt an LLM with the aggregated viewpoint descriptions of each zone to obtain a general zone description along with a list of the objects available in it.

\noindent\textbf{Diverse Mission Synthesis.} We first generate household profiles for diverse scenarios using an LLM. For each annotated zone, we extract its description and properties together with those of its neighbors, and prompt an LLM to pick the most suitable profile. Conditioned on the household characteristics and zone information, the LLM brainstorms candidate mission scenarios, each comprising a task demand and a list of subtasks with targets and constraints. Synthesized missions violating a set of well-formedness principles are filtered out.

\noindent\textbf{Mission Refinement.} Based on the synthesized task information, we further prompt an LLM with the available viewpoints containing the targets of each subtask to obtain the destination viewpoint together with a fine-grained atomic instruction per subtask. The instantiated task then undergoes a makespan optimization phase that searches over constraint-consistent execution orders and agent allocations, mutating the drafted allocation to optimize the schedule. A subsequent filtering step eliminates consecutive subtasks that share the same viewpoint or lie too close to each other.

\noindent\textbf{Verified Instruction Rendering.} We prompt a VLM with each subtask's atomic instruction and observations rendered along the agent's trajectory from start locations to subtask destinations to verify and refine the instruction. An LLM then consolidates the refined atomic instructions into task instructions of decentralized, centralized and centralized-implicit styles. Finally, an LLM reconstructs an ordered arrival sequence based on the instructions, which is checked against the ground-truth task constraints to certify fidelity; failed cases are given a limited number of retries before being filtered out. We further assess instruction reliability through human evaluation.

\subsection{Dataset Statistics}
\label{sec:stats}
\Cref{tab:benchmarks} compares MAVLN, our proposed Systematic Multi-Agent VLN benchmark, with existing collaborative embodied benchmarks. MAVLN is the first multi-agent VLN benchmark supporting flexible team sizes, and the first embodied navigation benchmark to introduce complex dependency and resource constraints as well as inter-agent conflict measurement. It contains 11{,}724 collaborative episodes spanning 145 scenes with an average path length of 35.9 meters, providing sufficient area coverage in unseen environments. The episodes derive from 3{,}908 unique missions, each rendered under the three instruction regimes below; missions are partitioned scene-disjointly into 3{,}079 training (128 scenes), 386 validation-unseen (7 scenes), and 443 test-unseen (10 scenes) missions.

The episodes are divided into three splits by instruction regime, each with val-unseen and test-unseen scenarios. The decentralized split distributes an explicit instruction to each agent, and every subtask permits only its designated agent as the executor. The centralized split provides one consolidated instruction specifying the whole team's mission; unlike the decentralized split, every subtask permits any agent as an allowed executor. The centralized-implicit split further removes agent attribution from the centralized instruction, and the team must additionally infer who performs what.

\begin{figure*}[t]
  \centering
  \subfloat[Subtasks per episode]{\includegraphics[width=0.24\textwidth]{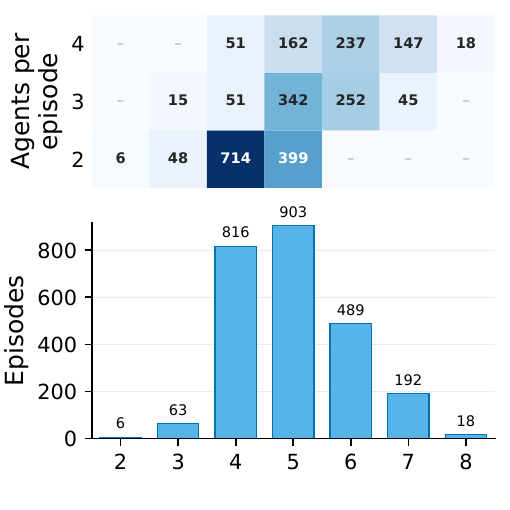}\label{fig:stats_task_scale}}\hfil
  \subfloat[Constraint decomposition]{\includegraphics[width=0.24\textwidth]{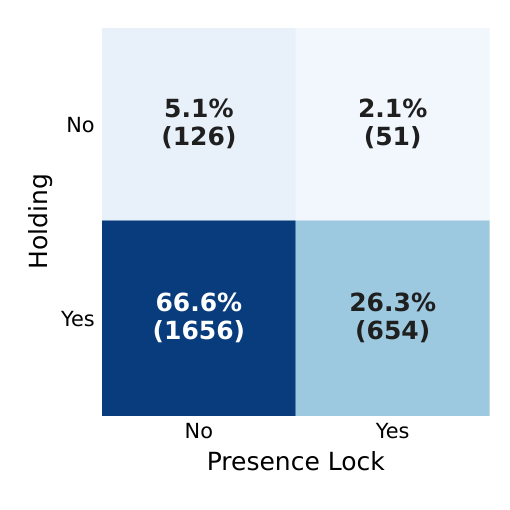}\label{fig:constraint_decomposition}}\hfil
  \subfloat[Trajectory length]{\includegraphics[width=0.24\textwidth]{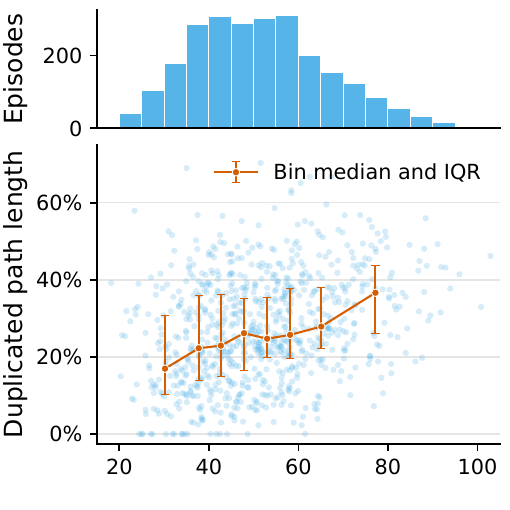}\label{fig:trajectory_overlap}}\hfil
  \subfloat[Multi-agent makespan]{\includegraphics[width=0.24\textwidth]{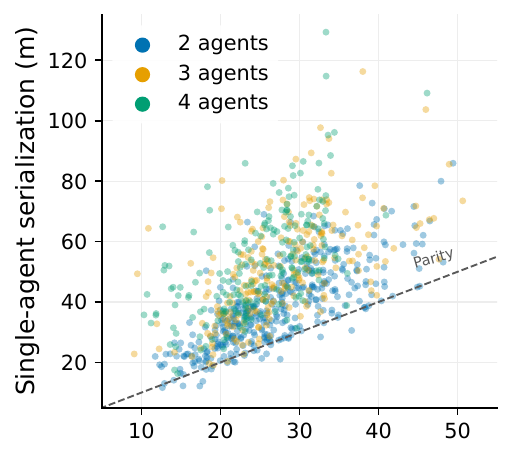}\label{fig:makespan_scatter}}
  \caption{Statistics of 2,487 evaluation episodes. \textbf{(a)} Joint distribution of team size (rows) and subtask count (columns), with the marginal over subtask counts below. \textbf{(b)} Decomposition of resource constraints including holding chains and presence locks. \textbf{(c)} Team trajectory length (top) against the fraction of it duplicated by more than one subtask (bottom). \textbf{(d)} Optimal multi-agent makespan against the best single-agent serialization of the same mission.}
  \label{fig:overview_stats}
\end{figure*}

As illustrated in \Cref{fig:overview_stats}, episodes carry five subtasks on average and up to eight, over half of them field three or more agents, and subtask counts grow with team size. Dependency edges are universal by construction, while holding chains appear in $92.9\%$ of missions and presence locks in $28.3\%$, with $26.3\%$ exercising both primitives at once and only $5.1\%$ reducing to a pure DAG. Across successive subtasks, a team spends approximately a quarter of its total travel distance revisiting previously traversed paths, with this proportion increasing for longer trajectories. Such retracing quantifies the headroom available to persistent and shared memory. The optimal multi-agent makespan is shorter than the minimum single-agent completion time in all but several resource-constrained cases. This gap widens with team size, demonstrating that larger teams enable greater parallelism.

\subsection{Evaluation Metrics}
We extend standard VLN evaluation to the multi-agent, constraint-aware setting. Arrivals are credited within a $3$\,m radius of any navigable anchor of the target instance.
\textbf{Success Rate (SR)} is the fraction of missions in which every subtask succeeds under the task's constraint semantics.
\textbf{Success weighted by Path Length (SPL)} weights SR by the ratio between the length of the optimal path and total traversed path of all agents.
\textbf{Conditional Subtask Success Rate (CSR)} is the fraction of subtasks whose arrival lies within the success radius, honoring all dependency, locking, and holding constraints.
\textbf{Independent / Conditional Subtask SPL (ISPL / CSPL)} measure per-subtask path efficiency, gated by success without and with task constraints, respectively. Both normalize by the length the agent traverses within the segment; ISPL takes as reference the shortest path from the terminal position of its preceding subtask to the current goal, while CSPL takes the subtask's fragment of the optimal reference path.
\textbf{Task Completion (TC)} is the fraction of subtasks the team consumes irrespective of whether the arrival is correct.
\textbf{Time Steps (TS)} counts synchronized simulation steps until termination. Unlike planned makespan \cite{2026conav}, this realized makespan captures congestion, waiting, and deadlock during execution.
\textbf{Multi-Agent Conflict (MAC)} is the number of time steps in which any two agents lie within twice the radius of each other, divided by the product of TS and team size.
Arrival-to-subtask matching can be ambiguous: multiple target instances may lie within the success radius, and an incorrect stop may fall near another target. We pair the embeddings of scheduler's emitted atomic instructions (\Cref{sec:scheduler}) with the ground-truth ones one-to-one through all-MiniLM-L6-v2 to map from atomic instructions to subtask IDs. This requires exposing a pool of atomic instructions; matching for systems that consume the mission instruction undecomposed is left to future work.

\begin{table*}[t]
    \centering
    \caption{Comparison with related benchmarks on collaborative embodied tasks.}
    \small
    \renewcommand{\arraystretch}{1.05}
    \setlength{\aboverulesep}{0pt}
    \setlength{\belowrulesep}{0pt}
    \setlength{\tabcolsep}{2pt}
    \begin{tabular}{lccccccccc}
        \toprule
        \textbf{Benchmark} & \textbf{Max Agents} & \textbf{Task} & \textbf{Perception} & \textbf{Dependency} & \textbf{Resource} & \textbf{Conflict} & \textbf{Path Length} & \textbf{Scenes} & \textbf{Episodes} \\
        \midrule
        R2R~\cite{2018r2r} & 1 & VLN & \checkmark & - & - & - & 9.4 & 90 & 21,567 \\
        REVERIE~\cite{2020reverie} & 1 & VLN & \checkmark & - & - & - & 9.7 & 90 & 21,702 \\
        IR2R~\cite{2023ivln} & 1 & VLN & \checkmark & Chain & - & - & 9.4 & 90 & 21,567 \\
        LHPR-VLN~\cite{2025lhvln} & 1 & VLN & \checkmark & Chain & - & - & 25.2 & 216 & 3,260 \\
        CollaVN~\cite{2021collavn} & 4 & Img.-Goal Nav. & \checkmark & - & - & $\times$ & - & 572 & 1,180K \\
        PARTNR~\cite{2025partnr} & 2 & Planning & $\times$ & DAG & $\times$ & $\times$ & - & 60 & 100K \\
        Habitat-MAS~\cite{2025emos} & 4 & Planning & $\times$ & DAG & $\times$ & $\times$ & - & 61 & 1,113 \\
        CoNavBench~\cite{2026conav} & 2 & VLN & \checkmark & Chain & $\times$ & $\times$ & 19.4 & 128 & 4,048 \\
        \midrule
        \textbf{MAVLN (Ours)} & \textbf{4} & \textbf{VLN (w/ Planning)} & \textbf{\checkmark} & \textbf{DAG} & \textbf{\checkmark} & \textbf{\checkmark} & 35.9 & 145 & 11,724 \\
        \bottomrule
    \end{tabular}
    \label{tab:benchmarks}
\end{table*}

\begin{figure*}[t]
\centering
    \includegraphics[scale=0.55]{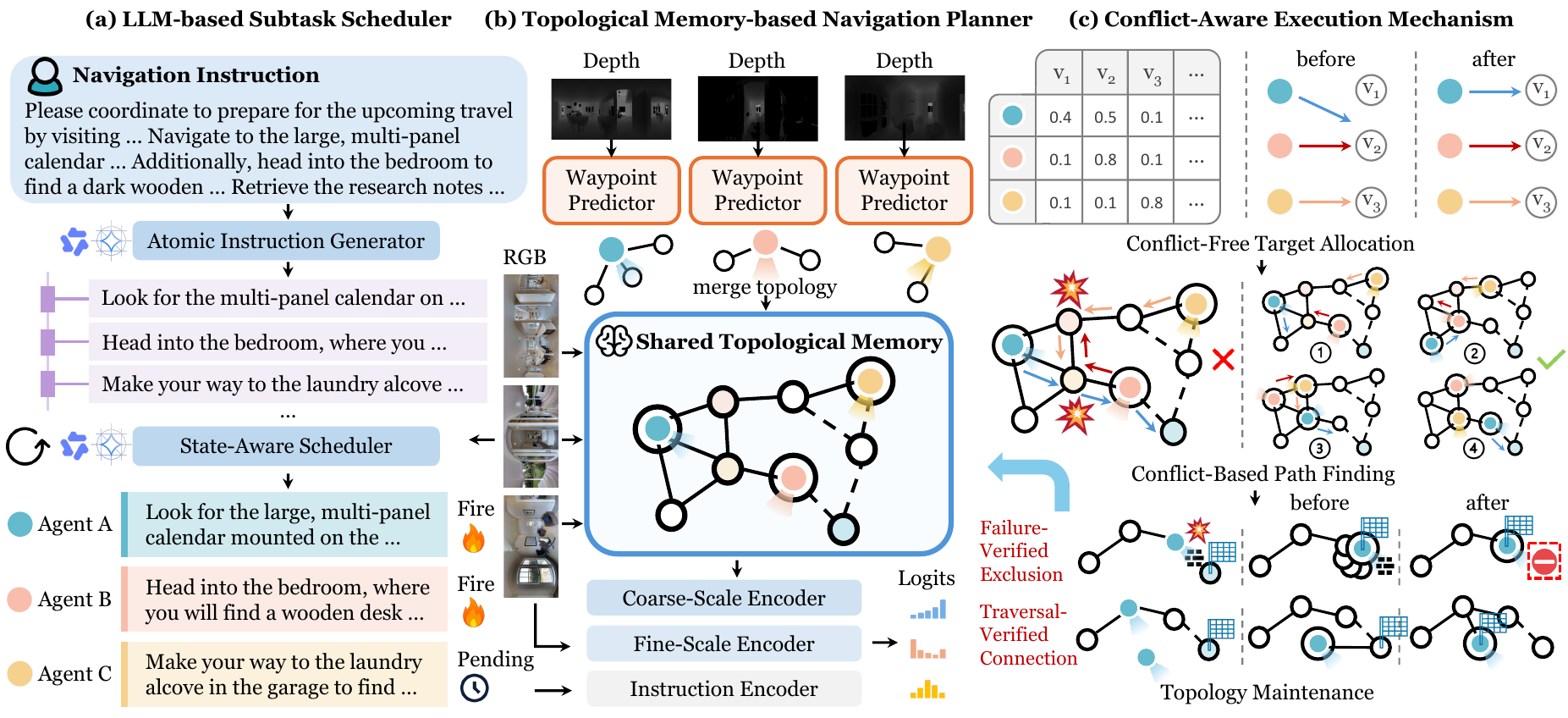}
    \caption{Details of Topological Memory-based Collaborative Navigation and Subtask Scheduling. \textbf{(a)} Atomic instructions are extracted and allocated by an atomic instruction scheduler. \textbf{(b)} A navigation planner predicts viewpoint-level actions over the shared topology for each agent. \textbf{(c)} We deploy conflict-free target allocation and plan agent paths to avoid collisions. A low-level controller outputs actions while the overall topology is protected.}
    \label{fig:method}
\end{figure*}

\section{Method}
This section presents TRISS for the MAVLN benchmark. As shown in \Cref{fig:method}, our method comprises three modules: a subtask scheduler that dynamically dispatches subtasks, a shared memory-based navigation planner for viewpoint selection, and a conflict-aware mechanism that executes paths while maintaining topological consistency.

\subsection{LLM-based Subtask Scheduler}
\label{sec:scheduler}
To facilitate subtask scheduling, the task instructions $\mathcal{L}$ are first processed by an LLM-based atomic instruction extractor, which identifies subtasks and yields a shared pool of atomic navigation instructions $\mathcal{I}=\{\ell_i\}_{i=1}^L$, where $L$ denotes the number of subtasks identified by the extractor. An LLM-based atomic instruction scheduler then allocates unspent atomic instructions in successive scheduling rounds, reasoning about the dependency and resource constraints implied by the raw task instructions $\mathcal{L}$. A round is triggered once every previously dispatched navigation has concluded, at which point the scheduler emits an assignment $I_n\in(\mathcal{I}\cup\{\varnothing\})^M$ granting each agent at most one atomic instruction. A blank assignment is reserved for agents that are locked, blocked or finished, and the mission is completed once $I_n$ is blank for every agent.

In practice, although one agent's next subtask may depend on another's, the agent can travel toward its destination in advance as long as no presence lock pins it elsewhere, because arrival alone does not complete a subtask. A stop action with no unmet dependency fires automatically upon arrival, whereas a gated action waits for an explicit firing signal. Such pre-allocation overlaps travel with waiting, shortening the realized makespan and directly improving TS. We therefore maintain a pending list $P_n\subseteq\mathcal{A}$ of agents whose newly allocated subtask will be waiting to be fired, and an ordered firing list $F_n$ of previously pending agents whose actions fire at the beginning of this round. The scheduler jointly emits:

\begin{equation}
I_n, P_n, F_n = \mathrm{LLM}_{sched}\bigl(\mathcal{L}, \mathcal{I}, \{I_m, P_m, F_m\}_{m<n}\bigr).
\label{eq:scheduler}
\end{equation}

\subsection{Topological Memory-based Navigation Planner}
To facilitate experience sharing to support informed decisions, we first construct a shared topological memory, consisting of a topological representation of the environment and visual features accumulated through viewpoint prediction and merging. On top of this memory, a DUET-style \cite{2022duet} cross-modal planner combines each agent's atomic instruction and observation to produce navigation decisions.

\noindent\textbf{Shared Topological Memory.} To facilitate cooperative reasoning, all agents jointly construct one topological map on the fly. Similar to prior works \cite{2024etpnav}, the graph $\mathcal{G}_t=(\mathcal{V}_t,\mathcal{E}_t)$ represents the constructed topology after $t$ navigation steps, with nodes divided into explored viewpoints $\mathcal{V}^{e}_t$, containing all nodes previously visited by any agent, and unexplored viewpoints $\mathcal{V}^{u}_t$, proposed by viewpoint prediction but never visited. A depth-only waypoint predictor \cite{2022bridging} consumes the depth view representations $d_t=\{d_t^{(i)}\}_{i=1}^{12}$ and proposes $K$ neighboring viewpoints $Y_t=\{y_i\}_{i=1}^{K}$ around the agent. Every position the team produces is localized against the shared node set through:

\begin{equation}
    v^{*}(q) = \underset{v \in \mathcal{V}_t}{\arg\min}\, \operatorname{dist}(v, q).
\label{eq:localize}
\end{equation}

\noindent A viewpoint $y_i \in Y_t$ is matched if $\operatorname{dist}(v^{*}(y_i), y_i) \leq \varepsilon_{\text{loc}}$, yielding the matched viewpoints $\mathcal{V}_t'$ and the unmatched remainder $Y_t'$; letting $v_t$ denote the agent's current viewpoint, the map is updated in one pass as $\mathcal{V}_{t+1} = \mathcal{V}_t \cup Y_t'$ and $\mathcal{E}_{t+1} = \mathcal{E}_t \cup \{(v_t, v) \mid v \in \mathcal{V}_t' \cup Y_t',\ v \neq v_t\}$. Since all agents apply the same operator to the same node set, duplicate proposals from different teammates merge rather than fork. Unlike single-agent topological mapping, which assumes each agent occupies its intended viewpoint before the next planning step, we include an additional localization phase before viewpoint prediction: the agent's measured position $q_t$ is localized through \Cref{eq:localize}, and its current viewpoint is set to $v_t = v^{*}(q_t)$ if the match lies within $\varepsilon_{\text{loc}}$. This ensures that, when backtracking fails, the intended viewpoint is not mistakenly treated as the agent’s new viewpoint or updated using observations collected from the agent’s actual position.

To obtain the viewpoint representations, we feed the view representations $r_t$, $d_t$ and the orientation encodings captured at viewpoint $v_t$ into a panorama encoder to obtain the contextual panoramic embeddings $\hat{r}_t=\{\hat{r}_t^{(i)}\}_{i=1}^{12}$. The representation of the current viewpoint $v_t$ is set to the average pooling of $\hat{r}_t$. A newly instantiated viewpoint $v \in Y_t'$ adopts $\hat{r}_t^{(i)}$ of the view through which it was proposed, while a matched unexplored viewpoint $v \in \mathcal{V}_t' \cap \mathcal{V}^{u}_t$ appends $\hat{r}_t^{(i)}$ to its accumulated embeddings and updates its representation to their mean.

\noindent \textbf{Cross-Modal Planning.}
Each agent receives an atomic instruction $\ell \in I_n$ from the subtask scheduler and encodes it into textual representations $\hat{\ell}$. Following DUET \cite{2022duet}, we employ a coarse-scale encoder for node-level reasoning and a fine-scale encoder for neighborhood-level reasoning. To bound the planning space, for agent $a$, we denote by $\mathcal{V}_t^{(a)}\subseteq \mathcal{V}_t^e$ the nodes it has visited in the current round and construct the candidate set $\hat{\mathcal{V}}_t^{(a)}=\mathcal{V}_t^e\cup\mathcal{N}\!\left(\mathcal{V}_t^{(a)}\right)$ containing all explored nodes and the neighboring frontiers of its current-round trajectory. This restriction prevents candidate explosion in continuous environments, particularly as the team size grows. The coarse-scale encoder processes candidate viewpoint features $X = [x_0, x_1, \ldots, x_{|\hat{\mathcal{V}}_t^{(a)}|}]$ together with $\hat{\ell}$ through a graph transformer, where $x_0$ is the stop token. Global action logits $s_i^{(c)}$ are computed from $\hat{x}_i$ for each viewpoint, where $\hat{x}_i \in \hat{X}$ denotes the output of the coarse-scale encoder, providing node-level navigation preferences. Comparatively, fine-scale planning processes the immediate contextual panoramic embeddings $\hat{r}_t$. We prepend a stop token $r_0$ and feed the instruction embedding $\hat{\ell}$ and the sequence $R = [r_0; \hat{r}_t]$ into the fine-scale encoder, yielding local action logits $s_i^{(f)}$ for the current neighbors $v_i \in Y_t' \cup \mathcal{V}_t'$, which are converted to the global action space:
\begin{equation}
    s_i^{(f')} = \begin{cases}
    s_i^{(f)}, & \text{if } v_i \in Y_t' \cup \mathcal{V}_t', \\
    s_{\text{back}}, & \text{otherwise},
    \end{cases}
\label{eq:convert}
\end{equation}
\noindent where $s_{\text{back}}$ aggregates the logits of the visited neighbors of $v_t$. To support dual-scale navigation planning, we dynamically fuse the two granularities using a scalar gate $\sigma_t$ inferred through the concatenated features $[\hat{r}_0;\hat{x}_0]$. The final logit for each candidate is computed as $s_i = \sigma_t\, s_i^{(f')} + (1-\sigma_t)\, s_i^{(c)}$.

\subsection{Conflict-Aware Execution Mechanism}
To facilitate conflict-free execution, we first adopt a conflict-free target allocation algorithm to prevent multiple agents from selecting the same viewpoint as their destination. Subsequently, we leverage a conflict-based path finding algorithm to plan conflict-free path sequences toward the targets, carried out by low-level controllers. Finally, we introduce a topology maintenance mechanism to avoid topology corruption.

\noindent \textbf{Conflict-Free Target Allocation.} Multi-agent navigation lets several agents select target viewpoints on a shared graph, introducing potential conflicts where agents jockey for the same viewpoint, leading to collisions. We therefore model multi-agent viewpoint allocation as a maximum-score bipartite assignment problem, which is solved by the Hungarian algorithm. Every agent is thus dispatched to a distinct viewpoint, maximizing the team's total policy score; an agent whose preferred frontier is claimed by a higher-scoring teammate is rerouted to its best remaining alternative within the same solve.

\noindent \textbf{Conflict-Based Path Finding.} Agents traverse multiple viewpoints before arriving at their destinations, which regularly gives rise to temporal conflicts when several of them occupy the same viewpoint along their paths. We leverage Conflict-Based Search \cite{2015cbs} to resolve them over the shared topology. A space-time A* first plans each agent a short-horizon trajectory across explored nodes, with unexplored nodes enterable only as goals and waiting admitted as a move, requiring goal occupancy only at the end of the horizon so that an agent may pause or temporarily vacate its goal to let a teammate pass. A bounded search then resolves the remaining conflicts on the synchronized timeline, returning the least-conflict route set once its expansion budget is exhausted. The resulting routes are executed jointly through a low-level controller \cite{2024etpnav}.

\noindent \textbf{Topology Maintenance.} When executing low-level actions along the searched $H$-step path, agents commonly encounter two types of failures: mapping failures and drifting issues, both of which corrupt a topology that every teammate subsequently consumes. We propose failure-verified exclusion to tackle mapping failures. A predicted viewpoint $v_H \in \mathcal{V}_t^u$ might be unreachable, so that navigation terminates without any position change while the waypoint predictor keeps proposing the same location, forming a mapping failure. Therefore we propose a collective ban list $B=\{(u, y)\}$, pairing the viewpoint $u$ from which the failed attempt departed with the position $y$ of the proposal. Whenever a navigation from $u$ fails to produce a position change, we remove $y$ from the memory and insert $(u, y)$ into $B$. Subsequent proposals are then filtered as

\begin{equation}
Y_t \leftarrow \{y_i \in Y_t \mid
\operatorname{dist}(y,y_i) \geq \varepsilon_{\text{ban}}, \forall (v_t,y)\in B\},
\label{eq:fve}
\end{equation}

\noindent where $\varepsilon_{\text{ban}}$ indicates the radius of the banned area. Moreover, we propose traversal-verified connection to tackle drifting issues. During the localization phase, a naive approach connects the viewpoint $v_t$ with the penultimate viewpoint $v_{H-1}$ of the path. However, the controller might drift away from the explored viewpoint or exit early upon collision, so that $v_{H-1}$ is not where the agent physically came from and the added edge would assert a traversal that never happened. Instead, we identify the last viewpoint of the searched path that the agent verifiably passed through without early exit, reducing the risk of spurious shortcuts caused by detours or recovery.

\subsection{Student-Forced Multi-Agent Learning}
The navigation planner is trained by imitation learning with the ground-truth atomic navigation instructions. To avoid dependency-induced waiting and multi-agent simulation overhead during training, we augment the single-agent VLN pipeline with shared topological memory. We group the training set so that episodes assigned to one team take pairwise distinct starts, and preferentially exhibit path overlap between teammates' reference trajectories to reward reading teammate experience. We use student-forcing to encourage exploration during training and compute the loss at each step through:

\begin{equation}
\label{eq:loss}
\small
\begin{aligned}
\mathcal{L}_{\mathrm{IL}} &= -\sum_{a=1}^{M} \sum_{t} \log p\bigl(v^{\star}_{a,t} \mid \ell_a, \mathcal{G}_t, \mathcal{O}_{a,t}\bigr),
\end{aligned}
\end{equation}
\normalsize
\noindent where $v^{\star}_{a,t}$ denotes the viewpoint closest to the agent’s goal.

\begin{table*}[t]
\renewcommand{\arraystretch}{1.1}
\setlength{\aboverulesep}{0pt}
\setlength{\belowrulesep}{0pt}
\setlength{\tabcolsep}{1.6pt}
\centering
\caption{Performance on MAVLN under oracle and LLM-based subtask schedulers across the three instruction regimes}
\resizebox{\textwidth}{!}{
{
\begin{threeparttable}
\begin{tabular}{lc cccccccc c cccccccc}
\toprule
&& \multicolumn{8}{c}{\textbf{Val-Unseen}} && \multicolumn{8}{c}{\textbf{Test-Unseen}} \\
\cmidrule(lr){3-10} \cmidrule(lr){12-19}
\textbf{Sched.} & \textbf{Nav.} & \textbf{\texttt{SR}} $\uparrow$ & \textbf{\texttt{SPL}} $\uparrow$ & \textbf{\texttt{CSR}}$\uparrow$ & \textbf{\texttt{CSPL}}$\uparrow$ & \textbf{\texttt{ISPL}}$\uparrow$ & \textbf{\texttt{TC}}$\uparrow$ & \textbf{\texttt{TS}}$\downarrow$ & \textbf{\texttt{MAC}}$\downarrow$ && \textbf{\texttt{SR}}$\uparrow$ & \textbf{\texttt{SPL}}$\uparrow$ & \textbf{\texttt{CSR}}$\uparrow$ & \textbf{\texttt{CSPL}}$\uparrow$ & \textbf{\texttt{ISPL}}$\uparrow$ & \textbf{\texttt{TC}}$\uparrow$ & \textbf{\texttt{TS}}$\downarrow$ & \textbf{\texttt{MAC}}$\downarrow$ \\
\midrule
\midrule

Ora. (SA) & Ora.
& 67.1 & 63.1 & 74.9 & 68.5 & 91.6 & 100.0 & 412 & 0.00
&& 69.1 & 64.4 & 78.0 & 70.8 & 90.4 & 100.0 & 419 & 0.00 \\

Ora. (LE) & Ora.
& 100.0 & 99.7 & 100.0 & 99.5 & 92.6 & 100.0 & 261 & 1.16
&& 100.0 & 99.6 & 100.0 & 99.4 & 92.0 & 100.0 & 248 & 1.39 \\

Ora. & Ora.
& 100.0 & 99.8 & 100.0 & 99.6 & 92.6 & 100.0 & 215 & 1.75
&& 100.0 & 99.6 & 100.0 & 99.4 & 92.0 & 100.0 & 206 & 2.10 \\

Ora. & ETP
& 2.8 & 1.9 & 27.2 & 21.6 & 30.5 & 100.0 & 394 & 0.87
&& 5.6 & 4.2 & 28.7 & 22.4 & 30.6 & 100.0 & 403 & 0.85 \\

Ora. (SA) & Ours
& 5.4 & 3.3 & 26.0 & 15.3 & 25.2 & 100.0 & 2056 & 0.00
&& 5.9 & 3.8 & 25.2 & 17.4 & 27.7 & 100.0 & 1665 & 0.00 \\

Ora. & Ours
& 9.1 & 4.0 & 36.4 & 19.7 & 26.9 & 100.0 & 1588 & 0.82
&& 9.3 & 5.2 & 33.3 & 20.0 & 27.9 & 100.0 & 1494 & 1.42 \\

Rand. & Ours
& 0.3 & 0.1 & 7.9 & 4.3 & 23.6 & 100.0 & 1621 & 0.91
&& 0.2 & 0.1 & 6.7 & 4.1 & 24.1 & 100.0 & 1547 & 1.12 \\

Seq. & Ours
& 3.4 & 1.6 & 23.6 & 12.8 & 24.9 & 100.0 & 1593 & 0.90
&& 2.5 & 1.2 & 21.8 & 12.6 & 26.5 & 100.0 & 1443 & 1.28 \\

\midrule
\multicolumn{19}{c}{\textbf{Decentralized Instruction Style}}\\

Gemma4 & Ours
& 7.0\scriptsize{$\pm$0.9} & 2.9\scriptsize{$\pm$0.4} & 33.6\scriptsize{$\pm$0.9} & \textbf{18.2}\scriptsize{$\pm$0.6}
& 24.0\scriptsize{$\pm$0.4} & 94.6\scriptsize{$\pm$0.5} & \textbf{1570}\scriptsize{$\pm$37} & 0.95\scriptsize{$\pm$0.1}
&& 8.3\scriptsize{$\pm$0.1} & \textbf{4.4}\scriptsize{$\pm$0.0} & \textbf{33.2}\scriptsize{$\pm$0.3} & \textbf{19.5}\scriptsize{$\pm$0.3}
& \textbf{26.5}\scriptsize{$\pm$0.2} & 95.8\scriptsize{$\pm$0.2} & \textbf{1447}\scriptsize{$\pm$13} & 1.22\scriptsize{$\pm$0.0} \\

Qwen3.6 & Ours
& 7.3\scriptsize{$\pm$0.7} & 2.8\scriptsize{$\pm$0.3} & \textbf{34.7}\scriptsize{$\pm$0.4} & \textbf{18.2}\scriptsize{$\pm$0.1}
& \textbf{24.9}\scriptsize{$\pm$0.3} & \textbf{97.0}\scriptsize{$\pm$0.3} & 1613\scriptsize{$\pm$59} & 0.87\scriptsize{$\pm$0.1}
&& 7.9\scriptsize{$\pm$0.8} & 4.1\scriptsize{$\pm$0.4} & 32.1\scriptsize{$\pm$0.3} & 18.7\scriptsize{$\pm$0.2}
& 25.9\scriptsize{$\pm$0.4} & \textbf{97.4}\scriptsize{$\pm$0.2} & 1521\scriptsize{$\pm$25} & 1.26\scriptsize{$\pm$0.1} \\

\midrule
\multicolumn{19}{c}{\textbf{Centralized Instruction Style}}\\

Gemma4 & Ours
& 7.5\scriptsize{$\pm$1.2} & \textbf{3.1}\scriptsize{$\pm$0.5} & 34.3\scriptsize{$\pm$1.0} & 18.1\scriptsize{$\pm$0.5}
& 23.9\scriptsize{$\pm$0.5} & 94.4\scriptsize{$\pm$0.5} & \textbf{1570}\scriptsize{$\pm$40} & 0.83\scriptsize{$\pm$0.0}
&& 6.8\scriptsize{$\pm$0.1} & 3.9\scriptsize{$\pm$0.1} & 31.6\scriptsize{$\pm$0.1} & 18.6\scriptsize{$\pm$0.1}
& 25.4\scriptsize{$\pm$0.1} & 95.0\scriptsize{$\pm$0.2} & 1475\scriptsize{$\pm$10} & 1.13\scriptsize{$\pm$0.0} \\

Qwen3.6 & Ours
& \textbf{7.6}\scriptsize{$\pm$0.1} & \textbf{3.1}\scriptsize{$\pm$0.2} & 34.4\scriptsize{$\pm$0.3} & \textbf{18.2}\scriptsize{$\pm$0.2}
& 24.0\scriptsize{$\pm$0.4} & 95.7\scriptsize{$\pm$0.1} & 1619\scriptsize{$\pm$29} & 0.85\scriptsize{$\pm$0.0}
&& \textbf{8.4}\scriptsize{$\pm$0.9} & 4.2\scriptsize{$\pm$0.5} & 32.4\scriptsize{$\pm$0.8} & 19.1\scriptsize{$\pm$0.7}
& 26.3\scriptsize{$\pm$0.5} & 96.5\scriptsize{$\pm$0.4} & 1470\scriptsize{$\pm$7} & 1.03\scriptsize{$\pm$0.0} \\

\midrule
\multicolumn{19}{c}{\textbf{Centralized Implicit Instruction Style}}\\

Gemma4 & Ours
& 5.5\scriptsize{$\pm$0.1} & 2.3\scriptsize{$\pm$0.1} & 32.1\scriptsize{$\pm$0.8} & 16.5\scriptsize{$\pm$0.2}
& 22.5\scriptsize{$\pm$0.2} & 92.5\scriptsize{$\pm$0.4} & 1611\scriptsize{$\pm$44} & \textbf{0.76}\scriptsize{$\pm$0.0}
&& 7.1\scriptsize{$\pm$0.3} & 3.7\scriptsize{$\pm$0.2} & 29.6\scriptsize{$\pm$0.6} & 17.0\scriptsize{$\pm$0.5}
& 23.3\scriptsize{$\pm$0.3} & 92.8\scriptsize{$\pm$0.4} & 1512\scriptsize{$\pm$37} & \textbf{0.97}\scriptsize{$\pm$0.1} \\

Qwen3.6 & Ours
& 6.2\scriptsize{$\pm$0.6} & 2.7\scriptsize{$\pm$0.3} & 31.5\scriptsize{$\pm$0.2} & 16.6\scriptsize{$\pm$0.0}
& 23.2\scriptsize{$\pm$0.2} & 96.8\scriptsize{$\pm$0.2} & 1662\scriptsize{$\pm$16} & 0.78\scriptsize{$\pm$0.0}
&& 6.8\scriptsize{$\pm$0.7} & 3.5\scriptsize{$\pm$0.4} & 28.8\scriptsize{$\pm$0.6} & 16.5\scriptsize{$\pm$0.5}
& 24.1\scriptsize{$\pm$0.5} & 96.5\scriptsize{$\pm$0.5} & 1562\scriptsize{$\pm$26} & 1.03\scriptsize{$\pm$0.1} \\

\bottomrule
\end{tabular}
\begin{tablenotes}
    \footnotesize
    \item Sched.: scheduler; Nav.: navigator; Ora.: oracle; (SA): single agent variant; (LE): legacy variant without pre-allocation; ETP: ETPNav trained on MAVLN; Rand./Seq.: ground-truth atomic instructions dispatched to free agents in random / sequential order. Bold: best among LLM schedulers.
\end{tablenotes}
\end{threeparttable}}}
\label{tab:main}
\end{table*}

\section{Experiments}
\subsection{Experimental Setup}
\noindent \textbf{Training Settings.}
We adopt a pretraining stage for the planner following ETPNav \cite{2024etpnav}. We construct pretraining data from the NavRAG \cite{2025navrag} training set together with ground-truth pairs of atomic instructions and subtask trajectories extracted from the MAVLN training split. Since NavRAG is also built on HM3D, we remove all trajectories belonging to the unseen splits of MAVLN beforehand to prevent data leakage. We pretrain for 200{,}000 steps and select the checkpoint with the best navigation performance on val-unseen. For imitation learning, we set the maximum trajectory length to 30, let four agents share the topological memory, and train for 30{,}000 steps, selecting the checkpoint that maximizes the sum of SR and SPL. Training runs on a single A100 GPU.

\noindent \textbf{Inference Settings.}
Given the size and complexity of the evaluation splits, we adopt two open-source VLMs to keep evaluation tractable: Qwen3.6-27b, a lightweight yet capable model on agentic tasks, and Gemma4-31b, the latest open-source release from Google. Both serve as subtask schedulers within our framework, and all LLM-dependent results are averaged over three independent runs. Unless stated otherwise, ablation studies replace the LLM scheduler with an oracle scheduler that emits ground-truth atomic instructions, which isolates navigation quality from scheduling variance and improves time efficiency. Inference runs on a single A100 GPU.

\begin{table}[t]
\renewcommand{\arraystretch}{1.1}
\setlength{\aboverulesep}{0pt}
\setlength{\belowrulesep}{0pt}
\setlength{\tabcolsep}{1.4pt}
\centering
\caption{Ablation of system components and training recipes}
\resizebox{0.95\linewidth}{!}{
\begin{tabular}{lc cccc cc}
\toprule
&& \multicolumn{6}{c}{\textbf{Val-Unseen}} \\
\cmidrule(lr){3-8}
\textbf{Component}
& \textbf{Variant}
& \textbf{\texttt{SR}}$\uparrow$
& \textbf{\texttt{SPL}}$\uparrow$
& \textbf{\texttt{CSR}}$\uparrow$
& \textbf{\texttt{CSPL}}$\uparrow$
& \textbf{\texttt{TS}}$\downarrow$
& \textbf{\texttt{MAC}}$\downarrow$ \\
\midrule

\multicolumn{8}{c}{\textbf{Framework}}\\

\multirow{2}{*}{Shared Memory}
& Disabled
& 6.2
& 3.0
& 33.4
& 19.5
& \textbf{1065}
& \textbf{0.7} \\

& Enabled
& \textbf{9.1}
& \textbf{4.0}
& \textbf{36.4}
& \textbf{19.7}
& 1588
& 0.8 \\

\midrule

\multirow{2}{*}{Planning Space}
& Unbounded
& 7.8
& 3.7
& 35.0
& \textbf{20.2}
& \textbf{1316}
& 0.9 \\

& Bounded
& \textbf{9.1}
& \textbf{4.0}
& \textbf{36.4}
& 19.7
& 1588
& \textbf{0.8} \\

\midrule

\multirow{2}{*}{Planner}

& Coarse-Scale
& 0.0 & 0.0 & 8.5 & 1.2 & 2468 & 1.9 \\

& Dual-Scale
& \textbf{9.1}
& \textbf{4.0}
& \textbf{36.4}
& \textbf{19.7}
& \textbf{1588}
& \textbf{0.8} \\

\midrule

\multirow{2}{*}{Step Limit}
& 20
& 6.0
& 2.6
& 34.7
& \textbf{20.0}
& \textbf{1272}
& 0.9 \\

& 30
& \textbf{9.1}
& \textbf{4.0}
& \textbf{36.4}
& 19.7
& 1588
& \textbf{0.8} \\

\midrule

\multicolumn{8}{c}{\textbf{Training}}\\

\multirow{2}{*}{Teacher-Forcing}
& 50\%
& 6.0
& 2.8
& 33.1
& \textbf{21.6}
& \textbf{1010}
& 1.1 \\

& 0\%
& \textbf{9.1}
& \textbf{4.0}
& \textbf{36.4}
& 19.7
& 1588
& \textbf{0.8} \\

\midrule

\multirow{2}{*}{Concurrent Agents}
& 2
& 4.9
& 2.5
& 32.3
& 18.2
& 1589
& 1.1 \\

& 4
& \textbf{9.1}
& \textbf{4.0}
& \textbf{36.4}
& \textbf{19.7}
& \textbf{1588}
& \textbf{0.8} \\

\midrule

\multirow{2}{*}{Accum. Rounds}
& 1
& \textbf{9.1}
& \textbf{4.0}
& 36.4
& \textbf{19.7}
& \textbf{1588}
& \textbf{0.8} \\

& 2
& 8.5
& 3.4
& \textbf{37.3}
& 19.4
& 1760
& 0.9 \\

\midrule

\multirow{3}{*}{Pretrain Source}
& MAVLN
& 5.7
& 2.4
& 30.0
& 18.1
& \textbf{1404}
& 1.0 \\

& +REVERIE
& 7.3
& 3.4
& 31.7
& 18.0
& 1667
& 1.1 \\

& +NavRAG
& \textbf{9.1}
& \textbf{4.0}
& \textbf{36.4}
& \textbf{19.7}
& 1588
& \textbf{0.8} \\

\bottomrule
\end{tabular}
}
\label{tab:ablation}
\end{table}

\subsection{Quantitative Analysis}
\Cref{tab:main} compares navigation performance across schedulers and instruction regimes. With oracle scheduling and navigation, TRISS achieves 100.0 SR with 215 and 206 time steps on val-unseen and test-unseen, respectively. Removing pre-allocation increases TS to 261 and 248, demonstrating the benefit of overlapping travel with dependency-induced waiting. Restricting execution to one agent reduces SR to 67.1 and 69.1 and increases TS to 412 and 419, highlighting the feasibility and parallelism benefits of multi-agent execution.

Replacing oracle navigation with our navigator reduces test-unseen SPL from 99.6 to 5.2. The gap is not specific to our navigator: ETPNav trained on MAVLN reaches only 2.8 and 5.6 SR, although it attains higher CSPL and ISPL by reaching fewer subtasks along more direct paths. Multi-agent execution still outperforms its single-agent counterpart (3.8 SPL). Replacing the oracle scheduler with an LLM further reduces SPL to 4.4 (Gemma4, decentralized), yet it clearly outperforms random and sequential dispatch, whose ISPL stays close to the oracle scheduler's while CSR collapses, confirming that failures stem from violated constraints rather than navigation. Centralized-implicit instructions are hardest, as allocation must be inferred on top of constraint reasoning (3.5 SPL for Qwen3.6). Lower MAC does not necessarily accompany better task performance, and Qwen3.6 outperforms Gemma4 on SR in four of the six scenarios.

\begin{table}[t]
\renewcommand{\arraystretch}{1.1}
\setlength{\aboverulesep}{0pt}
\setlength{\belowrulesep}{0pt}
\setlength{\tabcolsep}{3pt}
\centering
\caption{Ablation of the conflict-resolving mechanisms}
\resizebox{0.9\linewidth}{!}{
\begin{threeparttable}
\begin{tabular}{cc cccccc c}
\toprule
&& \multicolumn{7}{c}{\textbf{Val-Unseen}} \\
\cmidrule(lr){3-9}

\textbf{CFTA} & \textbf{CBPF}
& \textbf{\texttt{SR}}$\uparrow$
& \textbf{\texttt{SPL}}$\uparrow$
& \textbf{\texttt{CSR}}$\uparrow$
& \textbf{\texttt{CSPL}}$\uparrow$
& \textbf{\texttt{ISPL}}$\uparrow$
& \textbf{\texttt{TS}}$\downarrow$
& \textbf{\texttt{MAC}}$\downarrow$ \\
\midrule

&
& 8.8
& \textbf{4.1}
& 34.9
& 19.7
& 27.2
& \textbf{1230}
& 2.2 \\

\checkmark &
& 8.3
& 3.8
& 35.2
& \textbf{19.8}
& \textbf{27.5}
& 1265
& 1.4 \\

\checkmark & \checkmark
& \textbf{9.1}
& 4.0
& \textbf{36.4}
& 19.7
& 26.9
& 1588
& \textbf{0.8} \\

\bottomrule
\end{tabular}
\begin{tablenotes}
    \footnotesize
    \item CFTA: Conflict-Free Target Allocation; CBPF: Conflict-Based Path Finding.
\end{tablenotes}
\end{threeparttable}
}
\label{tab:conflict}
\end{table}

\begin{table}[t]
\renewcommand{\arraystretch}{1.1}
\setlength{\aboverulesep}{0pt}
\setlength{\belowrulesep}{0pt}
\setlength{\tabcolsep}{3.0pt}
\centering
\caption{Ablation of topology mapping and maintenance}
\resizebox{0.95\linewidth}{!}{
\begin{threeparttable}
\begin{tabular}{ccc ccccc c}
\toprule
&&& \multicolumn{6}{c}{\textbf{Val-Unseen}} \\
\cmidrule(lr){4-9}

\textbf{Localization} & \textbf{FVE} & \textbf{TVC}
& \textbf{\texttt{SR}}$\uparrow$
& \textbf{\texttt{SPL}}$\uparrow$
& \textbf{\texttt{CSR}}$\uparrow$
& \textbf{\texttt{CSPL}}$\uparrow$
& \textbf{\texttt{ISPL}}$\uparrow$
& \textbf{\texttt{MAC}}$\downarrow$ \\
\midrule

& &
& 6.5
& 2.7
& 34.0
& 18.3
& 25.4
& 1.1 \\

\checkmark & \checkmark &
& 7.8
& 3.5
& 36.0
& 19.6
& 26.1
& \textbf{0.8} \\

\checkmark & & \checkmark
& 7.3
& 3.0
& 34.1
& 18.6
& 25.1
& 1.0 \\

\checkmark & \checkmark & \checkmark
& \textbf{9.1}
& \textbf{4.0}
& \textbf{36.4}
& \textbf{19.7}
& \textbf{26.9}
& \textbf{0.8} \\

\bottomrule
\end{tabular}
\begin{tablenotes}
    \footnotesize
    \item FVE: Failure-Verified Exclusion; TVC: Traversal-Verified Connection.
\end{tablenotes}
\end{threeparttable}
}
\label{tab:localization}
\end{table}

\subsection{Ablation Studies \& Analyses}
\noindent\textbf{Framework Components.}
\Cref{tab:ablation} ablates the components of TRISS. Disabling the shared topological memory reduces SR from 9.1 to 6.2 and SPL from 4.0 to 3.0, validating memory sharing as the central mechanism of the framework. The same mechanism raises TS from 1065 to 1588 because conflict-resolving takes time. Bounding the planning space during cross-modal reasoning raises SR from 7.8 to 9.1 by filtering noisy distant candidates, while the unbounded variant retains marginally better CSPL from its wider horizon. Dual-scale planning has the largest observed effect: the coarse-scale variant achieves 0.0 SR and 8.5 CSR, compared with 9.1 and 36.4 for the dual-scale planner. A shared map results in far more candidate nodes per step and the waypoint predictor contributes additional noise, and therefore dual-scale reasoning anchors attention on the local neighborhood. This is not a stylistic inheritance from DUET \cite{2022duet} but a requirement of the multi-agent regime. Finally, reducing navigation step limit to 20 saves time but lowers SR from 9.1 to 6.0, as exploration is curtailed before decisions can be grounded.

\begin{table}[t]
\renewcommand{\arraystretch}{1.1}
\setlength{\aboverulesep}{0pt}
\setlength{\belowrulesep}{0pt}
\setlength{\tabcolsep}{2.2pt}
\centering
\caption{Comparison of arrival-sequence matching strategies.}
\resizebox{0.95\linewidth}{!}{
\begin{tabular}{ll cccccc}
\toprule
&& \multicolumn{6}{c}{\textbf{Val-Unseen}} \\
\cmidrule(lr){3-8}
\textbf{Instruction Style}
& \textbf{Matching}
& \textbf{\texttt{SR}}$\uparrow$
& \textbf{\texttt{SPL}}$\uparrow$
& \textbf{\texttt{CSR}}$\uparrow$
& \textbf{\texttt{CSPL}}$\uparrow$
& \textbf{\texttt{ISPL}}$\uparrow$
& \textbf{\texttt{TC}}$\uparrow$ \\
\midrule

Decentralized
& greedy
& 8.3 & 3.3 & 35.3 & 18.5 & 22.5 & 98.1 \\

& embed
& 8.3 & 3.3 & 35.2 & 18.4 & 25.0 & 97.3 \\

\rowcolor{gray!15}
& $\Delta$
& +0.0 & +0.0 & -0.1 & -0.1 & +2.5 & -0.8 \\

\midrule

Centralized
& greedy
& 7.3 & 2.8 & 37.8 & 19.4 & 22.8 & 97.3 \\

& embed
& 7.5 & 3.1 & 34.2 & 18.0 & 23.6 & 95.6 \\

\rowcolor{gray!15}
& $\Delta$
& +0.2 & +0.3 & -3.6 & -1.4 & +0.8 & -1.7 \\

\midrule

Centralized Implicit
& greedy
& 5.1 & 2.1 & 35.2 & 18.3 & 22.3 & 98.1 \\

& embed
& 5.4 & 2.3 & 31.5 & 16.6 & 23.1 & 96.6 \\

\rowcolor{gray!15}
& $\Delta$
& +0.3 & +0.2 & -3.7 & -1.7 & +0.8 & -1.5 \\

\bottomrule
\end{tabular}
}
\label{tab:metric}
\end{table}

\begin{figure}[t]
\centering
    \includegraphics[scale=0.48]{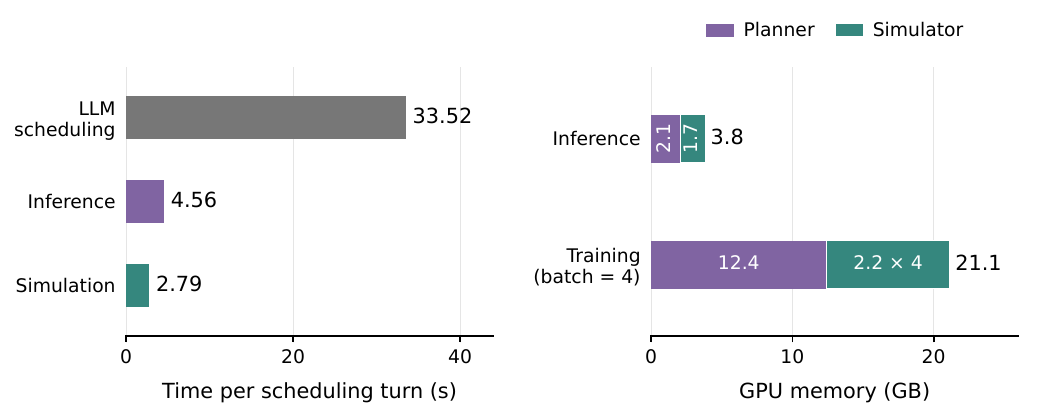}
    \caption{Latency breakdown per scheduling round and GPU memory consumption during training and inference.}
    \label{fig:computational_cost}
\end{figure}

\begin{table}[t]
\renewcommand{\arraystretch}{1.1}
\setlength{\aboverulesep}{0pt}
\setlength{\belowrulesep}{0pt}
\setlength{\tabcolsep}{2pt}
\centering
\caption{Performance of observation-guided scheduler}
\resizebox{0.95\linewidth}{!}{
{
\begin{threeparttable}
\begin{tabular}{c ccccc ccccc}
\toprule
& \multicolumn{5}{c}{\textbf{Val-Unseen}}
& \multicolumn{5}{c}{\textbf{Test-Unseen}} \\
\cmidrule(lr){2-6} \cmidrule(lr){7-11}
\textbf{Obs.}
& \textbf{\texttt{SR}}$\uparrow$
& \textbf{\texttt{SPL}}$\uparrow$
& \textbf{\texttt{CSR}}$\uparrow$
& \textbf{\texttt{CSPL}}$\uparrow$
& \textbf{\texttt{ISPL}}$\uparrow$
& \textbf{\texttt{SR}}$\uparrow$
& \textbf{\texttt{SPL}}$\uparrow$
& \textbf{\texttt{CSR}}$\uparrow$
& \textbf{\texttt{CSPL}}$\uparrow$
& \textbf{\texttt{ISPL}}$\uparrow$ \\
\midrule
\midrule

\multicolumn{11}{c}{\textbf{Decentralized Instruction Style}}\\

& \textbf{8.3} & \textbf{3.3} & \textbf{35.2} & 18.4 & 25.0 & 7.0 & 3.9 & 31.7 & 18.4 & \textbf{25.6} \\

\checkmark
& 7.8 & 3.0 & 34.7 & \textbf{18.5} & \textbf{25.3} & \textbf{7.7} & \textbf{4.0} & \textbf{33.0} & \textbf{19.5} & 23.6 \\

\midrule
\multicolumn{11}{c}{\textbf{Centralized Instruction Style}}\\

& \textbf{7.5} & 3.1 & \textbf{34.2} & \textbf{18.0} & \textbf{23.6} & \textbf{7.2} & \textbf{3.6} & \textbf{31.8} & 18.5 & \textbf{25.7} \\

\checkmark
& 7.3 & \textbf{3.3} & 33.0 & 17.7 & 23.1 & 6.5 & 3.5 & \textbf{31.8} & \textbf{19.3} & 25.2 \\

\midrule
\multicolumn{11}{c}{\textbf{Centralized Implicit Instruction Style}}\\

& 5.4 & 2.3 & \textbf{31.5} & 16.6 & \textbf{23.1} & 7.4 & 4.0 & 28.9 & 16.7 & \textbf{24.3} \\

\checkmark
& \textbf{5.7} & \textbf{2.4} & 30.6 & \textbf{16.8} & 22.2
& \textbf{7.9} & \textbf{4.2} & \textbf{31.2} & \textbf{18.6} & 23.9 \\

\bottomrule
\end{tabular}
\end{threeparttable}
}}
\label{tab:multimodal}
\end{table}

\begin{table}[t]
\renewcommand{\arraystretch}{1.1}
\setlength{\aboverulesep}{0pt}
\setlength{\belowrulesep}{0pt}
\setlength{\tabcolsep}{2.2pt}
\centering
\caption{Zero-shot navigation performance on a 50-episode subset}
\resizebox{\linewidth}{!}{
\begin{tabular}{ll ccccc cc}
\toprule
&& \multicolumn{7}{c}{\textbf{Val-Unseen}} \\
\cmidrule(lr){3-9}
\textbf{Instruction Style}
& \textbf{Planner}
& \textbf{\texttt{SR}}$\uparrow$
& \textbf{\texttt{SPL}}$\uparrow$
& \textbf{\texttt{CSR}}$\uparrow$
& \textbf{\texttt{CSPL}}$\uparrow$
& \textbf{\texttt{ISPL}}$\uparrow$
& \textbf{\texttt{TS}}$\downarrow$
& \textbf{\texttt{MAC}}$\downarrow$ \\
\midrule

Decentralized
& Default
& \textbf{10.0} & \textbf{2.8} & \textbf{37.2} & 16.9 & 23.2 & 1453 & 1.2 \\

& Hybrid
& 4.0 & 1.4 & 33.2 & \textbf{20.4} & \textbf{24.3} & 1343 & 1.2 \\

& VLM
& 2.0 & 0.7 & 20.8 & 13.2 & 17.3 & \textbf{977} & \textbf{0.9} \\

\midrule

Centralized
& Default
& \textbf{10.0} & 2.5 & \textbf{34.9} & 16.8 & \textbf{23.6} & 1502 & 0.9 \\

& Hybrid
& 6.0 & \textbf{2.9} & 29.8 & \textbf{17.9} & 22.5 & 1222 & \textbf{0.7} \\

& VLM
& 2.0 & 1.3 & 20.6 & 12.7 & 16.1 & \textbf{882} & \textbf{0.7} \\

\midrule

Centralized Implicit
& Default
& \textbf{6.0}& \textbf{2.3} & \textbf{36.0} & \textbf{18.0} & \textbf{24.8} & 1649 & 1.0 \\

& Hybrid
& 4.0 & 2.0 & 26.8 & 16.0 & 21.6 & 1172 & 0.8 \\

& VLM
& 2.0 & 1.1 & 12.6 & 9.1 & 14.0 & \textbf{862} & \textbf{0.4} \\

\bottomrule
\end{tabular}
}
\label{tab:zero_shot}
\end{table}

\noindent\textbf{Planner Training.}
\Cref{tab:ablation} also compares training recipes. Fully student-forced training outperforms 50\% teacher forcing in SR (9.1 vs.\ 6.0), although teacher forcing yields lower TS and higher CSPL, indicating that exposure to demonstrator trajectories narrows the exploration the student must ultimately perform alone. Training with four rather than two agents improves SR by 4.2 points and SPL by 1.5, supporting training with broader shared-memory exposure. We further ask whether pre-filling that memory helps, by persisting the topology across rounds; SR instead drops slightly when accumulation rises from one round to two, suggesting that an over-rich memory encourages reliance on inherited structure, which hurts precisely when little experience is available. Pretraining with REVERIE \cite{2020reverie} improves SPL from 2.4 to 3.4 over MAVLN-only pretraining, while the NavRAG configuration \cite{2025navrag} performs best at 4.0, consistent with its broader HM3D scene coverage. Basic navigation competence therefore scales with the knowledge acquired in pretraining.

\noindent\textbf{Conflict Resolving.}
\Cref{tab:conflict} isolates conflict-free target allocation (CFTA) and conflict-based path finding (CBPF). The vanilla framework attains the highest MAC at 2.2, together with the lowest TS and the highest SPL, since agents drive straight at their goals. Enabling CFTA lowers MAC to 1.4 at a slight cost in success, reflecting the compromise imposed by mutually exclusive target selection. Adding CBPF further reduces MAC to 0.8, indicating its effectiveness in avoiding interference at the price of 323 additional time steps.

\noindent\textbf{Topology Mapping and Maintenance.}
\Cref{tab:localization} examines the additional localization phase together with failure-verified exclusion (FVE) and traversal-verified connection (TVC). Localization lets an agent recover the most plausible viewpoint after control deviation, rather than inheriting the assumed viewpoint. Enabling either mechanism on top of localization improves performance, by 0.8 and 0.3 SPL respectively. The gain is larger for FVE, since control deadlocks against obstacles are more common. The full configuration performs best, and these results support verifying both failed proposals and traversed connections before updating shared topology.

\noindent\textbf{Metric Calculation.}
\Cref{tab:metric} compares our embedding-based instruction-to-subtask matching with a greedy strategy that maximizes task completion and uses conditional success to break ties. Mission-level metrics change little, but subtask metrics are more sensitive: greedy matching reports higher CSR by 0.1, 3.6 and 3.7 points under decentralized, centralized and centralized-implicit instructions respectively, while reporting lower ISPL throughout. The inflation tracks allocation freedom: it is negligible when each subtask admits only its designated agent, and grows once any agent may execute any subtask and the space of consistent interpretations widens accordingly. Faithful measurement therefore requires an explicit instruction-based mapping rather than a maximizing search.

\noindent\textbf{Computational Efficiency.}
\Cref{fig:computational_cost} reports inference latency and GPU memory. LLM scheduling dominates the reported per-round latency at 33.52\,s, compared with 4.56\,s for navigation planning and 2.79\,s for simulation, identifying scheduling as the principal target for future optimization and confirming that topological planning is efficient. The planner requires only 2.1\,GB of GPU memory at inference, comparable to the simulator itself, and training fits within 21.1\,GB in total, keeping the system reproducible on widely available hardware.

\begin{figure*}[t]
\centering
    \includegraphics[scale=0.5]{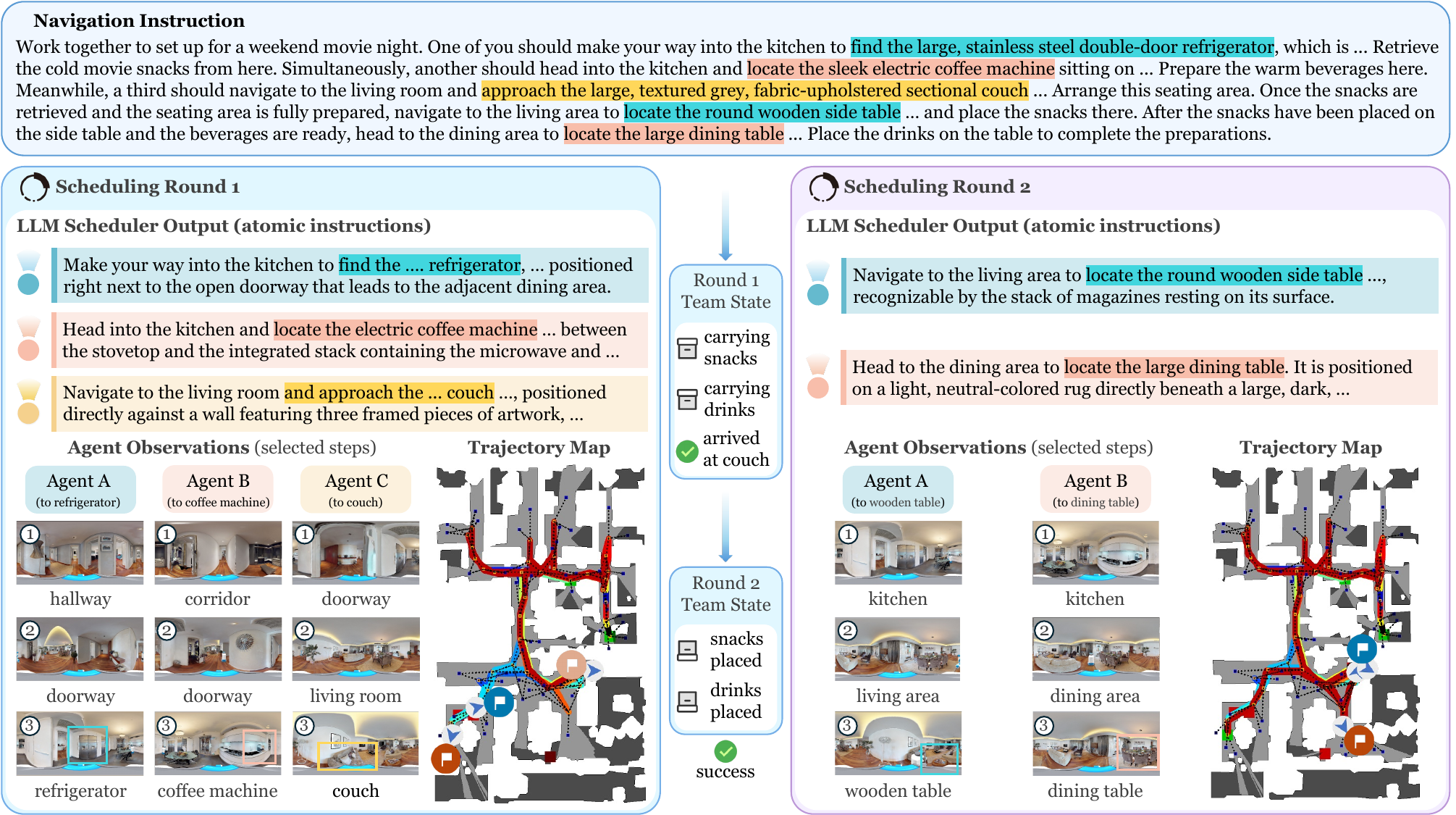}
    \caption{Scheduling and navigation process of TRISS on a representative episode. Agents A and B enter the holding state after the first scheduling round, and the mission completes at the end of the second. Subtask goals are drawn as flags in each agent's color.}
    \label{fig:case_study}
\end{figure*}

\begin{table}[t]
\renewcommand{\arraystretch}{1.1}
\setlength{\aboverulesep}{0pt}
\setlength{\belowrulesep}{0pt}
\setlength{\tabcolsep}{2.2pt}
\centering
\caption{Comparison against an oracle planner}
\resizebox{\linewidth}{!}{
\begin{tabular}{l ccccc ccccc}
\toprule
& \multicolumn{5}{c}{\textbf{Val-Unseen}}
& \multicolumn{5}{c}{\textbf{Test-Unseen}} \\
\cmidrule(lr){2-6} \cmidrule(lr){7-11}

\textbf{Planner}
& \textbf{\texttt{SR}}$\uparrow$
& \textbf{\texttt{SPL}}$\uparrow$
& \textbf{\texttt{CSR}}$\uparrow$
& \textbf{\texttt{CSPL}}$\uparrow$
& \textbf{\texttt{TS}}$\downarrow$
& \textbf{\texttt{SR}}$\uparrow$
& \textbf{\texttt{SPL}}$\uparrow$
& \textbf{\texttt{CSR}}$\uparrow$
& \textbf{\texttt{CSPL}}$\uparrow$
& \textbf{\texttt{TS}}$\downarrow$ \\
\midrule

Ours
& 9.1
& 4.0
& 36.4
& 19.7
& 1588
& 9.3
& 5.2
& 33.3
& 20.0
& 1494 \\

Oracle
& \textbf{89.1}
& \textbf{73.4}
& \textbf{96.3}
& \textbf{81.8}
& \textbf{640}
& \textbf{81.0}
& \textbf{64.0}
& \textbf{91.0}
& \textbf{74.9}
& \textbf{769} \\

\bottomrule
\end{tabular}
}
\label{tab:planner}
\end{table}

\noindent\textbf{Observation-Guided Scheduler.}
Decentralized instructions distribute the agent assignment explicitly, whereas centralized and centralized-implicit instructions progressively withhold it, leaving the scheduler to infer an allocation. We analyze whether scheduling benefits from observations that hint at promising locations. \Cref{tab:multimodal} compares the language-only scheduler against an observation-guided variant on Qwen3.6. Centralized-implicit instructions show small gains. Centralized / decentralized instructions show no consistent improvement, suggesting that perceptual context can also distract reasoning. Closing this gap charts a robust scheduler that grounds allocation in the team's spatial state and approaches optimality.

\noindent\textbf{Zero-Shot Navigation.}
\Cref{tab:zero_shot} compares the trained planner with Qwen3.6 as a zero-shot VLM planner and a hybrid that queries the VLM only when the trained planner is uncertain. The trained planner outperforms the zero-shot VLM in SR across all three instruction regimes, by 8.0, 8.0, and 4.0 points. The hybrid variant recovers part of the gap but still trails the trained planner. Notably, the VLM planner attains the lowest TS and MAC while failing most missions, which reflects early termination rather than efficiency. Executing coarse-grained atomic instructions over a shared topology therefore remains beyond current zero-shot VLMs.

\begin{table}[t]
\renewcommand{\arraystretch}{1.1}
\setlength{\aboverulesep}{0pt}
\setlength{\belowrulesep}{0pt}
\setlength{\tabcolsep}{3.5pt}
\centering
\caption{Failure statistics of conflict-based path finding}
\resizebox{0.9\linewidth}{!}{
\begin{tabular}{l ccc ccc}
\toprule
& \multicolumn{3}{c}{\textbf{Val-Unseen}}
& \multicolumn{3}{c}{\textbf{Test-Unseen}} \\
\cmidrule(lr){2-4} \cmidrule(lr){5-7}

\textbf{Scheduler}
& \textbf{Total}
& \textbf{Unres.}$\downarrow$
& \textbf{Inter.}$\downarrow$
& \textbf{Total}
& \textbf{Unres.}$\downarrow$
& \textbf{Inter.}$\downarrow$ \\
\midrule

Qwen3.6
& 18,508
& 3.78
& 1.02
& 21,057
& 4.41
& 1.26 \\

Oracle
& 18,087
& 3.98
& 1.19
& 21,062
& 5.05
& 1.55 \\

\bottomrule
\end{tabular}
}
\label{tab:failure}
\end{table}

\noindent\textbf{Topological Planning Limitations.}
\Cref{tab:planner} bounds the current paradigm with an oracle planner, raising SR from 9.1 to 89.1 on val-unseen and from 9.3 to 81.0 on test-unseen, identifying planning as a major bottleneck. Even granted perfect node selection, topological planning is thus capped well below saturation. Viewpoint prediction noise compounds over an episode, so a target may simply never be proposed, and the resulting spurious nodes corrupt memory. This charts future work on robust mapping modules, or on planning paradigms that dispense with predicted viewpoints while preserving memory-wise coordination.

\noindent\textbf{Conflict Resolving Limitations.}
\Cref{tab:failure} reflects failures of the bounded search to return conflict-free routes. 3.78-3.98\% of transitions remain unresolved on val-unseen and 4.41-5.05\% on test-unseen. Geometrically intersecting transitions occur in 1.02-1.55\% of cases. Such residual rates would remain unacceptable in physical deployments, charting stronger search and geometric checks during execution.

\begin{figure*}[t]
\centering
    \includegraphics[scale=0.5]{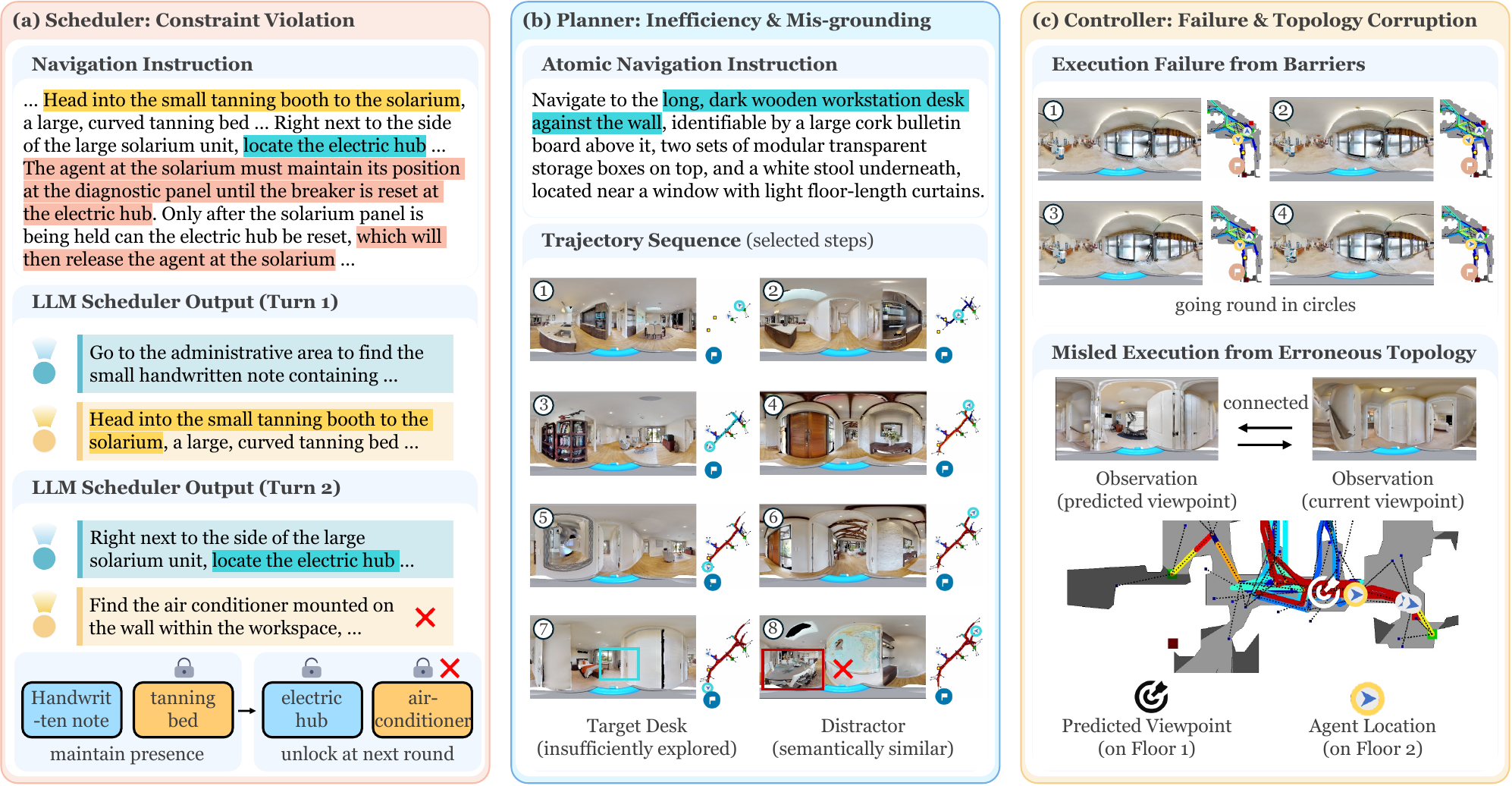}
    \caption{Failure modes. The scheduler case violates a presence lock by re-dispatching an agent that should have held position; the planner case stops at a distractor while leaving the target insufficiently explored; the controller case shows an execution failure and an erroneous edge misleading later execution.}
    \label{fig:failure_cases}
\end{figure*}

\subsection{Qualitative Analysis}
\Cref{fig:case_study} illustrates successful coordination across five subtasks and two scheduling rounds. In the first round, agents A and B reach the refrigerator and coffee machine, entering holding states associated with carrying snacks and drinks, while agent C reaches the couch. In the second round, A and B are assigned the side table and dining table, respectively, and complete the mission under the task constraints. This example illustrates how TRISS preserves agent assignments across holding chains while coordinating dependent subtasks. \Cref{fig:failure_cases} highlights three complementary failure modes. The scheduler violates an explicit presence constraint by re-dispatching the agent at the solarium before its teammate resets the electric hub. The planner repeatedly revisits candidate areas and stops at a semantically similar distractor, leaving the target desk insufficiently explored. The controller encounters execution failures at obstacles that produce deadlocks, and these failures can lead to nonexistent connections in the shared map, while an erroneous connection between viewpoints on different floors misleads subsequent navigation.

Across the analyses above, the headroom MAVLN leaves is attributable to no single module. Schedulers fail to keep language-stated constraints live through execution, which neither perceptual context nor their dominant share of latency currently redeems; planners stay capped by waypoint proposal even under oracle node selection, and the task remains well beyond zero-shot VLMs; controllers still suffer from failures, leaving residual conflicts unresolved.

\section{Conclusion}
This work introduces Systematic Multi-Agent Vision-and-Language Navigation, formulating embodied multi-agent navigation as a coordination problem in which a mission decomposes into subtasks carrying dependency and resource constraints. We present an automatic generation engine that yields MAVLN released with constraint-aware metrics. We further present TRISS, a coordination-ready navigation system integrating an LLM-based subtask scheduler, a shared topological memory and a conflict-aware execution mechanism. Experiments establish TRISS as a comprehensive baseline while measuring how far the task remains from solved, as absolute success rates stay in the single digits, and the headroom is distributed across scheduling, planning, and execution rather than concentrated in any one. We release the benchmark, platform and code to support the work of closing it.

\bibliographystyle{IEEEtran}
\bibliography{shortstrings,references}

%








\section{Biography Section}

\vspace{-21pt}
\begin{IEEEbiography}[{\includegraphics[width=1in,height=1.25in,clip,keepaspectratio]{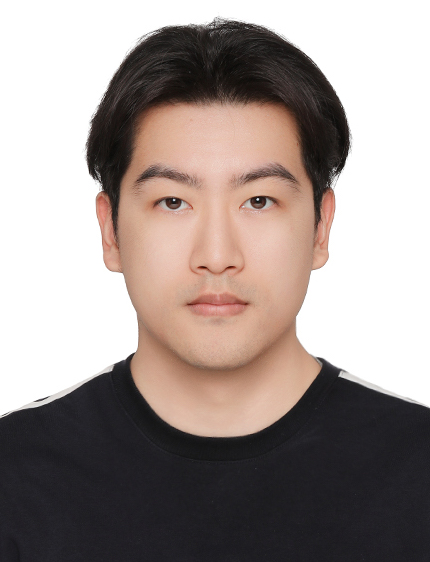}}]{Yunzhe Xu} received the bachelor's degree in software engineering from Harbin Institute of Technology in 2022. He is currently pursuing the Ph.D. degree in computer science and technology with Shanghai Jiao Tong University. His research interests include embodied navigation systems, robotic learning and large language model agents. He has published papers in top-tier AI venues, including IEEE T-PAMI, AAAI and NeurIPS.
\end{IEEEbiography}

\vspace{-21pt}
\begin{IEEEbiography}[{\includegraphics[width=1in,height=1.25in,clip,keepaspectratio]{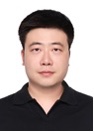}}]{Zhe Liu} received the Ph.D. degree in control technology and control engineering from Shanghai Jiao Tong University, Shanghai, China, in 2016. From 2017 to 2020, he was a Post-Doctoral Fellow with the Department of Mechanical and Automation Engineering, The Chinese University of Hong Kong, Hong Kong. From 2020 to 2022, he was a Research Associate with the Department of Computer Science and Technology, University of Cambridge, Cambridge, U.K. From 2022 to 2025, he has been an Associate Professor with the AI institute, Shanghai Jiao Tong University, where he is currently an Associate Professor with the Department of Automation. His current research interests include multi-robot cooperation and autonomous driving system. He is a Senior Editor of IEEE Transactions on Automation Science and Engineering, an Associate Editor of IEEE Transactions on Robotics, and an Editor of IEEE International Conference on Robotics and Automation.
\end{IEEEbiography}
\vfill

\end{document}


\title{Supplementary Material for ``Systematic Multi-Agent Vision-and-Language Navigation: Formulation, Benchmark, and Method''}

\maketitle

\noindent This supplementary material is organized as follows. \Cref{app:impl} documents the simulation platform, perception backbone, pretraining, imitation learning and inference runtime. \Cref{app:data} analyzes the generation funnel and the composition of MAVLN. \Cref{app:exp} reports additional experiments. \Cref{app:eval} formalizes the goal anchors, the constraint-aware metrics and the instruction-to-subtask matching. \Cref{app:method} unfolds the three modules of TRISS, including the full episode loop, the scheduler protocol, the shared-topology maintenance, the planner architecture, conflict-aware execution and the training stream. \Cref{app:crafting} details verified task crafting. \Cref{tab:hyper} consolidates the constants of the generation pipeline, the evaluation protocol and TRISS. The prompt templates are reproduced verbatim on the last pages of this document, grouped by the pipeline step they serve: \Cref{fig:prompt_stage12} for grounded scene understanding and mission synthesis, \Cref{fig:prompt_stage3} for mission refinement, \Cref{fig:prompt_stage4} for verified instruction rendering, \Cref{fig:prompt_sched} for the subtask scheduler of TRISS, and \Cref{fig:prompt_nav} for its VLM-assisted planners.

\section{Implementation Details}
\label{app:impl}

\subsection{Simulation Platform}
\label{app:platform}
Inference and evaluation run on a synchronized multi-agent platform that we build on Habitat~3 (habitat-lab 0.3.1) with HM3D scenes, whereas planner training reuses the single-agent VLN-CE stack of ETPNav (habitat-lab 0.1.7) and shares the topological memory across simulator instances (\Cref{app:training}). Each episode instantiates up to four Fetch articulated agents in kinematic mode. Every agent carries twelve RGB cameras ($224\times224$) and twelve depth cameras ($256\times256$, range $[0,10]$\,m, normalized), all with a $90^{\circ}$ horizontal field of view and oriented at $30^{\circ}$ heading increments, so that each step yields the panoramic observation.

Low-level motion is realized by a per-agent velocity controller integrated over a $1$\,s time step: a unit linear command translates the agent by $0.25$\,m and a unit angular command rotates it by $15^{\circ}$, reproducing the action granularity of VLN-CE. A single environment step advances all agents simultaneously, so the step counter defines one synchronized clock for the whole team. The Time Steps (TS) metric counts these steps, and the Multi-Agent Conflict (MAC) metric treats each agent as a disc of radius $\rho=0.25$\,m. Subtask arrivals are credited within a $3$\,m geodesic radius of the nearest navigable anchor of the target instance (\Cref{app:anchors}).

\subsection{Perception Backbone and Waypoint Prediction}
\label{app:backbone}
Following ETPNav, RGB views are encoded by a CLIP image encoder into $512$-dimensional vectors, and depth views are encoded by a ResNet-50 pretrained with DD-PPO point-goal navigation on Gibson, whose $128\times4\times4$ feature maps are average-pooled into $128$-dimensional vectors. Both encoders are kept frozen and in evaluation mode throughout training. Neighboring viewpoints are proposed by the transformer-based waypoint predictor, using the depth-driven checkpoint released with ETPNav for a $90^{\circ}$ field of view, which is likewise frozen. The predictor outputs a heatmap over $120$ heading bins ($3^{\circ}$ each) and $12$ distance bins ($0.25$\,m to $3$\,m in $0.25$\,m increments), which exposes the planner to plausible topologies.

\subsection{Pretraining}
\label{app:pretrain}
The cross-modal planner is initialized from BERT-base-uncased and pretrained on discrete navigation graphs following DUET and ETPNav. The corpus merges the NavRAG training set with the ground-truth pairs of atomic instructions and subtask trajectories extracted from the MAVLN training split: every subtask yields one pair whose instruction is its verified atomic instruction, whose path is the navigation-graph shortest path from the goal of the previous subtask of the same agent (or its start pose) to the current goal viewpoint, and whose positive endpoint is that goal viewpoint. All NavRAG trajectories located in the val-unseen and test-unseen scenes of MAVLN are removed beforehand. Two proxy tasks are optimized jointly. Masked Language Modeling (MLM) randomly masks instruction tokens and recovers them from the cross-modal text states conditioned on the topological graph built along the full demonstration. Single-step Action Prediction (SAP) truncates the demonstration at a random step, builds the graph from the observed prefix, and supervises the expert next node with fused logits (\Cref{app:planner}). Pretraining runs for $200{,}000$ steps, and the checkpoint with the best navigation performance on val-unseen (step $180{,}000$) initializes imitation learning.

\subsection{Imitation Learning}
\label{app:il}
Imitation learning uses the per-subtask episodes of the MAVLN training split: every subtask of the decentralized rendering becomes one single-agent episode whose instruction is the verified atomic instruction, whose start pose is the goal pose of the previous subtask of the same agent (or the mission start pose for its first subtask), whose initial heading faces the final view of that previous trajectory, and whose goal is the subtask goal viewpoint with a $3$\,m success radius. Four simulator instances run concurrently, each hosting one agent, and instances whose current episodes belong to the same scene share one topological memory (\Cref{app:training}). Each iteration rolls out one episode per instance for at most $T_{\max}=30$ planning steps under student forcing, and performs one AdamW update with learning rate $10^{-5}$. Training lasts $30{,}000$ iterations with a checkpoint every $1{,}000$ iterations, and the checkpoint maximizing the sum of SR and SPL on val-unseen is retained. The visual encoders and the waypoint predictor stay frozen, while the language, panorama and cross-modal encoders are updated. Training fits on a single A100 GPU.

\subsection{Inference Runtime}
\label{app:runtime}
The LLM scheduler is served locally through an OpenAI-compatible endpoint with JSON-formatted responses, using Qwen3.6-27b (FP8) or Gemma4-31b (W4A16 quantization-aware checkpoint); the same backbone performs instruction decomposition, scheduling and, when enabled, observation description and zero-shot planning (\Cref{app:hybrid}). The observation-guided and VLM-assisted variants additionally tag each candidate view with the RAM++ image tagger (Swin-L backbone) at $384\times384$ resolution. All LLM-dependent results are averaged over three runs.

\section{Dataset Details}
\label{app:data}

\begin{figure}[t]
\centering
\includegraphics[width=\linewidth]{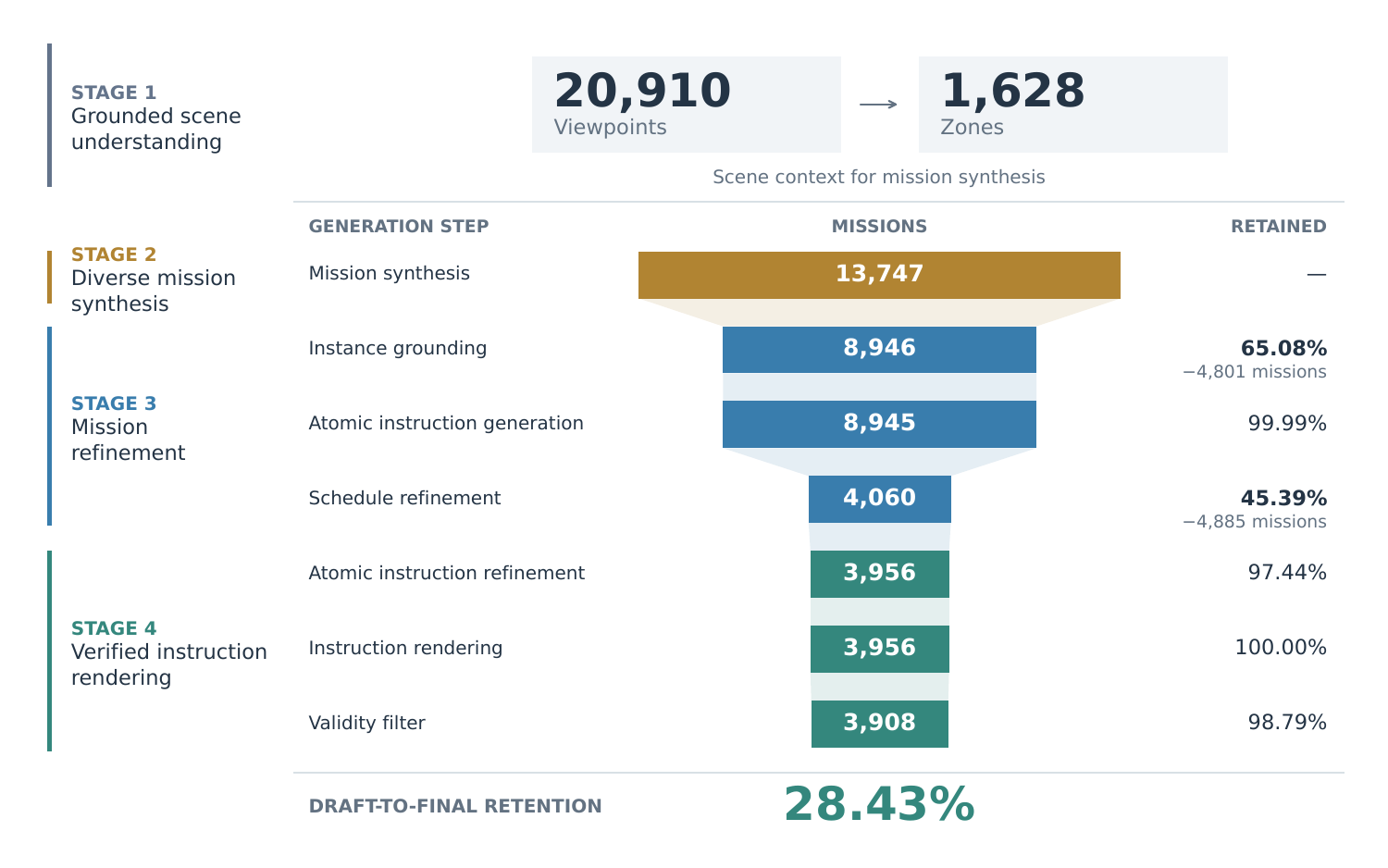}
\caption{Generation funnel of MAVLN. Stage~1 produces the viewpoint and zone annotations consumed by mission synthesis; subsequent rows report the missions retained after each generation step and the retention ratio relative to the preceding step.}
\label{fig:funnel}
\end{figure}

\subsection{Generation Funnel}
\label{app:funnel}
\Cref{fig:funnel} traces every mission draft through the four crafting stages detailed in \Cref{app:crafting}. Grounded scene understanding annotates $20{,}910$ viewpoints and aggregates them into $1{,}628$ repaired zones, which form the scene context of mission synthesis. From a synthesis budget of $14{,}000$ missions distributed over scenes and zones in proportion to their grounded viewpoints, $13{,}747$ drafts are returned in a valid format. Instance grounding retains $8{,}946$ of them ($65.08\%$): the $4{,}801$ discarded drafts violating the well-formedness rules (\Cref{tab:wellformed}). Atomic instruction generation loses a single draft. Schedule refinement is the most selective step, retaining $4{,}060$ of $8{,}945$ missions ($45.39\%$), because it rewrites the allocation by the makespan-optimal schedule and then enforces the navigation-graph and continuous-space leg bounds, the start-pose constraints and the start-to-target separation on every candidate instance combination; a mission is released only if at least one combination satisfies all of them. Perception-verified refinement of atomic instructions retains $3{,}956$ missions ($97.44\%$), dropping those for which the VLM cannot bind the instruction to a candidate instance with a visible mask. All $3{,}956$ missions are rendered in the three instruction regimes, and the validity filter finally retains $3{,}908$ ($98.79\%$), discarding every mission whose instructions fail round-trip reconstruction or human assessment (val-unseen and test-unseen split) in any regime. The draft-to-final retention is therefore $28.43\%$. Two observations support the reliability of MAVLN. First, the heaviest attrition is caused by programmatic geometric and scheduling checks rather than by LLM-judged quality, so the released distribution is shaped by verifiable constraints. Second, the last filter removes only $1.21\%$ of the rendered missions, which indicates that instructions produced from perception-verified atomic instructions under our rendering rules are already faithful to the constraint structure before filtering, and that the filter acts as a safeguard rather than as the primary source of quality.

\begin{figure}[t]
\centering
\subfloat[Scenario domains]{\includegraphics[width=0.48\linewidth]{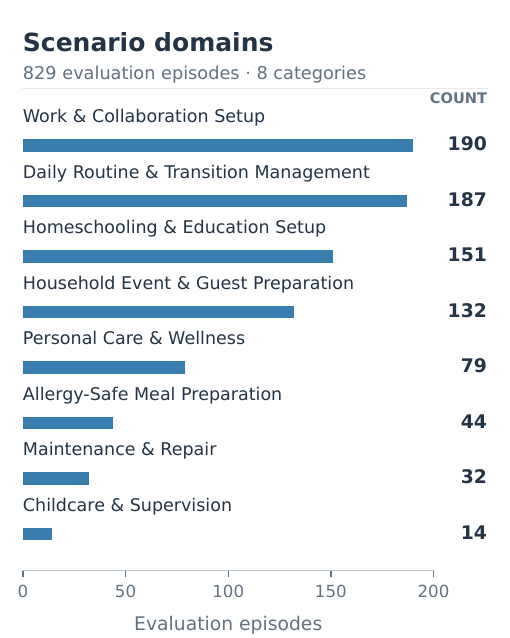}\label{fig:scenario_taxonomy}}\hfil
\subfloat[Target categories]{\includegraphics[width=0.48\linewidth]{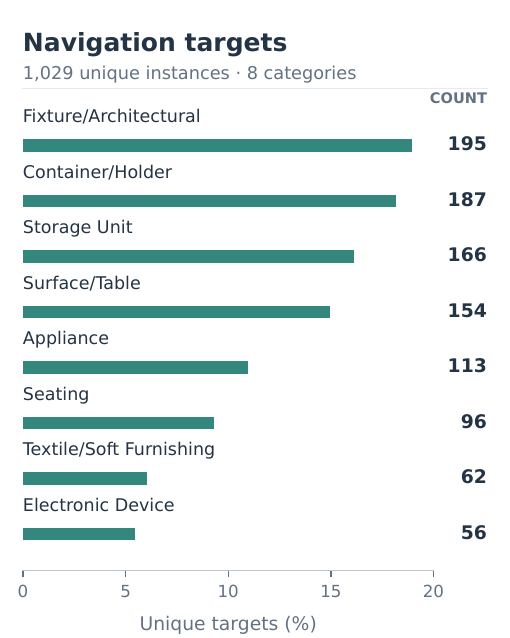}\label{fig:target_taxonomy}}
\caption{Composition of the evaluation splits. \textbf{(a)} Scenario domains of the $829$ evaluation missions. \textbf{(b)} Categories of the $1{,}029$ unique navigation target instances.}
\label{fig:taxonomy}
\end{figure}

\subsection{Scenario and Target Composition}
\label{app:taxonomy}
\Cref{fig:taxonomy} characterizes the $829$ evaluation missions (val-unseen and test-unseen) and their navigation targets. The scenario domains are led by work and collaboration setup ($190$, $22.9\%$), daily routine and transition management ($187$, $22.6\%$), homeschooling and education setup ($151$, $18.2\%$), and household event and guest preparation ($132$, $15.9\%$), followed by personal care and wellness ($79$, $9.5\%$), allergy-safe meal preparation ($44$, $5.3\%$), maintenance and repair ($32$, $3.9\%$), and childcare and supervision ($14$, $1.7\%$). The four leading domains cover $79.6\%$ of the evaluation missions, reflecting the household profiles that most often match the evaluation scenes, while the long tail keeps constraint-heavy scenarios such as supervision and care, in which continuous presence is a natural source of presence locks. The $1{,}029$ unique target instances are spread over eight categories without a dominant one: fixtures and architectural elements ($195$, $19.0\%$), containers and holders ($187$, $18.2\%$), storage units ($166$, $16.1\%$), surfaces and tables ($154$, $15.0\%$), appliances ($113$, $11.0\%$), seating ($96$, $9.3\%$), textiles and soft furnishings ($62$, $6.0\%$), and electronic devices ($56$, $5.4\%$). No category exceeds a fifth of the targets, and a substantial portion consists of small or visually ambiguous objects such as containers, holders and devices, which must be distinguished from same-category distractors through the fine-grained attributes injected by perception-verified refinement.

\begin{figure}[t]
\centering
\includegraphics[width=0.45\linewidth]{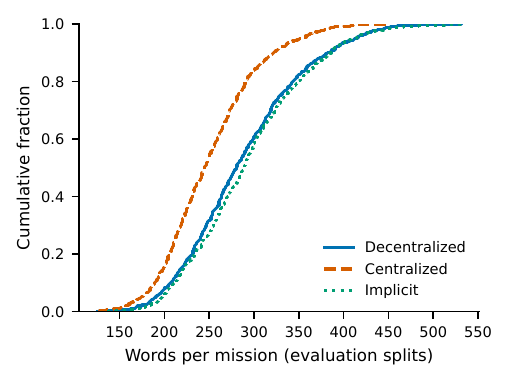}
\caption{Empirical cumulative distribution of words per mission on the evaluation splits under the decentralized, centralized and centralized-implicit regimes.}
\label{fig:ecdf}
\end{figure}

\subsection{Instruction Length}
\label{app:length}
\Cref{fig:ecdf} plots the empirical cumulative distribution of words per mission on the evaluation splits for the three regimes. Mission instructions run to several hundred words, far longer than typical single-agent VLN instructions, because every atomic destination is described with the perception-verified attributes that separate it from nearby distractors and because each regime additionally verbalizes the coordination structure. The decentralized regime distributes the mission over per-agent instructions, each of which restates the cross-agent conditions gating its subtasks; the centralized regime consolidates the plan into one instruction; and the centralized-implicit regime removes agent attribution while keeping every completion dependency, so its length reflects the constraint content rather than the allocation. Consequently, the regimes differ in how the constraint structure is expressed rather than in the amount of navigational content, which is shared across them by construction.

\begin{figure}[t]
\centering
\includegraphics[width=\linewidth]{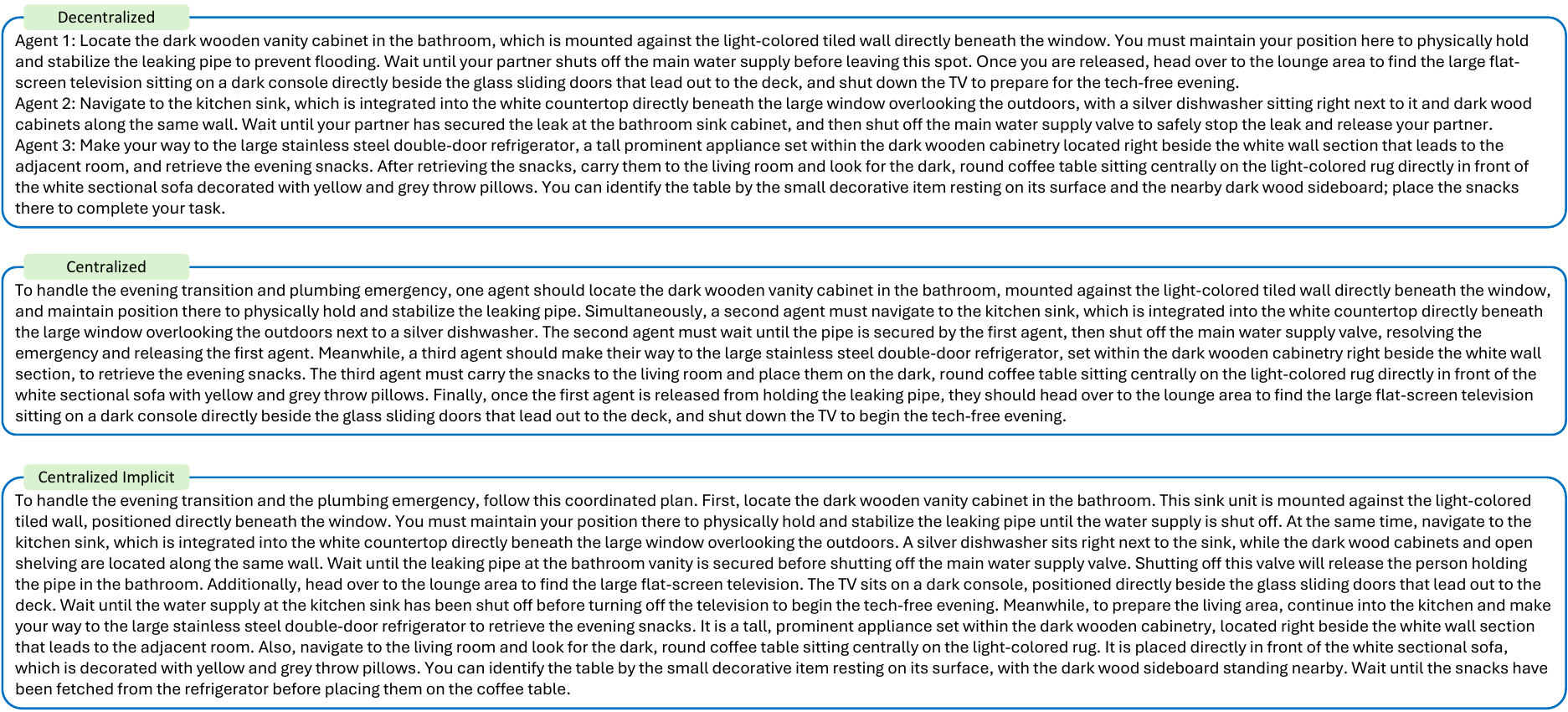}
\caption{Mission instructions of the example episode, rendered under the decentralized, centralized and centralized-implicit regimes.}
\label{fig:instruction}
\end{figure}

\begin{figure}[t]
\centering
\includegraphics[width=\linewidth]{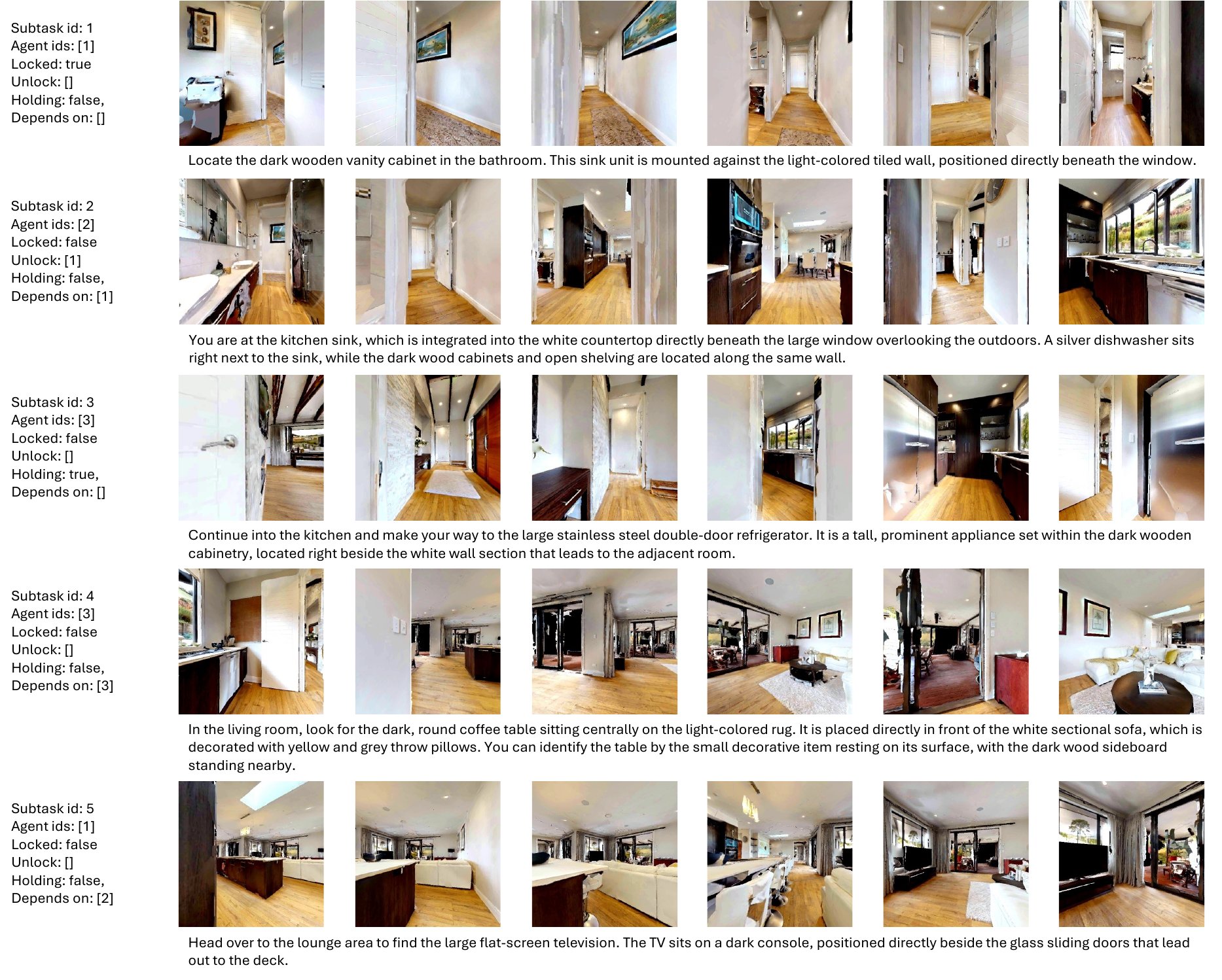}
\caption{Constraint structure, per-subtask atomic instructions and rendered agent trajectories of the same example episode.}
\label{fig:trajectory}
\end{figure}

\subsection{Episode Visualization}
\label{app:example}
\Cref{fig:instruction,fig:trajectory} illustrate one released episode end to end, so that the artifacts described above can be inspected on a concrete mission. \Cref{fig:instruction} shows the three renderings of the same mission. The decentralized rendering assigns one instruction to every agent and restates, inside each of them, the cross-agent conditions that gate its subtasks; the centralized rendering merges the plan into a single instruction that still names the executor of every destination; and the centralized-implicit rendering removes all agent attribution while preserving every completion dependency, which is exactly the information a system must recover in order to allocate the subtasks itself. \Cref{fig:trajectory} shows the underlying structure of the same episode: the subtask graph with its dependency, holding and locking attributes, the atomic instruction of every subtask, and the rendered trajectory of every agent through its goal viewpoints. Reading the two figures together exposes the correspondence that verified task crafting establishes and that evaluation relies upon.

\section{Additional Experiments}
\label{app:exp}

\begin{table*}[t]
\renewcommand{\arraystretch}{1.1}
\setlength{\aboverulesep}{0pt}
\setlength{\belowrulesep}{0pt}
\setlength{\tabcolsep}{1.6pt}
\centering
\caption{Performance on MAVLN under oracle navigation with LLM scheduling.}
\resizebox{\textwidth}{!}{
\begin{threeparttable}
\begin{tabular}{lc cccccc c cccccc}
\toprule
&& \multicolumn{6}{c}{\textbf{Val-Unseen}} && \multicolumn{6}{c}{\textbf{Test-Unseen}} \\
\cmidrule(lr){3-8} \cmidrule(lr){10-15}
\textbf{Scheduler} & \textbf{Navigator}
& \textbf{\texttt{SR}} $\uparrow$
& \textbf{\texttt{SPL}} $\uparrow$
& \textbf{\texttt{CSR}} $\uparrow$
& \textbf{\texttt{CSPL}} $\uparrow$
& \textbf{\texttt{ISPL}} $\uparrow$
& \textbf{\texttt{TC}} $\uparrow$
&& \textbf{\texttt{SR}} $\uparrow$
& \textbf{\texttt{SPL}} $\uparrow$
& \textbf{\texttt{CSR}} $\uparrow$
& \textbf{\texttt{CSPL}} $\uparrow$
& \textbf{\texttt{ISPL}} $\uparrow$
& \textbf{\texttt{TC}} $\uparrow$ \\
\midrule

\multicolumn{15}{c}{\textbf{Decentralized Instruction Style}} \\
Gemma4 & Oracle
& 82.6\scriptsize{$\pm$1.3} & 82.5\scriptsize{$\pm$1.3} & 94.5\scriptsize{$\pm$0.5} & 94.4\scriptsize{$\pm$0.5} & 87.6\scriptsize{$\pm$0.5} & 94.5\scriptsize{$\pm$0.5}
&& 84.2\scriptsize{$\pm$0.8} & 84.2\scriptsize{$\pm$0.8} & 95.5\scriptsize{$\pm$0.1} & 95.5\scriptsize{$\pm$0.1} & 88.3\scriptsize{$\pm$0.2} & 95.8\scriptsize{$\pm$0.2} \\
Qwen3.6 & Oracle
& 92.3\scriptsize{$\pm$0.7} & 92.2\scriptsize{$\pm$0.7} & 96.6\scriptsize{$\pm$0.3} & 96.4\scriptsize{$\pm$0.3} & 89.6\scriptsize{$\pm$0.3} & 96.9\scriptsize{$\pm$0.4}
&& 92.7\scriptsize{$\pm$0.6} & 92.7\scriptsize{$\pm$0.6} & 97.0\scriptsize{$\pm$0.4} & 96.8\scriptsize{$\pm$0.4} & 89.7\scriptsize{$\pm$0.3} & 97.4\scriptsize{$\pm$0.3} \\

\midrule
\multicolumn{15}{c}{\textbf{Centralized Instruction Style}} \\
Gemma4 & Oracle
& 82.0\scriptsize{$\pm$1.0} & 81.4\scriptsize{$\pm$1.1} & 93.9\scriptsize{$\pm$0.6} & 92.7\scriptsize{$\pm$0.6} & 87.6\scriptsize{$\pm$0.4} & 94.6\scriptsize{$\pm$0.5}
&& 84.2\scriptsize{$\pm$1.0} & 83.1\scriptsize{$\pm$1.0} & 94.5\scriptsize{$\pm$0.3} & 92.8\scriptsize{$\pm$0.3} & 87.5\scriptsize{$\pm$0.1} & 95.1\scriptsize{$\pm$0.2} \\
Qwen3.6 & Oracle
& 86.9\scriptsize{$\pm$0.7} & 86.1\scriptsize{$\pm$0.9} & 94.0\scriptsize{$\pm$0.2} & 92.5\scriptsize{$\pm$0.4} & 88.5\scriptsize{$\pm$0.0} & 95.8\scriptsize{$\pm$0.0}
&& 90.7\scriptsize{$\pm$0.2} & 89.4\scriptsize{$\pm$0.2} & 95.3\scriptsize{$\pm$0.4} & 93.5\scriptsize{$\pm$0.4} & 88.9\scriptsize{$\pm$0.4} & 96.8\scriptsize{$\pm$0.4} \\

\midrule
\multicolumn{15}{c}{\textbf{Centralized Implicit Instruction Style}} \\
Gemma4 & Oracle
& 70.2\scriptsize{$\pm$0.9} & 67.2\scriptsize{$\pm$0.9} & 89.4\scriptsize{$\pm$0.7} & 82.4\scriptsize{$\pm$0.8} & 85.0\scriptsize{$\pm$0.3} & 92.6\scriptsize{$\pm$0.3}
&& 71.4\scriptsize{$\pm$0.7} & 65.9\scriptsize{$\pm$0.6} & 89.9\scriptsize{$\pm$0.6} & 81.2\scriptsize{$\pm$0.4} & 84.9\scriptsize{$\pm$0.4} & 93.1\scriptsize{$\pm$0.5} \\
Qwen3.6 & Oracle
& 80.0\scriptsize{$\pm$1.2} & 76.2\scriptsize{$\pm$1.2} & 91.1\scriptsize{$\pm$0.9} & 83.8\scriptsize{$\pm$0.9} & 88.8\scriptsize{$\pm$0.5} & 97.0\scriptsize{$\pm$0.3}
&& 82.1\scriptsize{$\pm$0.1} & 75.7\scriptsize{$\pm$0.1} & 92.0\scriptsize{$\pm$0.2} & 82.7\scriptsize{$\pm$0.1} & 87.9\scriptsize{$\pm$0.4} & 96.8\scriptsize{$\pm$0.5} \\

\bottomrule
\end{tabular}
\end{threeparttable}}
\label{tab:oracle_all_results}
\end{table*}

\subsection{Scheduling with Oracle Navigation}
\label{app:oraclenav}
\Cref{tab:oracle_all_results} isolates the LLM scheduler by replacing planning and control with the oracle navigator of \Cref{app:oracles}. Under the oracle scheduler, oracle navigation completes every mission, so any deficit in this table is attributable to scheduling. Scheduling alone already costs a substantial share of missions. Under decentralized instructions, Qwen3.6 reaches $92.3$ and $92.7$ SR on val-unseen and test-unseen, and Gemma4 reaches $82.6$ and $84.2$. Under centralized instructions, Qwen3.6 drops to $86.9$ and $90.7$ SR, while Gemma4 remains at $82.0$ and $84.2$. The centralized-implicit regime is the hardest, with $80.0$ and $82.1$ SR for Qwen3.6 and $70.2$ and $71.4$ for Gemma4, confirming that inferring the allocation on top of the constraint structure is a distinct source of difficulty.

The subtask-level metrics locate the failures. TC stays between $92.6$ and $97.4$ and CSR between $89.4$ and $97.0$ in all configurations, so most failures are not missing subtasks but a few constraint violations or premature terminations that invalidate an otherwise completed mission. Path efficiency degrades with allocation freedom: under decentralized instructions SPL is within $0.1$ of SR, whereas under centralized-implicit instructions the gap grows to $3.0$--$6.4$ points and CSPL falls $7.0$--$9.3$ points below CSR, indicating that when the scheduler must choose executors, it often chooses feasible but longer allocations than the makespan-optimal reference. ISPL remains between $84.9$ and $89.7$ throughout, below the $92.6$ and $92.0$ attained under oracle scheduling, which shows that scheduling also affects how far agents travel between declarations. Qwen3.6 outperforms Gemma4 in SR in every regime and split under oracle navigation, although the two backbones perform comparably with the learned navigator; the navigation bottleneck therefore masks scheduling differences that become visible once navigation is solved. Consequently, improvements in scheduling are necessary but not sufficient: their benefit is bounded by navigation quality, while their absence caps performance even under perfect navigation.

\begin{table*}[t]
\renewcommand{\arraystretch}{1.1}
\setlength{\aboverulesep}{0pt}
\setlength{\belowrulesep}{0pt}
\setlength{\tabcolsep}{3.0pt}
\centering
\caption{Comparison of embedding-based and LLM-annotated instruction-to-subtask matching.}
\resizebox{0.9\linewidth}{!}{
\begin{tabular}{ll cccccc cccccc}
\toprule
&& \multicolumn{6}{c}{\textbf{Val-Unseen}}
& \multicolumn{6}{c}{\textbf{Test-Unseen}} \\
\cmidrule(lr){3-8} \cmidrule(lr){9-14}
\textbf{Instruction Style}
& \textbf{Matching}
& \textbf{\texttt{SR}}$\uparrow$
& \textbf{\texttt{SPL}}$\uparrow$
& \textbf{\texttt{CSR}}$\uparrow$
& \textbf{\texttt{CSPL}}$\uparrow$
& \textbf{\texttt{ISPL}}$\uparrow$
& \textbf{\texttt{TC}}$\uparrow$
& \textbf{\texttt{SR}}$\uparrow$
& \textbf{\texttt{SPL}}$\uparrow$
& \textbf{\texttt{CSR}}$\uparrow$
& \textbf{\texttt{CSPL}}$\uparrow$
& \textbf{\texttt{ISPL}}$\uparrow$
& \textbf{\texttt{TC}}$\uparrow$ \\
\midrule

Decentralized & embed
& 8.3 & 3.3 & 35.2 & 18.4 & 25.0 & 97.3
& 7.0 & 3.9 & 31.7 & 18.4 & 25.6 & 97.0 \\

& annotated
& 8.3 & 3.3 & 35.1 & 18.3 & 25.0 & 97.3
& 7.0 & 3.8 & 31.7 & 18.4 & 25.6 & 97.0 \\

\rowcolor{gray!15}
& $\Delta$
& 0.0 & 0.0 & +0.1 & +0.1 & +0.1 & +0.1
& 0.0 & +0.0 & 0.0 & 0.0 & 0.0 & 0.0 \\

\midrule

Centralized & embed
& 7.5 & 3.1 & 34.2 & 18.0 & 23.6 & 95.6
& 7.2 & 3.6 & 31.8 & 18.5 & 25.7 & 95.9 \\

& annotated
& 7.5 & 3.1 & 34.3 & 18.0 & 23.6 & 95.8
& 7.4 & 3.7 & 31.9 & 18.6 & 25.6 & 96.2 \\

\rowcolor{gray!15}
& $\Delta$
& 0.0 & 0.0 & -0.1 & 0.0 & 0.0 & -0.2
& -0.2 & -0.1 & -0.1 & -0.1 & +0.1 & -0.3 \\

\midrule

Centralized Implicit & embed
& 5.4 & 2.3 & 31.5 & 16.6 & 23.1 & 96.6
& 7.4 & 4.0 & 28.9 & 16.7 & 24.3 & 96.7 \\

& annotated
& 5.4 & 2.3 & 31.4 & 16.7 & 23.1 & 96.6
& 7.2 & 3.9 & 28.8 & 16.6 & 24.3 & 96.9 \\

\rowcolor{gray!15}
& $\Delta$
& 0.0 & 0.0 & +0.1 & 0.0 & 0.0 & -0.1
& +0.2 & +0.1 & +0.2 & +0.1 & 0.0 & -0.2 \\

\bottomrule
\end{tabular}
}
\par\smallskip
\begin{minipage}{0.95\linewidth}
\footnotesize
Embed denotes sentence-embedding matching; annotated denotes Qwen3.6 annotation.
\end{minipage}
\label{tab:metric_supp}
\end{table*}

\subsection{Robustness of Instruction-to-Subtask Matching}
\label{app:matchexp}
\Cref{tab:metric_supp} compares the embedding-based matching of \Cref{app:matching} with an LLM-annotated mapping, in which Qwen3.6 reads each emitted atomic instruction together with the ground-truth pool and annotates the subtask it refers to, on the same Qwen3.6 scheduling runs. Mission-level metrics are identical on val-unseen and differ by at most $0.2$ points on test-unseen, and every subtask-level difference lies within $\pm0.3$ points. Since the two matchers rely on entirely different mechanisms, this agreement indicates that the reported metrics do not hinge on the particular matcher, and it justifies the lightweight sentence-embedding matcher as a reproducible default. The largest differences appear in TC under the centralized and centralized-implicit regimes, where any agent may execute any subtask, so that surplus or paraphrased declarations are plausibly harder to attribute.

\section{Evaluation Details}
\label{app:eval}

\begin{figure}[t]
\centering
\subfloat[]{\includegraphics[width=0.32\linewidth]{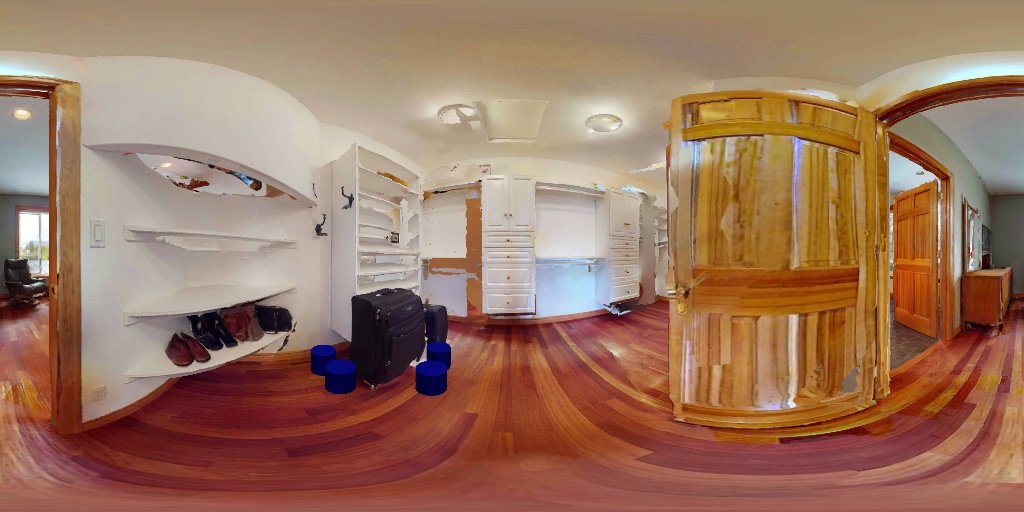}\label{fig:anchor1}}\hfil
\subfloat[]{\includegraphics[width=0.32\linewidth]{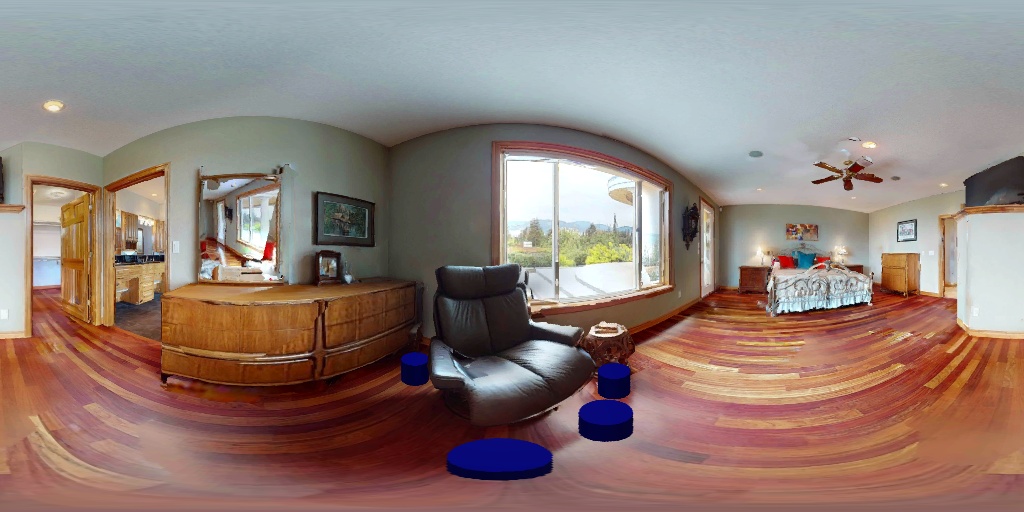}\label{fig:anchor2}}\hfil
\subfloat[]{\includegraphics[width=0.32\linewidth]{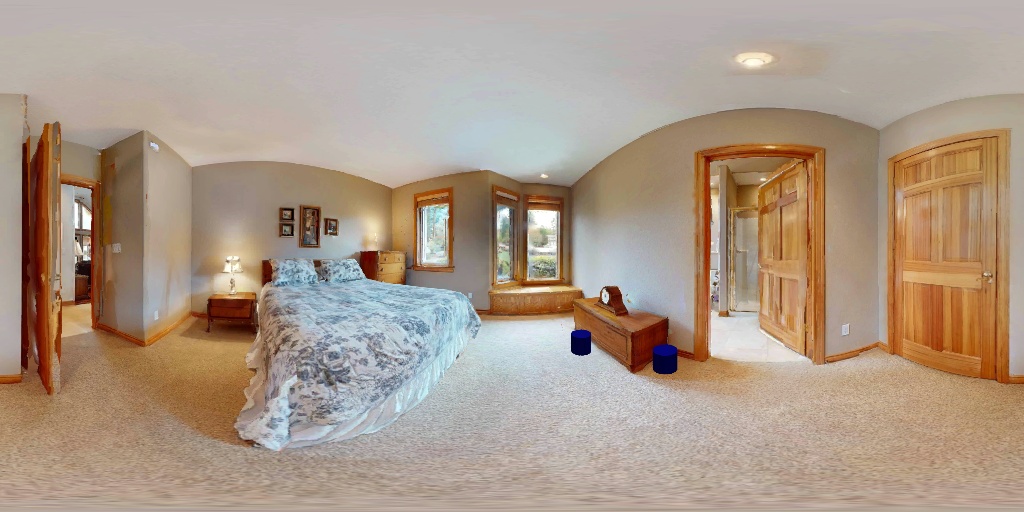}\label{fig:anchor3}}
\caption{Navigable goal anchors rendered as blue cylinders in panoramas captured at goal viewpoints. \textbf{(a)} Anchors around a suitcase in a walk-in closet. \textbf{(b)} Anchors around an armchair in a bedroom. \textbf{(c)} Anchors of a wooden chest in a bedroom, where deduplication and reachability leave two anchors.}
\label{fig:anchors}
\end{figure}

\subsection{Navigable Goal Anchors}
\label{app:anchors}

Measuring arrivals against a single object center systematically mis-scores large furniture: the axis-aligned bounding-box (AABB) center of a bed or cabinet may project to an interior or unreachable point, while an agent legitimately standing near another side of the object would still be penalized. We therefore associate each target instance with a small set of navigable anchors. Specifically, we take the AABB center and its six face centers as candidate points and project each candidate onto the navigation mesh. Using a viewpoint that certifies the instance, we discard projected candidates that are not geodesically reachable from that viewpoint. We then scan the remaining projections in order and remove duplicates, retaining a point only if it is at least $0.5\,\mathrm{m}$ from all previously retained anchors. Instances for which no anchor remains are not grounded at that viewpoint. Let $d(x,o)$ denote the distance from a position $x$ to an instance; and the set of instances reached from $x$ is
\begin{equation}
\small
\begin{aligned}
\mathcal{R}(x) &= \bigl\{\, o \in \{g_1,\dots,g_N\} \;:\; d(x,o) < 3\,\mathrm{m} \,\bigr\}.
\end{aligned}
\label{eq:reached}
\end{equation}
\normalsize
\Cref{fig:anchors} visualizes the anchors of three evaluation targets. The anchors surround each object on its accessible sides and collapse to a few points for compact objects, so that an agent approaching from any accessible direction is credited consistently. The same anchors define the grounding radius at generation time (\Cref{app:stage1}), the geodesic supervision of the oracle components (\Cref{app:oracles}) and the arrival test at evaluation time, so generation and evaluation share one notion of reaching a target.

\subsection{Arrival Declarations and Metric Definitions}
\label{app:metrics}
An agent declares an arrival when its allocated atomic instruction completes (\Cref{app:loop}). The $j$-th declaration of an episode is the tuple $\delta_j=(a_j,t_j,\mathcal{R}_j,\hat{\ell}_j,\kappa_j)$, where $a_j$ is the declaring agent, $t_j$ the synchronized time step of the declaration, $\mathcal{R}_j=\mathcal{R}(x_{a_j}^{t_j})$ the set of instances reached at that moment, $\hat{\ell}_j$ the atomic instruction the agent was executing, and $\kappa_j$ the ground-truth subtask matched to $\hat{\ell}_j$ (\Cref{app:matching}), with $\kappa_j=\varnothing$ when no matching is used. The evaluator of \Cref{app:evaluator} consumes the declaration sequence and returns three subtask sets: the completed subtasks $\mathcal{C}\subseteq\mathcal{T}$, which were consumed by some declaration; the atomically successful subtasks $\mathcal{S}^{\circ}\subseteq\mathcal{C}$, whose target was reached irrespective of constraints; and the successful subtasks $\mathcal{S}\subseteq\mathcal{S}^{\circ}$, whose declaration additionally honors all dependency, holding and locking constraints and whose dependencies are themselves successful. For the declaration $\delta_j$ consumed by $\tau_k$, let $j^-$ be the previous declaration of the same agent (the episode start if none), let $p_k=\sum_{t=t_{j^-}+1}^{t_j}\lVert x_{a_j}^{t}-x_{a_j}^{t-1}\rVert$ be the distance the agent traveled in between, let $\lambda^{\circ}_k=d(x_{a_j}^{t_{j^-}},g_k)$ be the geodesic distance from its position at $t_{j^-}$ to the target, and let $\lambda^{\star}_k$ be the precomputed length of the fragment of $\tau_k$ in the optimal reference execution. With the total traveled distance $P=\sum_{a\in\mathcal{A}}\sum_{t=1}^{T}\lVert x_{a}^{t}-x_{a}^{t-1}\rVert$ over the $T$ synchronized steps of the episode, the metrics are
\begin{equation}
\small
\begin{aligned}
\mathrm{SR} &= \mathbb{I}\bigl[\mathcal{S}=\mathcal{T}\bigr], &
\mathrm{SPL} &= \mathrm{SR}\cdot\frac{\sum_{k}\lambda^{\star}_k}{\max\bigl(P,\sum_{k}\lambda^{\star}_k\bigr)}, \\
\mathrm{CSR} &= \frac{|\mathcal{S}|}{N}, &
\mathrm{CSPL} &= \frac{1}{N}\sum_{k=1}^{N}\mathbb{I}\bigl[\tau_k\in\mathcal{S}\bigr]\frac{\lambda^{\star}_k}{\max(\lambda^{\star}_k,p_k)}, \\
\mathrm{TC} &= \frac{|\mathcal{C}|}{N}, &
\mathrm{ISPL} &= \frac{1}{N}\sum_{k=1}^{N}\mathbb{I}\bigl[\tau_k\in\mathcal{S}^{\circ}\bigr]\frac{\lambda^{\circ}_k}{\max(\lambda^{\circ}_k,p_k)}, \\
\mathrm{TS} &= T, &
\mathrm{MAC} &= \frac{1}{T\,M}\sum_{t=1}^{T}\mathbb{I}\Bigl[\exists\, a\neq a' : \lVert x_{a}^{t}-x_{a'}^{t}\rVert<2\rho\Bigr],
\end{aligned}
\label{eq:metrics}
\end{equation}
\normalsize
where fractions are set to zero when their denominator vanishes, and all metrics are averaged over episodes. ISPL rewards efficient approach to the target from wherever the agent actually was, independent of whether it should have gone there, whereas CSPL compares against the optimal team execution and rewards only constraint-consistent arrivals. TS is the realized makespan, which, unlike a planned makespan, absorbs congestion, waiting and deadlock during execution. MAC counts the synchronized steps with at least one pair of agents closer than $2\rho$ and normalizes by the team size, so that values remain comparable across team sizes. Presence locks are verified on the realized trajectory. If $\tau_k\in\mathcal{S}$ is locked and was consumed by $\delta_j$, the mission fails if the executing agent leaves the success radius of any instance credited at $\delta_j$ before the release.

\begin{algorithm}[t]
\SetEndCharOfAlgoLine{}
\SetKwComment{Comment}{// }{}
\SetKw{KwAnd}{and}
\SetKw{KwOr}{or}
\SetKwInOut{Input}{Input}
\SetKwInOut{Output}{Output}
\SetKwFunction{FCommit}{Commit}
\SetKwProg{Fn}{Function}{:}{}
\Input{\\\hspace{-3.6em}\small
\begin{tabular}[t]{l @{\hspace{.3em}} l}%
$\mathcal{T}$ & Subtask DAG with attributes $(g_k,A_k,D_k,h_k,l_k,U_k)$ \\
$\{\delta_j\}_{j=1}^{J}$ & Declarations $(a_j,t_j,\mathcal{R}_j,\hat{\ell}_j,\kappa_j)$ \\
\end{tabular}%
}
\BlankLine
$\mathcal{H}\leftarrow\{h_0\}$ with all subtasks incomplete and $\mathrm{last}(a)\leftarrow\varnothing$ for all $a$ \\
\For{$j=1$ \KwTo $J$}{
  $\mathcal{H}'\leftarrow\emptyset$ \\
  \For{hypothesis $h\in\mathcal{H}$}{
    $\Phi\leftarrow\{\tau_k \text{ incomplete in } h : D_k \text{ completed in } h,\; a_j\in A_k,\; \kappa_j\in\{\varnothing,k\}\}$ \Comment{ready subtasks}
    $\mathcal{R}'\leftarrow\{g_{\kappa_j}\}$ \textbf{if} $\kappa_j\neq\varnothing$ \KwAnd $g_{\kappa_j}\in\mathcal{R}_j$; $\{\varnothing\}$ \textbf{if} $\kappa_j\neq\varnothing$ otherwise; $\mathcal{R}_j$ (or $\{\varnothing\}$ if empty) \textbf{if} $\kappa_j=\varnothing$ \\
    \For{target $r\in\mathcal{R}'$}{
      $\Phi_r\leftarrow\{\tau_k\in\Phi : g_k=r\}$ \\
      \uIf{$\Phi_r\neq\emptyset$}{\lFor{$\tau_k\in\Phi_r$}{$\mathcal{H}'\leftarrow\mathcal{H}'\cup\{\FCommit(h,j,\tau_k,\mathrm{true})\}$}}
      \uElseIf{$\Phi\neq\emptyset$}{\lFor{$\tau_k\in\Phi$}{$\mathcal{H}'\leftarrow\mathcal{H}'\cup\{\FCommit(h,j,\tau_k,\mathrm{false})\}$}}
      \lElse{$\mathcal{H}'\leftarrow\mathcal{H}'\cup\{\FCommit(h,j,\varnothing,\mathrm{false})\}$}
    }
  }
  $\mathcal{H}\leftarrow\mathcal{H}'$ \\
}
\Output{Hypothesis selected lexicographically, its sets $\mathcal{C},\mathcal{S}^{\circ},\mathcal{S}$ after failure propagation, and the next oracle assignment}
\BlankLine
\Fn{\FCommit{$h,j,\tau,\mathit{correct}$}}{
  $h'\leftarrow$ copy of $h$; $\tau^{-}\leftarrow\mathrm{last}(a_j)$ in $h'$ \\
  \lIf{$\tau^{-}\neq\varnothing$ \KwAnd $\tau^{-}$ is still locked}{mark $\tau^{-}$ failed, and mark $\tau$ failed as well if $\tau\neq\varnothing$ and $\mathit{correct}$}
  \lIf{$\tau=\varnothing$}{\Return $h'$}
  mark $\tau$ completed; mark $\tau$ atomically successful and successful \textbf{if} $\mathit{correct}$, failed otherwise \\
  \If{$\mathit{correct}$}{
    \lIf{$\tau^{-}$ is holding \KwAnd ($\tau^{-}\notin D_{\tau}$ \KwOr the drafted executors of $\tau^{-}$ and $\tau$ differ)}{mark $\tau$ failed}
    \lIf{some $\tau_d\in D_{\tau}$ is incomplete or failed}{mark $\tau$ failed}
    \lFor{$u\in U_{\tau}$ that is the last subtask of another agent}{unlock $u$}
  }
  $\mathrm{last}(a_j)\leftarrow\tau$; \Return $h'$ \\
}
\caption{Constraint-Aware Discrete-Event Evaluation}
\label{alg:evaluator}
\end{algorithm}

\subsection{Discrete-Event Evaluator}
\label{app:evaluator}
\Cref{alg:evaluator} evaluates a declaration sequence under the task semantics. Because several instances may lie within the success radius and a declaration may be attributable to several ready subtasks, the evaluator maintains a set of hypotheses, each holding a copy of the DAG state and the last subtask of every agent. For every declaration it branches over all consistent interpretations: a ready subtask of the declaring agent whose target was reached is consumed as a correct arrival; otherwise, any ready subtask of that agent may be consumed as an incorrect arrival, which counts toward completion but not toward success; if the agent has no ready subtask, the declaration consumes nothing but still breaks a presence lock held by that agent. When a subtask identifier $\kappa_j$ is available from matching, the branching collapses to that subtask, which is consumed correctly if and only if its own target is in $\mathcal{R}_j$. Failure propagates upward: a subtask fails whenever any of its dependencies fails, and the mission succeeds only if every subtask succeeds.

Among the final hypotheses, the evaluator retains those whose pending assignments are consistent with the task state and that admit a complete future execution, and selects the one that lexicographically maximizes TC, then maximizes CSR, then minimizes the number of future rounds, with the order of consumed subtasks as the final tie-breaker; when no hypothesis admits a complete future, the same key without the future rounds is applied to all hypotheses. The future execution is computed by a branch-and-bound search over round-wise assignments with pre-allocation (\Cref{app:oracles}), which also yields the next assignment of the oracle scheduler. Without matching, the evaluator thus resolves ambiguity in favor of the team, which corresponds to the greedy strategy that maximizes task completion and breaks ties by conditional success.

\subsection{Instruction-to-Subtask Matching}
\label{app:matching}
Since the scheduler emits atomic instructions that it extracted itself, a declaration does not identify which ground-truth subtask the agent attempted. We recover this mapping by text similarity. We use all-MiniLM-L6-v2 to obtain sentence embeddings of a case-folded, whitespace-normalized text. Let $\{\hat{\ell}_j\}_{j=1}^{J}$ be the instructions of the declarations of an episode, and let $\{\ell^{\star}_k\}_{k=1}^{N}$ be the ground-truth atomic instructions. Every declaration is a separate row, even when two declarations carry identical text. The mapping solves the rectangular assignment problem over one-to-one partial assignments of maximum cardinality, solved by the Hungarian algorithm. Each matched declaration receives $\kappa_j=k$; unmatched surplus declarations are recorded for auditing and removed from the declaration sequence before the metrics are recomputed. The mapping prevents an evaluator from crediting an arrival to whichever subtask happens to be satisfied, and it is robust to the choice of matcher (\Cref{tab:metric_supp}). It presupposes that the evaluated system exposes a pool of atomic instructions.

\subsection{Oracle Components}
\label{app:oracles}
\noindent\textbf{Oracle scheduler.} The oracle scheduler reveals the ground-truth atomic instructions together with the next round of the evaluator's optimal future execution. This execution is found by a branch-and-bound search over rounds: in each round, pending subtasks whose dependencies are complete fire first in subtask order; each free agent (neither locked nor pending) receives at most one undispatched subtask it is permitted to execute, where a holding agent may only receive the consumer of its held subtask with the same drafted executor, and a subtask consuming a held object is reserved for the holder once the object has been acquired; assigned subtasks with complete dependencies fire at the end of the round, whereas the others are pre-allocated and become pending; and at least one assigned subtask must be dependency-ready. Assignments are expanded in order of decreasing size and decreasing number of ready subtasks, and a branch is pruned when the rounds used cannot improve on the best complete plan. The legacy variant (LE) restricts dispatch to dependency-ready subtasks and thus disables pre-allocation.

\noindent\textbf{Oracle navigator.} The oracle navigator replaces planning and control by a geodesic follower that drives the agent toward the goal position of its allocated subtask along the navigation-mesh shortest path, moving $0.2$\,m per step when aligned with the next path point within $0.1$\,rad and turning by $10^{\circ}$ otherwise. It terminates when the agent is within $0.2$\,m of the goal on the horizontal plane, after $500$ steps, or immediately if the goal is unreachable.

\noindent\textbf{Oracle planner.} The oracle planner keeps the shared topology, conflict-based path finding and control of TRISS but replaces the policy decision, and hence the target allocation, by the expert action used for imitation learning (\Cref{app:training}): it stops when the geodesic distance to the goal of the allocated subtask is below $1.5$\,m, and otherwise selects the selectable candidate whose simulated position is geodesically closest to that goal.

\section{Method Details}
\label{app:method}

\begin{algorithm}[t]
\SetEndCharOfAlgoLine{}
\SetKwComment{Comment}{// }{}
\SetKw{KwAnd}{and}
\SetKw{KwOr}{or}
\SetKwInOut{Input}{Input}
\SetKwInOut{Output}{Output}
\Input{\\\hspace{-3.6em}\small
\begin{tabular}[t]{l @{\hspace{.3em}} l}%
$\mathcal{L}$ & Task instructions \\
$\mathcal{A}$ & Team of $M$ agents \\
\end{tabular}\hspace{-0.5em}%
\begin{tabular}[t]{l @{\hspace{.3em}} l}%
$T_{\max}$ & Planning step limit \\
$\pi_{\theta}$ & Trained planner \\
\end{tabular}%
}
\BlankLine
$\mathcal{I}\leftarrow\mathrm{LLM}_{dec}(\mathcal{L})$; $\mathcal{G}\leftarrow\emptyset$; $P_0\leftarrow\emptyset$ \Comment{atomic pool, shared topology, pending set}
\For{round $n=1,2,\dots$}{
  Reset visitation records and stop scores; localize all agents in $\mathcal{G}$ (\Cref{alg:topology}) \\
  $(I_n,P_n,F_n)\leftarrow$ \textsc{Schedule}$(\mathcal{L},\mathcal{I},n)$ \\
  \lFor{agent $a\in F_n$ in order}{declare the pending arrival of $a$; update the task state}
  \lIf{$I_n^{(a)}=\varnothing$ for all $a$ \KwAnd $F_n=\emptyset$}{\textbf{break}}
  Encode $\hat{\ell}^{(a)}$ for every agent with $I_n^{(a)}\neq\varnothing$; mark these agents active \\
  \For{step $t=0,\dots,T_{\max}-1$ \textbf{while} some agent is active}{
    \For{every active agent $a$}{
      Localize $a$ and update $\mathcal{G}$ from its panorama (\Cref{alg:topology}) \\
      encode its panorama, update node features, and compute fused logits $s^{(a)}$ over $\hat{\mathcal{V}}^{(a)}_t$
    }
    Assign targets to active agents by maximum-score bipartite matching over $\{s^{(a)}\}$ (\Cref{eq:cfta}) \\
    \For{every active agent $a$}{
      \uIf{$a$ selects stop \KwOr $t=T_{\max}-1$ \KwOr no candidate is selectable}{goal $\leftarrow$ node with the highest stop score of $a$ in this round; mark $a$ idle}
      \lElse{goal $\leftarrow$ selected node; remove it from $\mathcal{G}$ if unexplored and record the attempt}
    }
    Idle agents hold their current node; routes $\leftarrow$ \textsc{CBS}$(\mathcal{G},\text{starts},\text{goals})$ (\Cref{alg:cbs}) \\
    Execute all routes synchronously with the low-level controller (\Cref{app:controller}); collect verified anchors \\
  }
  \lFor{agent $a$ with $I_n^{(a)}\neq\varnothing$ \KwAnd $a\notin P_n$}{declare the arrival of $a$; update the task state}
}
\Output{Declaration sequence evaluated by \Cref{alg:evaluator}}
\caption{Episode Loop of TRISS}
\label{alg:triss}
\end{algorithm}

\subsection{Episode Loop of TRISS}
\label{app:loop}
\Cref{alg:triss} combines the three modules of TRISS into one episode. An episode alternates scheduling rounds and navigation rounds on the synchronized platform. At the beginning of round $n$, every agent is localized in the shared topology, the scheduler emits the assignment $I_n$, the newly pre-allocated agents $P_n$ and the ordered firing list $F_n$, and the previously pending actions listed in $F_n$ fire one by one, each as a separate declaration so that a release caused by an earlier firing is visible to later ones. The agents holding a non-blank instruction then navigate for at most $T_{\max}$ planning steps. At every step, the active agents update and read the shared topology, the planner scores their candidate nodes, the conflict-free target allocation assigns distinct targets, the conflict-based path finder converts the targets into synchronized routes over the shared topology, and the low-level controllers execute these routes. An agent becomes idle once it stops, reaches the step limit, or has no selectable candidate left. When all agents are idle, every allocated agent that is not pending declares its arrival, the evaluator updates the task state, and the next round begins. The topology persists across rounds, whereas per-agent visitation records and stop scores are reset at every round. The episode terminates when a round yields a blank assignment for every agent without firing any pending action.

\subsection{LLM-based Subtask Scheduler}
\label{app:scheduler}
\noindent\textbf{Atomic instruction extraction.} At the start of an episode, the extractor $\mathrm{LLM}$ decomposes $\mathcal{L}$ into the shared pool $\mathcal{I}=\{\ell_i\}_{i=1}^{L}$, addressed by integer indices. Its prompt defines a navigation destination as the specific target the agent interacts with together with everything that identifies it (its room, appearance, placement, surroundings and markers), so that a sentence introducing a room and a subsequent sentence describing a target within that room form a single destination. The extractor must preserve all identifying detail and the original wording, and may only remove on-arrival actions and coordination or dependency clauses. The call is repeated until it returns a non-empty list of strings.

\noindent\textbf{Scheduling protocol.} At each round, the scheduler receives the instruction(s) $\mathcal{L}$, the pool $\mathcal{I}$, the per-round history of issued indices, the per-round history of the pending set, its own previous scheduling plans, and the list of agent identifiers, and returns an updated scheduling plan describing future rounds only, the per-agent indices $I_n$, the ordered firing list $F_n$ and the newly pre-allocated agents $P_n$.

\noindent\textbf{Observation-guided variant.} The observation-guided scheduler additionally receives a situation report
\begin{equation}
\small
\begin{aligned}
S_n &= \bigl\{\bigl(a,\; v_n^{(a)},\; \mathrm{desc}(v_n^{(a)})\bigr)\bigr\}_{a\in\mathcal{A}}
\end{aligned}
\label{eq:situation}
\end{equation}
\normalsize
that pairs each agent's current viewpoint $v_n^{(a)}$ with a description of the place, and its prompt instructs the scheduler to use this report as the primary evidence for allocation whenever the instruction does not bind a subtask to an agent. The description is generated once per viewpoint at the start of a round by the same VLM from the views in the directions of the current waypoint proposals, their headings relative to the agent, and RAM++ landmark tags. Its prompt requires an orientation-agnostic description that anchors objects to architectural elements and to each other rather than to the viewer, emphasizes cues that distinguish the place from nearby alternatives, and, given the atomic pool as background, notes which targets are plausibly correlated with the current location while marking such inferences as uncertain.

\noindent\textbf{Round execution.} The environment realizes the protocol as follows. At the beginning of round $n$, each agent in $F_n$ fires its previously pending subtask as an individual declaration in the listed order, and is removed from the pending set only at its own firing, so that later firings in the same list are evaluated after earlier releases. The new pending agents $P_n$ are registered only after all firings of the round, and the instructions of $I_n$ are installed afterwards. At the end of the round, allocated agents outside the pending set declare their arrival, whereas pending agents remain at their destinations until they appear in a later firing list. Under the oracle scheduler, the same procedure is driven by the evaluator's optimal future execution (\Cref{app:oracles}) and additionally carries ground-truth subtask identifiers, so that each declaration is evaluated against its intended subtask.

\subsection{Shared Topology Maintenance}
\label{app:topology}
\noindent\textbf{Representation.} The shared topology $\mathcal{G}_t=(\mathcal{V}_t,\mathcal{E}_t)$ stores, for every explored node $v\in\mathcal{V}^e_t$, a measured position $\hat{p}_v$, and for every unexplored node $v\in\mathcal{V}^u_t$ the list of position estimates at which it has been proposed together with their mean $\hat{p}_v$. The explored set is shared by the team, whereas each agent $a$ additionally holds the set $\mathcal{V}^{(a)}_t\subseteq\mathcal{V}^e_t$ of nodes it has visited in the current round. Edges carry metric lengths, and the memory maintains the all-pairs shortest distance matrix $E$ together with a predecessor table from which shortest node sequences are reconstructed; $E$ is consumed as the distance bias of the coarse-scale encoder, in the positional features of the planner, and as transition cost and heuristic of the space-time search. The memory further stores node features (\Cref{app:planner}), per-agent stop scores for the current round, and the collective ban list $B$.

\noindent\textbf{Localization and node identity.} Two radii govern node identity. The identification radius $\varepsilon_{\text{loc}}=0.5$\,m decides whether a position is an existing node: before viewpoint prediction, the measured position $q_t$ of an agent is localized by \Cref{eq:localize_supp} restricted to explored nodes, and if the nearest one lies within $\varepsilon_{\text{loc}}$ it becomes the current viewpoint $v_t$. Otherwise, a new explored node is created at $\hat{p}_{v_t}=q_t$. Waypoint proposals are localized against all nodes except $v_t$, using the mean estimate for unexplored nodes, so that proposals from different agents and vantage points merge instead of forking,
\begin{equation}
\small
\begin{aligned}
v^{*}(q) &= \argmin_{v\in\mathcal{V}_t} \lVert \hat{p}_v - q \rVert, \qquad q \text{ is identified with } v^{*}(q) \iff \lVert \hat{p}_{v^{*}(q)} - q \rVert \le \varepsilon_{\text{loc}}.
\end{aligned}
\label{eq:localize_supp}
\end{equation}
\normalsize
The neighbor radius $\varepsilon_{\text{nbr}}=1.5$\,m governs stitching at the beginning of a round: an agent whose start position localizes to no explored node is instantiated as a new node and connected by metric edges to all explored nodes within $\varepsilon_{\text{nbr}}$, so that a freshly placed agent inherits the connectivity of its teammates rather than starting an isolated component. Within a round, a newly created node is connected only to the verified anchors returned by the controller (traversal-verified connection); a localized node is connected to these anchors if it differs from them. The localized $v_t$ is also connected to its predecessor.

\noindent\textbf{Unexplored-node lifecycle.} A proposal $y_i$ that matches an unexplored node appends its estimate to that node, updates the mean, and inherits no new edge from the matching itself. If the node is already connected, the geodesics are relaxed through each previously connected node; a proposal matching an explored node yields a direct edge to it. A proposal that matches nothing creates a new unexplored node, and every matched or created node is connected to $v_t$. After the update, unexplored neighbors of $v_t$ that were not proposed by any agent in the current step are pruned, so that stale frontiers do not accumulate. When an agent selects an unexplored node as its target, the node is removed from the memory at dispatch; the position the agent actually reaches is inserted as an explored node at the next localization, and its identity is decided by \Cref{eq:localize_supp}. Hence explored positions are always measured, never extrapolated from proposals.

\begin{algorithm}[t]
\SetEndCharOfAlgoLine{}
\SetKwComment{Comment}{// }{}
\SetKw{KwAnd}{and}
\SetKw{KwOr}{or}
\SetKwInOut{Input}{Input}
\SetKwInOut{Output}{Output}
\Input{\\\hspace{-3.6em}\small
\begin{tabular}[t]{l @{\hspace{.3em}} l}%
$\mathcal{G}_t$ & Shared topology with $E$, $B$ \\
$q_t,\phi_t$ & Measured pose of agent $a$ \\
$Y_t$ & Viewpoint proposals \\
\end{tabular}\hspace{-0.5em}%
\begin{tabular}[t]{l @{\hspace{.3em}} l}%
$\Lambda$ & Verified anchors from the controller \\
$\mathit{att}$ & Pending attempt $(u,y)$ of $a$ \\
$\varepsilon_{\text{loc}},\varepsilon_{\text{nbr}},\varepsilon_{\text{ban}}$ & Radii \\
\end{tabular}%
}
\BlankLine
\Comment{Localization}
$v^{\mathrm{e}}\leftarrow\argmin_{v\in\mathcal{V}^e_t}\lVert\hat{p}_v-q_t\rVert$ \\
\uIf{$\lVert\hat{p}_{v^{\mathrm{e}}}-q_t\rVert\le\varepsilon_{\text{loc}}$}{
  $v_t\leftarrow v^{\mathrm{e}}$; \lIf{$\mathit{att}\neq\varnothing$}{$B\leftarrow B\cup\{(u,y)\}$} \Comment{failure-verified}
  connect $v_t$ to every $\lambda\in\Lambda\setminus\{v_t\}$ \\
}
\Else{
  create explored node $v_t$ with $\hat{p}_{v_t}=q_t$ \\
  \lIf{first step of the round}{connect $v_t$ to all $v\in\mathcal{V}^e_t$ with $\lVert\hat{p}_v-q_t\rVert<\varepsilon_{\text{nbr}}$}
  \lElse{connect $v_t$ to every $\lambda\in\Lambda$} \Comment{traversal-verified}
}
$\mathit{att}\leftarrow\varnothing$ \\
\Comment{Proposal filtering and identification}
\lIf{$v_t$ has ban entries}{remove every $y_i$ violating \Cref{eq:fve_supp} from $Y_t$}
\For{$y_i\in Y_t$}{
  $v^{*}\leftarrow\argmin_{v\in\mathcal{V}_t\setminus\{v_t\}}\lVert\hat{p}_v-y_i\rVert$ \\
  \uIf{$\lVert\hat{p}_{v^{*}}-y_i\rVert\le\varepsilon_{\text{loc}}$}{
    \lIf{$v^{*}\in\mathcal{V}^u_t$}{append $y_i$ to the estimates of $v^{*}$ and update $\hat{p}_{v^{*}}$}
    $\mathcal{V}'_t\leftarrow\mathcal{V}'_t\cup\{v^{*}\}$ \\
  }
  \lElse{create unexplored node $v$ with $\hat{p}_v=y_i$; $Y'_t\leftarrow Y'_t\cup\{v\}$}
}
\For{$v\in\mathcal{V}'_t\cup Y'_t$}{
  \uIf{$v\in\mathcal{V}^e_t$}{add edge $(v_t,v)$ if absent}
  \Else{
    \lIf{$v$ was connected}{re-insert its edges with lengths to $\hat{p}_v$ and relax through each neighbor}
    add edge $(v_t,v)$ \\
  }
}
Add $v_t$ to $\mathcal{V}^{(a)}_t$ and $\mathcal{V}^e_{t}$ \\
Prune unexplored neighbors of $v_t$ not proposed in the current step \\
\Output{Updated $\mathcal{G}_{t+1}$, current node $v_t$, candidate neighbors $\mathcal{V}'_t\cup Y'_t$}
\caption{Shared Topology Update for One Agent}
\label{alg:topology}
\end{algorithm}

\subsection{Topological Memory-based Navigation Planner}
\label{app:planner}
\noindent\textbf{Text encoder.} An atomic instruction $\ell$ is tokenized and embedded by a stack of transformer layers into $\hat{\ell}\in\mathbb{R}^{|\ell|\times D}$ with hidden size $D=768$.

\noindent\textbf{Panorama encoder.} At viewpoint $v_t$, each of the twelve views is represented by its RGB feature $r_t^{(i)}$, depth feature $d_t^{(i)}$, orientation feature $o^{(i)}=[\sin\vartheta_i,\cos\vartheta_i,\sin\varphi_i,\cos\varphi_i]$ and a binary type $n^{(i)}$ indicating whether the view carries a waypoint proposal. Views carrying proposals are listed first, one entry per proposal, followed by the remaining views. The panorama encoder yields the contextual panoramic embeddings $\hat{r}_t=\{\hat{r}_t^{(i)}\}$. The representation of $v_t$ is overwritten by the average of $\hat{r}_t$ over valid views, whereas an unexplored node proposed through view $i$ accumulates $\hat{r}_t^{(i)}$ and is represented by the mean of its accumulated embeddings. A node that has become explored no longer accumulates proposal embeddings.

\noindent\textbf{Coarse-scale encoder.} The candidate set $\hat{\mathcal{V}}_t^{(a)}=\mathcal{V}_t^e\cup\mathcal{N}\!\left(\mathcal{V}_t^{(a)}\right)$ is realized within the connected component of $v_t$ as the union of three groups: the nodes $\mathcal{V}_t^{(a)}$ visited by the agent in the current round, the other explored nodes, and the unexplored nodes directly adjacent to $\mathcal{V}_t^{(a)}$. The first group provides context but is masked from selection. The coarse-scale encoder augments the candidate node features $x_i$ for $v_i \in \hat{\mathcal{V}}_t^{(a)}$ with positional encodings and a learned stop token $x_0$, and processes $X = [x_0, x_1, \ldots, x_{|\hat{\mathcal{V}}_t^{(a)}|}]$ together with $\hat{\ell}$ through cross-modal layers. Each layer first attends from $X$ to $\hat{\ell}$, then applies Graph-Aware Self-Attention (GASA),
\begin{equation}
\small
\begin{aligned}
\mathrm{GASA}(X) &= \mathrm{Softmax}\Bigl(\frac{XW_q(XW_k)^{\top}}{\sqrt{d}} + Z\Bigr)XW_v, \qquad Z = E\,w_e + b_e,
\end{aligned}
\label{eq:gasa}
\end{equation}
\normalsize
and finally a feed-forward block, where the distance bias $Z$ applies a learned scalar affine map to the pairwise shortest distances $E$ restricted to the candidate set (with zero distance for the stop token) and is shared across attention heads. Because $E$ is computed on the shared topology, the bias spans the union of the team's exploration. Global action logits are computed as $s_i^{(c)} = \mathrm{FFN}_c(\hat{x}_i)$, where $\hat{x}_i \in \hat{X}$ denotes the output of the coarse-scale encoder, providing node-level navigation preferences.

\noindent\textbf{Fine-scale encoder.} The fine-scale encoder processes $R=[r_0;\hat{r}_t]$, where the stop token $r_0$ is a zero vector, with fourteen-dimensional positional features that concatenate the positional feature of the agent's start node of the round, outputting contextual panoramic embeddings $\hat{r}_i$. Local action logits are computed as $s_i^{(f)} = \mathrm{FFN}_f(\hat{r}_i)$. Local logits are converted to the global action space by \Cref{eq:convert_supp}, where $s_{\text{back}}$ sums the local logits of the candidate views whose proposals localize to nodes already visited by the agent,
\begin{equation}
\small
\begin{aligned}
s_i^{(f')} &= \begin{cases} s_i^{(f)}, & \text{if } v_i \in Y_t' \cup \mathcal{V}_t', \\ s_{\text{back}}, & \text{otherwise}. \end{cases}
\end{aligned}
\label{eq:convert_supp}
\end{equation}
\normalsize

\noindent\textbf{Dynamic fusion and stop selection.} The scalar gate is $\sigma_t=\mathrm{Sigmoid}\bigl(\mathrm{FFN}_{\sigma}([\hat{r}_0;\hat{x}_0])\bigr)$, and the fused logit of every candidate, including the stop action $i=0$, is $s_i = \sigma_t\, s_i^{(f')} + (1-\sigma_t)\, s_i^{(c)}$. The stop probability $\mathrm{softmax}(s)_0$ is recorded as the stop score of $v_t$ for the agent. When the agent stops, it does not necessarily stop at $v_t$: following DUET, it travels to the node with the highest stop score among those it has scored in the current round.

\noindent\textbf{VLM-assisted variants.} \label{app:hybrid} The zero-shot VLM planner and the hybrid planner share the shared topology and conflict-aware execution. For the zero-shot planner, the VLM receives the atomic instruction, the per-round instruction history of all agents, the targets reached in previous rounds, the agent's own step history and planning notes, the images of the candidate views, textual action options and the list of admissible action strings, and returns a thought, a one-sentence plan and exactly one admissible action. Action options are verbalized as ``go forward'', ``turn slight left/right'', ``turn sharp left/right'' or ``turn around'' toward a viewpoint according to its relative heading (sectors bounded at $\pm30^{\circ}$, $\pm90^{\circ}$ and $\pm150^{\circ}$), or as ``backtrack to viewpoint'' for nodes already visited by the agent; nodes outside the current neighborhood are offered as backtracking options only if they are on the trained planner's shortlist (defined below) or were final targets of previous rounds. Each option is enriched with the RAM++ landmark tags observed toward the corresponding node, and the zero-shot planner stops at its current node. The hybrid planner queries the VLM only when the trained planner is uncertain. Let $p_{(1)}\ge p_{(2)}\ge\cdots$ be the softmax probabilities of the fused logits over selectable candidates and the stop action. The expert is considered confused if $p_{(1)}-p_{(2)}\le\kappa$, and it recommends the ranked shortlist $\{c : p_c\ge\gamma\,p_{(1)}\}$ truncated to $n_{\max}$ entries, which is provided to the VLM as a strong prior. Confident agents keep the conflict-free allocation of \Cref{eq:cfta}, whereas confused agents query the VLM sequentially, with the targets already claimed by idle agents, by confident agents and by previously queried agents removed from their options; stopping is admissible after the first step if the current node is unclaimed, or whenever no option remains. We use $\kappa=0.03$, $\gamma=0.5$ and $n_{\max}=5$.

\subsection{Conflict-Aware Execution}
\label{app:execution}
\noindent\textbf{Conflict-free target allocation.} Within an episode, let $\mathcal{A}^{\mathrm{act}}$ be the active agents, let $\mathcal{V}_I$ be the current nodes of idle agents, and let $\mathcal{W}=\bigcup_{a\in\mathcal{A}^{\mathrm{act}}}\{v\in\hat{\mathcal{V}}^{(a)}_t\setminus\mathcal{V}^{(a)}_t : v\notin\mathcal{V}_I\}$ be the pool of selectable nodes. Each agent additionally owns a private stop column $\varsigma_a$. Allocation solves the maximum-score bipartite assignment
\begin{equation}
\small
\begin{aligned}
\max_{z\in\{0,1\}^{\mathcal{A}^{\mathrm{act}}\times(\mathcal{W}\cup\{\varsigma_a\})}} \sum_{a,c} z_{ac}\,S_{ac}
\quad \text{s.t.}\quad \sum_{c} z_{ac}=1\;\;\forall a,\qquad \sum_{a} z_{ac}\le 1\;\;\forall c,
\end{aligned}
\label{eq:cfta}
\end{equation}
\normalsize
where $S_{av}=s^{(a)}_v$ for selectable nodes of agent $a$, $S_{a\varsigma_a}=s^{(a)}_0$. The problem is solved by the Hungarian algorithm; the private stop columns guarantee feasibility, and an agent assigned to its stop column executes the stop action. Every agent is thus dispatched to a distinct node, and an agent whose preferred frontier is claimed by a higher-scoring teammate is rerouted to its best remaining alternative within the same solve.

\begin{algorithm}[t]
\SetEndCharOfAlgoLine{}
\SetKwComment{Comment}{// }{}
\SetKw{KwAnd}{and}
\SetKw{KwOr}{or}
\SetKwInOut{Input}{Input}
\SetKwInOut{Output}{Output}
\Input{\\\hspace{-3.6em}\small
\begin{tabular}[t]{l @{\hspace{.3em}} l}%
$\mathcal{G}$ & Shared topology with $E$ \\
$s_{1:M}$ & Start nodes \\
$g_{1:M}$ & Allocated goal nodes \\
\end{tabular}\hspace{-0.5em}%
\begin{tabular}[t]{l @{\hspace{.3em}} l}%
$\delta$ & Collision distance \\
$B_{\text{cbs}}$ & Expansion budget \\
\end{tabular}%
}
\BlankLine
\Comment{Reachability and horizon}
\lIf{some $g_i$ is unreachable from $s_i$ by admissible moves}{\Return failure}
Compute $T_h$ by \Cref{eq:horizon} \\
\Comment{Root: independent space-time plans}
\For{agent $i = 1$ \KwTo $M$}{
  $\pi_i \leftarrow$ SpaceTimeAStar$(s_i, g_i, T_h, \emptyset)$; \lIf{$\pi_i$ does not exist}{\Return failure}
}
$R \leftarrow (\{\pi_i\}, \{\mathcal{C}_i=\emptyset\})$; push $R$ keyed by $(\#\text{conflicts}, \text{cost})$; $h^\star \leftarrow R$; $n_{\exp}\leftarrow0$ \\
\While{open list nonempty}{
  Pop best node $h$; \lIf{$h$ has fewer conflicts than $h^\star$, or as many at lower cost}{$h^\star\leftarrow h$}
  $n_{\exp}\leftarrow n_{\exp}+1$; \lIf{$n_{\exp}>B_{\text{cbs}}$}{\textbf{break}}
  $c \leftarrow$ earliest conflict in $h$: vertex or proximity ($<\delta$) at $t\ge1$, then swap or segment ($<\delta$) on $[t,t+1]$ \\
  \lIf{$c = \varnothing$}{\Return routes of $h$, which are conflict-free}
  \For{agent $i$ involved in $c$}{
    $h' \leftarrow h$; add to $\mathcal{C}_i$ the vertex constraint $(t,\pi_i(t))$ or the edge constraint $(t,\pi_i(t),\pi_i(t+1))$ \\
    Replan with the updated constraints; \lIf{feasible}{count conflicts and cost of $h'$; update $h^\star$; push $h'$}
  }
}
\Output{Routes of $h^\star$ with trailing waits trimmed, conflict-free if the search closed}
\caption{Conflict-Based Path Finding on the Shared Topology}
\label{alg:cbs}
\end{algorithm}

\noindent\textbf{Conflict-based path finding.} \label{app:cbs} Given the start node $s_i$ and goal node $g_i$ of every agent (idle agents use $s_i=g_i$), the planner first verifies by breadth-first search over admissible moves that every goal is reachable and derives the horizon
\begin{equation}
\small
\begin{aligned}
T_{h} &= \max\Bigl(5,\; \max_{i:\,s_i\neq g_i}\bigl(\mathrm{hops}_{\mathrm{BFS}}(s_i,g_i)+1\bigr) + M\Bigr),
\end{aligned}
\label{eq:horizon}
\end{equation}
\normalsize
which reserves one additional waiting slot per agent for conflict resolution. The low-level space-time A* plans each agent a sequence of exactly $T_h$ states starting at $t=0$. A move follows a direct edge to an explored node, or to the agent's own goal; an unexplored goal can only be entered at the final step, and no move departs from an unexplored node. Waiting is always admissible. The transition cost is the metric length of the move (zero for waiting), the heuristic is the stored shortest distance to the goal (zero if unknown), and a state is terminal only if $t\ge T_h-1$ and the agent occupies its goal, so an agent may temporarily vacate its goal to let a teammate pass and return before the horizon. Paths shorter than $T_h$ are padded by waiting at the goal. The high-level search (\Cref{alg:cbs}) scans the synchronized timeline for the earliest conflict. At every step $t\ge1$ it checks vertex conflicts, where two agents occupy the same node, and proximity conflicts, where their nodes are closer than the collision distance $\delta$. For every transition from $t$ to $t+1$ it checks swap conflicts and segment conflicts, where the minimum distance between the two agents over eleven synchronized interpolation points is below $\delta$; conflicts at $t=0$ between agents that localized to the same node are ignored, since several agents may legitimately depart from one abstract node. A vertex or proximity conflict at time $t$ adds the vertex constraint $(t,\text{own node})$ to each of the two agents in two child nodes, and a segment or swap conflict adds the edge constraint $(t,\text{own move})$. Children whose replanning succeeds are pushed into an open list ordered by the number of conflicts and then by total metric cost. The search returns the first conflict-free node; after $B_{\text{cbs}}$ expansions, it returns the incumbent with the fewest conflicts and, among those, the lowest cost, trading completeness for real-time operation. Returned paths are trimmed of trailing waits; all intermediate nodes are explored, and the goal is addressed by its measured position if explored and by its mean estimate otherwise. For agents moving to an unexplored target, the last node before the target on the returned route replaces the planned front node when recording the attempt.

\noindent\textbf{Failure-verified exclusion.} When an agent is dispatched toward an unexplored target, the memory records the attempt as the pair of the last explored node $u$ on its route and the target position $y$ (\Cref{alg:topology}). At the next localization, if the agent's measured position is identified with an existing explored node instead of instantiating a new one, the navigation did not reach new ground, and $(u,y)$ is appended to the collective ban list $B$; otherwise the attempt is cleared. Before the proposals $Y_t$ generated at $v_t$ enter the memory, every proposal lying within the ban radius of a failed target recorded from $v_t$ is excluded,
\begin{equation}
\small
\begin{aligned}
Y_t &\leftarrow \bigl\{\, y_i \in Y_t \;:\; \lVert y-y_i\rVert \geq \varepsilon_{\text{ban}},\;\; \forall\,(v_t,y)\in B \,\bigr\},
\end{aligned}
\label{eq:fve_supp}
\end{equation}
\normalsize
with $\varepsilon_{\text{ban}}=0.5$\,m. An excluded proposal is discarded and creates no node. Because $B$ is shared by the team, the exclusion is collective: a frontier that one agent proved unreachable from $u$ is not re-proposed at $u$ by any teammate that later localizes to the same node.

\begin{table}[t]
\renewcommand{\arraystretch}{1.15}
\caption{States of the low-level controller. Transitions follow the listed conditions.}
\label{tab:fsm}
\centering
\resizebox{\linewidth}{!}{
\begin{tabular}{lll}
\toprule
\textbf{State} & \textbf{Action} & \textbf{Next state} \\
\midrule
\textsc{Select} & Pop the next route node of the macro-step; rotate toward it; set $\lfloor d/0.25\rfloor$ forward steps & \textsc{Track}, or \textsc{Done} if no node remains \\
\textsc{Track} & Execute one forward step of $0.25$\,m per synchronized step & \textsc{Select} when steps are exhausted; \textsc{RecoverInit} upon collision \\
\textsc{RecoverInit} & Refund the collided step; turn $90^{\circ}$ to a random side; load the six probe headings & \textsc{Sweep} \\
\textsc{Sweep} & Rotate to the next probe heading and attempt one forward step & \textsc{Probe}, or \textsc{Select} if no heading remains \\
\textsc{Probe} & If displaced, rotate back to the blocked heading & \textsc{Retry} if displaced, otherwise \textsc{Sweep} \\
\textsc{Retry} & Execute the remaining forward steps & \textsc{Select} when exhausted or upon collision \\
\textsc{Done} & Idle until all agents finish the macro-step & --- \\
\bottomrule
\end{tabular}}
\end{table}

\noindent\textbf{Low-level controller.} \label{app:controller} Routes are executed in synchronized macro-steps: in each macro-step, every agent that still has a route node tracks exactly that node, agents without one idle, and the next macro-step starts only after all agents have finished, which preserves the shared timeline. To track a node, the controller of \Cref{tab:fsm} rotates the agent toward it in $15^{\circ}$ increments and issues $\lfloor d/0.25\rfloor$ forward commands of $0.25$\,m, where $d$ is the horizontal distance to the node. If a forward command collides, the consumed step is refunded and a recovery sweep starts: the agent turns $90^{\circ}$ to a uniformly chosen side and probes the headings $90^{\circ},60^{\circ},30^{\circ},-30^{\circ},-60^{\circ},-90^{\circ}$ relative to the blocked heading, sweeping from the chosen side to the opposite one, by attempting one forward step at each. At the first heading that yields a displacement, it turns back to the blocked heading and executes the remaining forward steps; a further collision abandons the node. If no probe yields a displacement, the node is abandoned as well. In both cases the controller proceeds to the next route node without reporting the abandoned one. The controller reports as verified anchor the last explored route node that it completed without abandonment, and the departure node if no route node was completed. These anchors are the only nodes to which a newly created node is connected (\Cref{alg:topology}), which realizes traversal-verified connection: an edge is added only for a traversal that the controller completed, so detours and recoveries cannot introduce spurious shortcuts. Conversely, a dispatch that ends at an already known position is caught by failure-verified exclusion, so control failures are converted into persistent, team-visible map knowledge.

\subsection{Student-Forced Multi-Agent Learning}
\label{app:training}
\noindent\textbf{Rollout and supervision.} During imitation learning, each simulator instance executes a single-agent episode (\Cref{app:il}), and instances in the same scene share one topological memory, in which the agents are indexed by their order within the batch. The loss of the main paper is normalized by the number $n_{\mathrm{act}}$ of executed decisions in the iteration. Actions are drawn from the categorical distribution of the fused logits; with probability $\beta$ the expert action is executed instead, decaying with the training iteration. Our student-forced recipe uses $\beta_0=0$, and the teacher-forcing ablation uses $\beta_0=0.5$. Stop and backtracking moves are executed by teleportation to explored nodes, as in ETPNav, whereas moves toward unexplored nodes are executed physically from the front node with the recovery sweep of \Cref{app:controller}. The localization phase, failure-verified exclusion, candidate restriction and the consumption of selected unexplored nodes are identical to inference.

\noindent\textbf{Memory persistence.} A per-scene counter controls how long a memory survives: a new memory is created for a scene whenever the counter reaches the number $K_{\text{traj}}$ of accumulation rounds, and is otherwise reused after clearing the visitation records and detaching the stored node features from the computation graph. Our final recipe uses $K_{\text{traj}}=1$, i.e., every iteration starts from an empty memory that is then built jointly by the concurrent agents, while the accumulation ablation uses $K_{\text{traj}}=2$.

\begin{algorithm}[t]
\SetEndCharOfAlgoLine{}
\SetKwComment{Comment}{// }{}
\SetKw{KwAnd}{and}
\SetKwInOut{Input}{Input}
\SetKwInOut{Output}{Output}
\Input{Episode list $\mathcal{E}=(e_1,\dots,e_{|\mathcal{E}|})$ with scene, start node and reference path; team size $N$}
\BlankLine
Shuffle $\mathcal{E}$; $i\leftarrow1$ \\
\While{$i\le|\mathcal{E}|$}{
  $\mathcal{R}\leftarrow\{j>i : \text{scene}(e_j)=\text{scene}(e_i)\}$ \Comment{same scene}
  $\mathcal{C}\leftarrow\{j\in\mathcal{R} : \text{start}(e_j)\notin\{\text{start}(e_k)\}_{k=\max(i-N,1)}^{i}\}$ \Comment{distinct start}
  $\mathcal{F}\leftarrow\{j\in\mathcal{C} : \text{path}(e_j)\cap\text{path}(e_i)\neq\emptyset\}$ \Comment{overlapping path}
  \uIf{$\mathcal{F}\neq\emptyset$}{swap $e_{i+1}$ with $e_j$, $j\sim\mathcal{U}(\mathcal{F})$}
  \uElseIf{$\mathcal{C}\neq\emptyset$}{swap $e_{i+1}$ with $e_j$, $j\sim\mathcal{U}(\mathcal{C})$}
  \uElseIf{$\mathcal{R}\neq\emptyset$}{
    \uIf{fewer than $100$ relocation attempts were made at position $i$}{
      $\mathcal{C}'\leftarrow$ positions $j<i$ in the contiguous same-scene block preceding $e_i$ whose episodes at positions $j-N,\dots,j+N-1$ do not share the start node of $e_i$ \\
      $\mathcal{F}'\leftarrow\{j\in\mathcal{C}' : \text{path}(e_j)\cap\text{path}(e_i)\neq\emptyset \text{ or } \text{path}(e_{j-1})\cap\text{path}(e_i)\neq\emptyset\}$ \\
      \uIf{$\mathcal{F}'\cup\mathcal{C}'\neq\emptyset$}{move $e_i$ to position $j\sim\mathcal{U}(\mathcal{F}')$, or $j\sim\mathcal{U}(\mathcal{C}')$ if $\mathcal{F}'=\emptyset$; $i\leftarrow i-1$}
      \lElse{swap $e_{i+1}$ with $e_j$, $j\sim\mathcal{U}(\mathcal{R})$}
    }
    \lElse{swap $e_{i+1}$ with $e_j$, $j\sim\mathcal{U}(\mathcal{R})$}
  }
  $i\leftarrow i+1$ \\
}
Split $\mathcal{E}$ into streams $\mathcal{S}_1,\dots,\mathcal{S}_N$ with $e_i\in\mathcal{S}_{(i \bmod N)+1}$ \\
\Output{Streams $\{\mathcal{S}_n\}$, consumed one episode per stream per iteration with one memory per scene}
\caption{Team-Correlated Episode Sorting}
\label{alg:episched}
\end{algorithm}

\noindent\textbf{Team-correlated episode sorting.} Sharing memory during training is useful only if concurrently simulated agents operate in the same scene, start from distinct poses, and traverse overlapping space. \Cref{alg:episched} sorts the training stream accordingly. Starting from a shuffled list, the algorithm fills the position following each episode $e_i$ with a same-scene episode whose start node differs from the start nodes of the episodes at positions $i-N,\dots,i$, preferring candidates whose reference path shares a node with that of $e_i$. If every same-scene candidate conflicts on the start node, the algorithm attempts, up to $100$ times per position, to move $e_i$ backward into the contiguous same-scene block, to a position whose neighborhood of $N$ episodes on either side contains no start conflict, preferring positions adjacent to an episode with an overlapping path, and then revisits the vacated position; otherwise it accepts a same-scene episode despite the conflict. The sorted list is split into $N$ streams in round-robin order, and each training iteration advances every stream by one episode, so that the $N$ concurrently simulated agents are consecutive episodes of the sorted list. The list is reshuffled and re-sorted whenever a stream is exhausted.

\section{Verified Task Crafting}
\label{app:crafting}
This section unfolds the four-stage crafting pipeline, including the zone-repair and schedule-refinement algorithms, the well-formedness rules, and the design rules governing every synthesis prompt. We leverage different VLMs and LLMs of the Gemini-3 family across the stages, matching model capacity and reasoning effort to the difficulty of each step (\Cref{tab:models}): Gemini-3.1-flash-lite handles viewpoint-level annotation, profile selection, and perception-verified refinement of atomic instructions; Gemini-3-flash handles zone-level annotation, mission instantiation, atomic instruction generation, and arrival-sequence reconstruction; and Gemini-3.1-pro handles mission synthesis and instruction rendering.

\begin{table}[t]
\renewcommand{\arraystretch}{1.15}
\caption{Model assignment of the crafting pipeline.}
\label{tab:models}
\centering
\resizebox{0.5\linewidth}{!}{
\begin{tabular}{llll}
\toprule
\textbf{Stage} & \textbf{Step} & \textbf{Model} & \textbf{Reasoning} \\
\midrule
\multirow{2}{*}{1} & Viewpoint annotation & Gemini-3.1-flash-lite & high \\
 & Zone annotation & Gemini-3-flash & medium \\
\midrule
\multirow{2}{*}{2} & Profile selection & Gemini-3.1-flash-lite & default \\
 & Mission synthesis & Gemini-3.1-pro & low \\
\midrule
\multirow{2}{*}{3} & Mission instantiation & Gemini-3-flash & low \\
 & Atomic instruction generation & Gemini-3-flash & low  \\
\midrule
\multirow{3}{*}{4} & Perception-verified refinement & Gemini-3.1-flash-lite & high \\
 & Instruction rendering & Gemini-3.1-pro & low \\
 & Arrival-sequence reconstruction & Gemini-3-flash & medium \\
\bottomrule
\end{tabular}}
\end{table}

\subsection{Stage 1: Grounded Scene Understanding}
\label{app:stage1}
\noindent\textbf{Scene parsing and rendering.} Viewpoints are taken from the navigation graphs of HM3D scenes. For each scene, we parse the semantic annotations into object instances, discarding instances whose category is empty or contains any of the non-target terms wall, frame, floor, sheet, void, stairs, unknown, ceiling, window, curtain, beam and decoration. At every viewpoint, we render twelve RGB and pixel-aligned semantic views at $30^{\circ}$ heading increments with a $90^{\circ}$ vertical field of view at $512\times512$ resolution, and certify which instances are observable in each view by the protocol below. In the same pass, we compute the navigable anchors of every certified instance together with their geodesic distances from the viewpoint.

\noindent\textbf{Observability certification.} An instance--view pair $(o,i)$ at viewpoint $u$ is certified observable if and only if the following tests hold. (i)~\emph{Proximity:} the distance from $u$ to the oriented bounding box of $o$ is below $\theta_{\text{obs}}=1.5$\,m, which admits objects in the immediate surroundings of the viewpoint that a subtask can meaningfully target. (ii)~\emph{Semantic visibility:} the semantic identifier of $o$ occurs in the pixel-aligned semantic rendering of view $i$, which rejects nearby objects hidden behind walls or outside the view. (iii)~\emph{Mask sufficiency:} the bounding box of the visible mask of $o$ in view $i$ covers at least a fraction $\theta_{\text{area}}=10^{-4}$ of the frame, which rejects empty or degenerate masks at frame borders. (iv)~\emph{Reachability:} the anchor set is non-empty, which rejects objects none of whose sides can be reached from the viewpoint. The categories of instances passing tests (i) and (ii) form the instance lists given to the viewpoint annotator, while the grounded instances used for synthesis must pass all four tests, be confirmed by the annotator, and have anchors within $\theta_{\text{gnd}}=3$\,m geodesic distance of the viewpoint. The anchors of a released target are those computed at a certifying viewpoint, and they are stored with the episode as the instance locations consumed by \Cref{eq:reached}. Hence every grounded instance in MAVLN is visible from the viewpoint that grounds it, describable by the annotator, and reachable by a navigating agent, and generation, training supervision and evaluation share one consistent notion of reaching a target.

\noindent\textbf{Viewpoint annotation.} A VLM annotates each viewpoint from six directional views at $60^{\circ}$ spacing (forward, front-right, back-right, backward, back-left and front-left), each paired with an instance list of the categories of instances that pass the proximity and semantic-visibility tests in that view. For each view, the VLM describes the layout and spatial relationships and lists all observable objects with a description and a function. Objects from the potential list must be named exactly, with a numeric suffix, and only if actually visible, since the list is treated as noisy candidates rather than ground truth; all other objects also receive numeric suffixes. Suffixes must reflect physical identity across views: the same physical object seen in two directions receives one suffix, whereas distinct look-alikes, objects co-occurring in a view, and objects in different views without evidence of identity receive distinct suffixes. The VLM then writes a summary of the viewpoint centered on the agent's position, emphasizing navigation-relevant structure and estimated distances, and a list of consolidated objects that merges all per-view entries by category without suffixes. A response is accepted only if every consolidated category appears among the per-view object names.

\noindent\textbf{Grounded instances.} An instance is grounded at a viewpoint in a given direction if it passes all certification tests, its category matches the name of an object listed by the VLM in that direction, its category appears among the consolidated objects, and its anchors are within $\theta_{\text{gnd}}=3$\,m geodesic distance of the viewpoint. The grounded instances of a viewpoint are the union over its six directions, and their category descriptions are taken from the consolidated list. This rule makes every textual reference traceable to pixels: a subtask can only name a category that the VLM verifiably described in a specific view of a specific viewpoint and that corresponds to an observable, reachable instance there.

\begin{algorithm}[t]
\SetEndCharOfAlgoLine{}
\SetKwComment{Comment}{// }{}
\SetKw{KwAnd}{and}
\SetKwInOut{Input}{Input}
\SetKwInOut{Output}{Output}
\Input{Initial zones $\{Z_j\}$ from region annotations; navigation graph $G_{\text{nav}}$; round cap $20$}
\BlankLine
\lFor{every empty zone}{move the least-connected viewpoint of the largest zone into it}
\For{round $=1$ \KwTo $20$}{
  \For{every zone $Z_j$ with more than one connected component on $G_{\text{nav}}$}{
    $C_0\leftarrow$ largest component of $Z_j$ \\
    \For{every other component $C$ of $Z_j$}{
      \Comment{Phase 1: bridge}
      $\mathcal{B}\leftarrow\{u\in Z_{j'},\,j'\neq j : u \text{ adjacent to } C_0 \text{ and to } C,\; |Z_{j'}|>1,\; Z_{j'}\setminus\{u\} \text{ connected}\}$ \\
      \lIf{$\mathcal{B}\neq\emptyset$}{move some $u\in\mathcal{B}$ into $Z_j$; $C_0\leftarrow C_0\cup\{u\}\cup C$; \textbf{continue}}
      \Comment{Phase 2: safe relocation}
      \For{$u\in C$}{
        $j^{*}\leftarrow\argmax_{j'\neq j}\,|\{w\in Z_{j'} : (u,w)\in G_{\text{nav}}\}|$ subject to $Z_{j'}\cup\{u\}$ connected and a positive score \\
        \lIf{$j^{*}$ exists}{move $u$ to $Z_{j^{*}}$}
      }
      \Comment{Phase 3: forced relocation}
      \For{$u\in C$ still in $Z_j$}{
        $j^{*}\leftarrow\argmax_{j'\neq j}\,|\{w\in Z_{j'} : (u,w)\in G_{\text{nav}}\}|$ \\
        \lIf{the score of $j^{*}$ is positive}{move $u$ to $Z_{j^{*}}$}
        \lElse{leave $u$ in $Z_j$ and log a warning}
      }
    }
  }
  \lIf{no viewpoint moved in this round}{\textbf{break}}
}
\Output{Repaired zones $\{Z_j\}$ and the list of unresolved zones}
\caption{Zone-Connectivity Repair}
\label{alg:zone}
\end{algorithm}

\noindent\textbf{Zone construction and repair.} Zones are initialized from the region annotations of HM3D. Region names are obtained from per-scene weighted votes over region labels, and regions labeled as unknown rooms are discarded. Each viewpoint is assigned to the region whose floor object has the smallest oriented-bounding-box distance to it, and empty regions are dropped. These raw assignments are unreliable planning units, because region boundaries frequently cut through navigable space. We therefore repair them by \Cref{alg:zone}, which iterates for at most $20$ rounds. In each round, every disconnected zone keeps its largest connected component on the navigation graph and processes each remaining fragment in three phases. The bridge phase borrows a single viewpoint from another zone that is adjacent to both the main component and the fragment, provided that its removal does not disconnect its source zone and does not empty it. If no bridge exists, the relocation phase moves each fragment viewpoint to the adjacent zone with the largest number of navigation-graph edges to it, provided that the receiving zone remains connected. Remaining viewpoints are force-relocated to the best-connected adjacent zone regardless of the receiving zone's connectivity, and viewpoints without any adjacent zone are left in place with a warning. The iteration stops when a round changes nothing. For mission synthesis, each zone is finally restricted to its largest connected component on the navigation graph, and two zones are adjacent if any navigation-graph edge joins them.

\noindent\textbf{Zone annotation.} An LLM aggregates the viewpoints of each zone, i.e., their summaries, grounded-instance descriptions and positions, into a zone description. The prompt supplies the voted region name as a reference, requires the zone to be characterized by its primary single room type while discounting objects of adjacent rooms visible near boundaries, and requires a description of every grounded instance of the zone under its exact name, merging observations from different viewpoints. A response is accepted only if the described instances coincide exactly with the zone's grounded instances. This stage yields $20{,}910$ annotated viewpoints aggregated into $1{,}628$ zone descriptions.

\subsection{Stage 2: Diverse Mission Synthesis}
\label{app:stage2}
\noindent\textbf{Household profiles.} Mission narratives are conditioned on one of $20$ structured household profiles, such as a family with young children, a multi-generational home, professional roommates, a household with disability accommodations, home-based healthcare, a household with rotating shift workers, or a household with special dietary needs. Each profile specifies a household type, its members with role, age, occupation, mobility, schedule and task preferences, and, where relevant, medical, accessibility, dietary or special needs, together with the household's coordination style, common activities and typical challenges. For each scene, a profile-selection pass receives the scene description and all profiles, and returns the indices of the profiles whose needs are best accommodated by the layout, considering zone functions and contained instances, zone connectivity, mobility requirements, schedule compatibility, special needs, age-appropriate spaces and supported activities, and prioritizing profiles in which several members' needs are met simultaneously. Profiles are the mechanism through which constraints arise organically: restricted mobility motivates fetch chains, overlapping schedules motivate parallelism, and continuous-care needs are the canonical source of presence locks.

\noindent\textbf{Mission budget and team size.} A budget of $14{,}000$ missions is distributed over scenes in proportion to the number of viewpoints in their zones that contain at least one grounded instance, and within each scene over zones in proportion to their viewpoints by largest-remainder rounding, zones whose share is below one mission receiving none. The admissible team size of a mission centered on a zone depends on the zone's number of viewpoints: at most two agents below $47$ viewpoints, at most three below $70$, and at most four otherwise. The team size of each mission is drawn uniformly between two and this maximum, and the selected profiles are cycled over the missions of a zone in a shuffled order.

\noindent\textbf{Mission drafting.} Conditioned on a profile, the center zone (its number of viewpoints, summary and grounded-instance descriptions), the adjacent zones with grounded instances, the textual zone connectivity and the team size, the LLM drafts a task demand and a list of subtasks, each with an identifier, a drafted agent, a zone, a navigation instruction, an exact target name, the dependency list, the lock and holding flags, the release list, and a natural-language description justifying every non-default attribute. A draft is accepted only if it contains a task demand, every target is a grounded instance of its zone, every zone is the center zone or one of its adjacent zones, at least one target lies in the center zone, subtask identifiers are unique, and every dependency refers to a previously listed subtask.

\begin{table}[t]
\renewcommand{\arraystretch}{1.15}
\caption{Well-formedness rules enforced before and after instantiation.}
\label{tab:wellformed}
\centering
\resizebox{0.95\linewidth}{!}{
\begin{tabular}{ll}
\toprule
\textbf{Rule} & \textbf{Condition for rejection} \\
\midrule
Unreleased lock & A locked subtask has a later subtask of the same agent depending on it, and no subtask releases it \\
Release by another agent & A released subtask is not a transitive dependency of the releasing subtask, or is found among its dependencies with the same drafted agent \\
Single consumer & A holding subtask has more than one dependent subtask of the same agent \\
Single held object & The consumer of a holding subtask also depends on another holding subtask \\
Distinct consecutive goals & A subtask shares its goal viewpoint with a same-agent dependency (after instantiation) \\
\bottomrule
\end{tabular}}
\end{table}

\noindent\textbf{Well-formedness.} Before instantiation, every release list is closed under dependencies, i.e., each subtask in $U_k$ is added to $D_k$, and the draft must pass the static checker of \Cref{tab:wellformed}. The checker operates on the DAG in which an edge points from each dependency to its dependent subtask, and it considers the drafted agents.

\subsection{Stage 3: Mission Refinement}
\label{app:stage3}
\noindent\textbf{Instance grounding.} For every subtask, the candidate goal viewpoints are the viewpoints of its zone at which its target is grounded. For every drafted agent, candidate viewpoint sequences over its subtasks are composed such that consecutive goals differ and their navigation-graph distance lies in $(3,20)$\,m; a draft is discarded if any agent has no valid sequence. The grounding mode depends on the size of the joint combination space. If the product of the per-agent sequence counts is below $100$, every joint combination is instantiated, scheduled and ranked by \Cref{alg:refine}, rewritten by its best schedule, validated against the leg bounds below, and retained only if distinct start viewpoints can be allocated; the LLM then selects one of the retained complete combinations, preferring the one whose viewpoints best support the subtasks' purposes and spatial context. Otherwise, agents with fewer than $100$ sequences are offered their complete sequences, while the subtasks of the remaining agents are offered their candidate viewpoints individually; the LLM must then select exactly one sequence for each constrained agent, assign one candidate viewpoint to each remaining subtask without reusing a viewpoint for consecutive subtasks of the same agent, and prioritize coverage and reasonable distances over contextual fit when concurrent subtasks have few options. In both modes, the LLM also rewrites every subtask instruction with visual details of the assigned viewpoint and a reference to the room, and its output is accepted only if subtask identifiers and drafted agents are preserved, each viewpoint belongs to the subtask's zone and grounds its target, and the instantiated draft passes \Cref{tab:wellformed}.

\noindent\textbf{Atomic instruction generation.} For every subtask, an LLM receives the grounded instruction and the directional views of the goal viewpoint in which the target is grounded, each with its layout description and target annotations, selects the view whose annotation is most distinctive and spatially precise, and rewrites the instruction into a destination-only atomic instruction. The candidate instances of the subtask are the grounded instances of the target category in the selected direction.

\begin{algorithm}[t]
\SetEndCharOfAlgoLine{}
\SetKwComment{Comment}{// }{}
\SetKw{KwAnd}{and}
\SetKwInOut{Input}{Input}
\SetKwInOut{Output}{Output}
\SetKwFunction{FTrav}{Traverse}
\SetKwProg{Fn}{Function}{:}{}
\Input{\\\hspace{-3.6em}\small
\begin{tabular}[t]{l @{\hspace{.3em}} l}%
$\mathcal{T}$ & Instantiated subtask DAG \\
$\mathcal{A}$ & Drafted agents \\
$s_{1:M}$ & Initial viewpoints \\
\end{tabular}\hspace{-0.5em}%
\begin{tabular}[t]{l @{\hspace{.3em}} l}%
$\mathrm{SP}(\cdot,\cdot)$ & Graph shortest paths \\
$C_{\max}$ & Enumeration cap \\
\end{tabular}%
}
\BlankLine
$\mathcal{S}\leftarrow\emptyset$ \\
\Fn{\FTrav{$\mathit{state}$}}{
  $\mathcal{Z}\leftarrow$ admissible single commitments $(a,\tau)$ at $\mathit{state}$ \\
  \For{$(a,\tau)\in\mathcal{Z}$}{
    $\mathit{state}'\leftarrow$ copy of $\mathit{state}$; complete $\tau$ with executor $a$; release every $u\in U_{\tau}$ held by a teammate \\
    append $\mathrm{SP}(\mathrm{pos}(a),\mathrm{vp}(\tau))$ to the itinerary of $a$ and empty legs to all other agents; $\mathrm{pos}(a)\leftarrow\mathrm{vp}(\tau)$ \\
    \lIf{$|\mathcal{S}|\le C_{\max}$}{\FTrav{$\mathit{state}'$}}
  }
  \lIf{no recursion took place \KwAnd every subtask is complete in $\mathit{state}$}{$\mathcal{S}\leftarrow\mathcal{S}\cup\{\mathit{state}\}$}
}
\FTrav{initial state at $s_{1:M}$} \\
\BlankLine
\For{candidate $\mathcal{P}\in\mathcal{S}$ with commitments $c=1,\dots,N$}{
  $m\leftarrow0$; all cursors at $c=1$ \\
  \For{$c=1$ \KwTo $N$}{
    advance lagging cursors to $c$; refresh the lock flag of the agent committing at $c$ \\
    \While{the leg of commitment $c$ is not exhausted}{
      \For{agent $a\in\mathcal{A}$}{
        \lIf{$a$ has an exhausted current leg \KwAnd $a$ is unlocked}{move the cursor of $a$ to its next non-empty leg}
        \lIf{the current leg of $a$ is non-empty}{advance $a$ by one node}
      }
      $m\leftarrow m+1$ \\
    }
  }
  $\mathrm{key}(\mathcal{P})\leftarrow\bigl(m,\;\#\mathrm{deviations}(\mathcal{P}),\;-\sum\mathrm{depth}(\mathrm{deviations}(\mathcal{P}))\bigr)$ \\
}
\Output{$\mathcal{P}^{\star}\leftarrow\argmin_{\mathcal{P}\in\mathcal{S}}\mathrm{key}(\mathcal{P})$; allocation rewritten by $\mathcal{P}^{\star}$}
\caption{Makespan-Optimal Schedule Refinement}
\label{alg:refine}
\end{algorithm}

\noindent\textbf{Makespan-optimal schedule refinement.} The instantiated mission is then scheduled by \Cref{alg:refine}. A schedule is a sequence of single agent--subtask commitments. At every state, the admissible commitments are determined by the task semantics: a locked agent receives nothing until a teammate completes a subtask that releases it; a holding agent may only commit to the consumer of its held subtask with the same drafted executor, and this parent is reserved for it; and every other agent may commit to any ready subtask, i.e., an incomplete subtask whose dependencies are complete, regardless of the drafted allocation. Each commitment completes the subtask, records its executor, releases the locks listed in its release list, and appends to the executor's itinerary the navigation-graph shortest path from its current viewpoint to the goal. Every complete commitment sequence is then replayed on a synchronized clock. At commitment $c$, the clock advances one node per step until the committing agent completes its leg, while every other agent whose current leg is exhausted and which is not locked pulls its next leg forward and advances in lockstep; lock flags are refreshed from the state recorded at each commitment, which reflects the releases performed so far. The global makespan is the number of steps of the replay. Candidates are ranked lexicographically by makespan, by the number of subtasks whose executor deviates from the drafted agent, and by the negated sum of the DAG depths of the deviating subtasks, and the top-ranked schedule is adopted: when it reassigns a subtask, the episode is rewritten accordingly, so that the released allocation is makespan-justified rather than LLM-asserted.

\noindent\textbf{Continuous-space validation and start allocation.} After rewriting, every consecutive leg of every agent must have a navigation-graph distance in $[3,20]$\,m and a geodesic distance of at most $20$\,m in the continuous simulator, since graph distances systematically underestimate continuous detours around furniture and through doorways. Missions failing these checks are discarded rather than repaired. Start viewpoints are then allocated by enumerating the combinations of candidate instances over all subtasks in random order. For each combination, a viewpoint is eligible as a starting location for an agent if it is not the first goal viewpoint of any agent, its navigation distance to the agent’s first goal viewpoint is between 4 and 15 m, the shortest path contains at least four viewpoints, and its distance to the oriented bounding box of the agent’s first target instance is between 3 and 15 m. Distinct starting viewpoints are then selected through a minimum-cost assignment with independent, uniformly sampled random costs for eligible agent–viewpoint pairs, randomizing the selection while ensuring that no two agents share the same start. The combination is rejected if some agent has no eligible start. Each agent's trajectory is then rendered along the graph shortest paths through its goals: at every node, the agent faces the next node, the view index is the corresponding $30^{\circ}$ heading bin, and at the goal the view of the selected direction is appended. A leg is rejected if its starting position lies within $3$\,m geodesic distance of any anchor of its target, so that every subtask requires navigation to be credited. The first combination passing all checks is released, together with the start positions and the start orientations facing the next node on each agent's first path.

\subsection{Stage 4: Verified Instruction Rendering}
\label{app:stage4}
\noindent\textbf{Perception-verified refinement.} For every subtask, the candidate instances at its goal viewpoint are painted with distinct colors on darkened copies of the six even-indexed views, and only the views in which some candidate is visible are kept; these composites serve for auditing. A VLM receives the trajectory frames labeled by navigation step, the clean goal views in which a candidate is visible, and a map listing each candidate's name, color and bounding boxes (restricted to boxes covering at least a fraction $\theta_{\text{area}}$ of the frame) in each goal view. It must first select the candidate that best matches the instruction by reasoning about its position, size, appearance and surroundings, and then rewrite the target description from what is actually visible for the selected instance, overriding any wrong or hallucinated attributes, specifying which part of the object the agent arrives at, and grounding it by its immediate neighbors at the arrival position. The output is accepted only if the selected name is one of the candidates, which fixes the target instance $g_k$ of the subtask; subtasks whose candidates have no valid bounding box, or whose refinement fails, invalidate the mission. This step converts generation-time language into perception-verified language: the released atomic instruction describes the instance as a navigating agent will observe it.

\noindent\textbf{Variant rendering.} Each mission is rendered by one call producing the decentralized and centralized instructions and by a separate call producing the centralized-implicit instruction. Both calls receive the task demand, the agent identifiers, the per-subtask information, viewpoint and zone descriptions, pairwise navigation-graph distances between goal viewpoints in steps, and the zone affiliation of each goal. The decentralized and centralized call additionally receives the drafted and final executor of each subtask and the subtask descriptions, and must organize the instructions according to the final executors whenever schedule refinement reassigned a subtask. The centralized-implicit call receives no executor information, so its single instruction treats all subtasks as one coordinated plan without attribution. In the released episodes, a decentralized subtask permits only its final executor, whereas centralized and centralized-implicit subtasks permit every agent of the team.

\noindent\textbf{Closed-loop filtering.} Every rendered instruction set faces a round-trip filter. An LLM receives only the instruction(s), the agent identifiers and the subtasks presented as an unordered list with their permitted agents, atomic instructions and goal viewpoints, and must return an interleaved sequence of $(\text{agent},\text{viewpoint})$ arrivals in which each agent's destinations follow the order implied by its instruction and the interleaving respects the stated waiting and coordination conditions. The sequence is replayed by a generation-time version of the evaluator in which an arrival consumes the first ready subtask of the agent whose goal viewpoint matches, lock and holding violations fail the involved subtasks as in \Cref{alg:evaluator}, and the mission is valid if and only if every subtask succeeds. Instructions failing the filter are regenerated and re-filtered, and a mission is released only if its instructions are valid in all three regimes, so that the three regimes share the same missions. The filter certifies the property that evaluation relies upon: the language alone, under the task's discrete-event semantics, determines a successful execution.

\noindent\textbf{Scene-disjoint splits.} \label{app:splits} Scenes are grouped by the largest team size among their valid missions, and scenes hosting at least one four-agent mission are sorted by their number of missions. All scenes without four-agent missions and six of the four-agent scenes form the training split of $128$ scenes, while seven four-agent scenes form val-unseen and ten form test-unseen. Interleaving the ranks balances scenes with many and few missions across splits, and every unseen scene hosts four-agent missions. Finally, we apply human assessment to the val-unseen and test-unseen split to remove erroneous missions and therefore ensure the dataset quality.

\begin{table}[t]
\renewcommand{\arraystretch}{1.12}
\caption{Consolidated hyperparameters.}
\label{tab:hyper}
\centering
\resizebox{\linewidth}{!}{
\begin{tabular}{lll}
\toprule
\textbf{Symbol} & \textbf{Meaning} & \textbf{Value} \\
\midrule
\multicolumn{3}{l}{\emph{Generation}} \\
$\theta_{\text{obs}}$ & Oriented-bounding-box proximity for observability & $1.5$\,m \\
$\theta_{\text{area}}$ & Minimum mask bounding-box area ratio & $10^{-4}$ \\
--- & Anchor deduplication radius & $0.5$\,m \\
$\theta_{\text{gnd}}$ & Maximum geodesic distance from a viewpoint to the anchors of a grounded instance & $3$\,m \\
--- & Mission synthesis budget; zone-size thresholds for team sizes $\le2/\le3/\le4$ & $14{,}000$; $<47$ / $<70$ / $\ge70$ viewpoints \\
--- & Consecutive-goal band on the navigation graph; continuous-space leg bound & $(3,20)$\,m; $\le20$\,m \\
--- & Start-to-first-goal band; minimum start path length & $(4,15)$\,m; $4$ nodes \\
--- & Start-to-first-target bounding-box band; minimum start-to-anchor geodesic & $(3,15)$\,m; $3$\,m \\
$C_{\max}$ & Schedule enumeration cap & $10^{5}$ \\
--- & Joint combinations below which grounding is exhaustive & $10^{2}$ \\
--- & Zone repair rounds & $20$ \\
\midrule
\multicolumn{3}{l}{\emph{Evaluation}} \\
--- & Success radius to navigable anchors & $3$\,m \\
$\rho$ & Agent radius for MAC & $0.25$\,m \\
\midrule
\multicolumn{3}{l}{\emph{Scheduling}} \\
--- & Rounds after which an all-blank round always terminates & $20$ \\
\midrule
\multicolumn{3}{l}{\emph{Shared topology and planning}} \\
$K$ & Maximum waypoint proposals per step & $5$ \\
$\varepsilon_{\text{loc}}$ & Node identification radius & $0.5$\,m \\
$\varepsilon_{\text{nbr}}$ & Round-start stitching radius & $1.5$\,m \\
$\varepsilon_{\text{ban}}$ & Ban radius & $0.5$\,m \\
$T_{\max}$ & Planning step limit per round & $30$ \\
\midrule
\multicolumn{3}{l}{\emph{Conflict-aware execution}} \\
$\delta$ & Collision distance in path finding & $0.4$\,m \\
$T_h$ & Space-time horizon & at least $5$ \\
$B_{\text{cbs}}$ & Expansion budget & $10^{3}$ \\
--- & Segment interpolation points & $11$ \\
--- & Forward quantum; turn quantum; recovery probes & $0.25$\,m; $15^{\circ}$; $\pm30^{\circ},\pm60^{\circ},\pm90^{\circ}$ \\
\midrule
\multicolumn{3}{l}{\emph{Learning}} \\
--- & Pretraining steps (selected checkpoint) & $200{,}000$ ($180{,}000$) \\
--- & Optimizer; learning rate & AdamW; $10^{-5}$ \\
--- & Imitation-learning iterations; checkpoint interval & $30{,}000$; $1{,}000$ \\
$M$ & Concurrent agents sharing memory during training & $4$ \\
$\beta_0$ & Initial teacher-forcing ratio (decay interval) & $0$ ($5{,}000$ iterations) \\
$K_{\text{traj}}$ & Memory accumulation rounds & $1$ \\
--- & Expert stop radius & $1.5$\,m \\
--- & Maximum instruction length & $512$ tokens \\
\bottomrule
\end{tabular}}
\end{table}

\newcommand{\promptfig}[1]{\includegraphics[width=0.485\linewidth,height=0.40\textheight,keepaspectratio]{appendix_fig/#1}}

\clearpage

\begin{figure}[p]
\centering
\subfloat[Viewpoint annotation.]{\promptfig{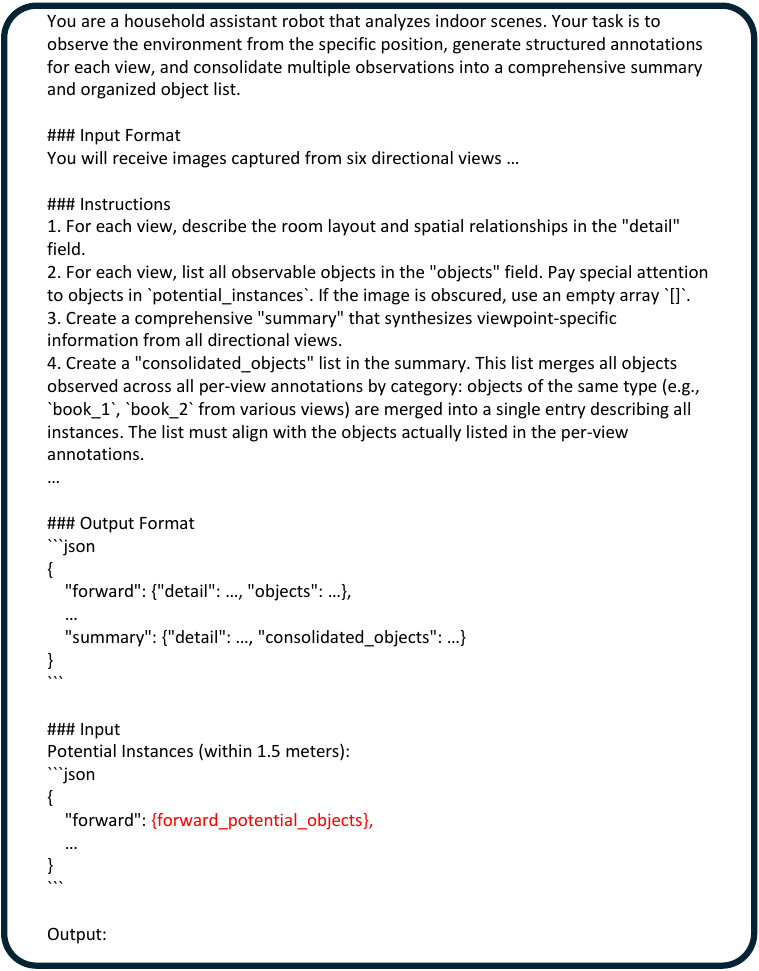}}\hfil
\subfloat[Zone annotation.]{\promptfig{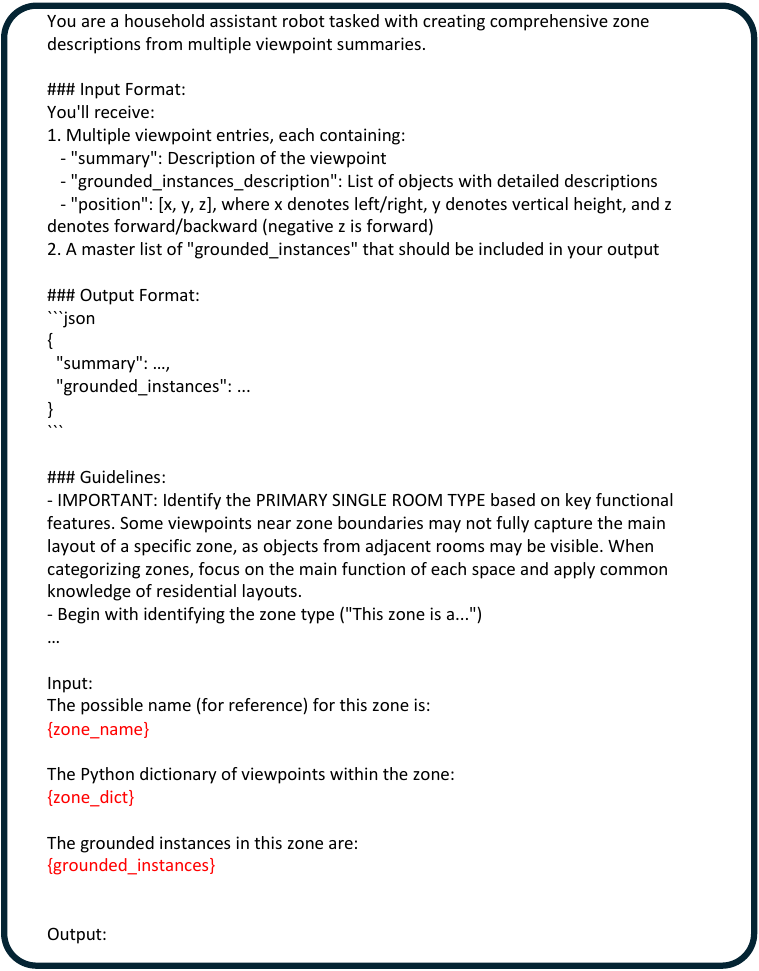}}
\\[2ex]
\subfloat[Household-profile selection.]{\promptfig{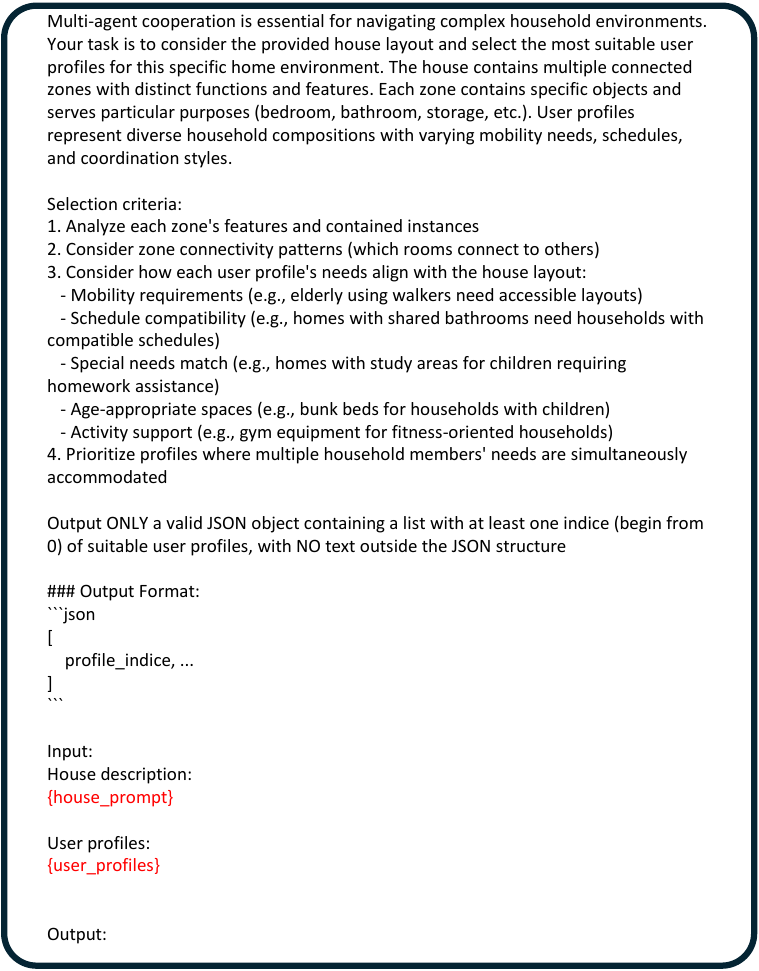}}\hfil
\subfloat[Mission drafting.]{\promptfig{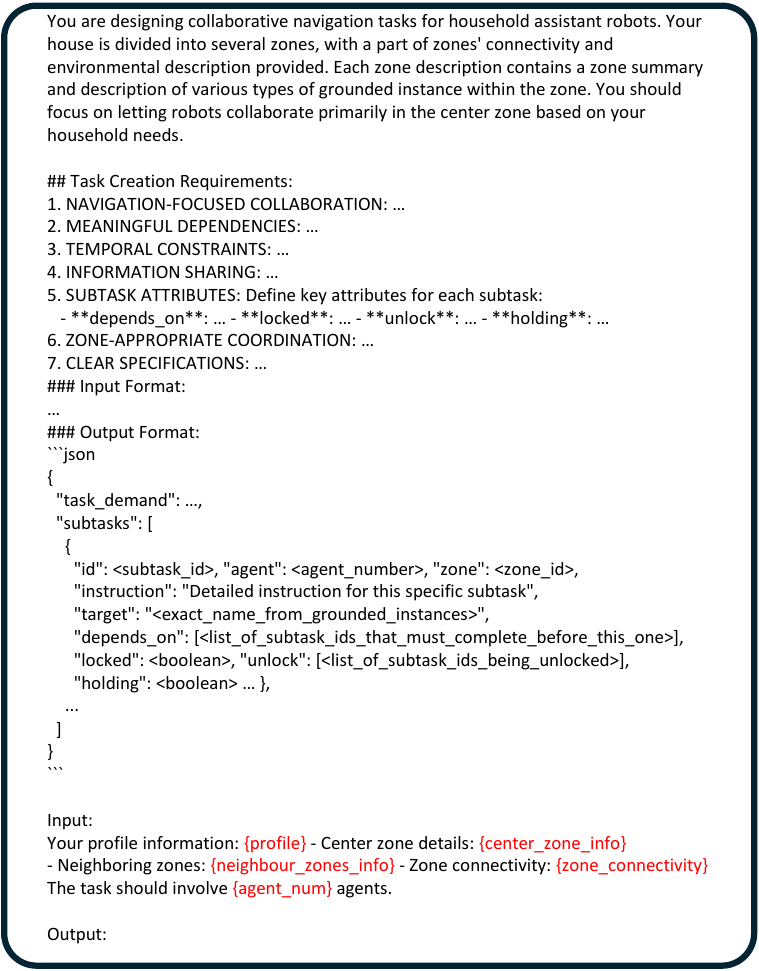}}
\caption{Prompt templates of grounded scene understanding (\Cref{app:stage1}) and diverse mission synthesis (\Cref{app:stage2}). \textbf{(a)} Per-viewpoint annotation from six directional views, with the potential-instance list of every view. \textbf{(b)} Aggregation of the annotated viewpoints of a zone into a zone description. \textbf{(c)} Selection of the household profiles that a scene can accommodate. \textbf{(d)} Drafting of the task demand and of the attributed subtask graph for one center zone, one profile and one team size.}
\label{fig:prompt_stage12}
\end{figure}

\begin{figure}[p]
\centering
\subfloat[Instance grounding, sequence mode.]{\promptfig{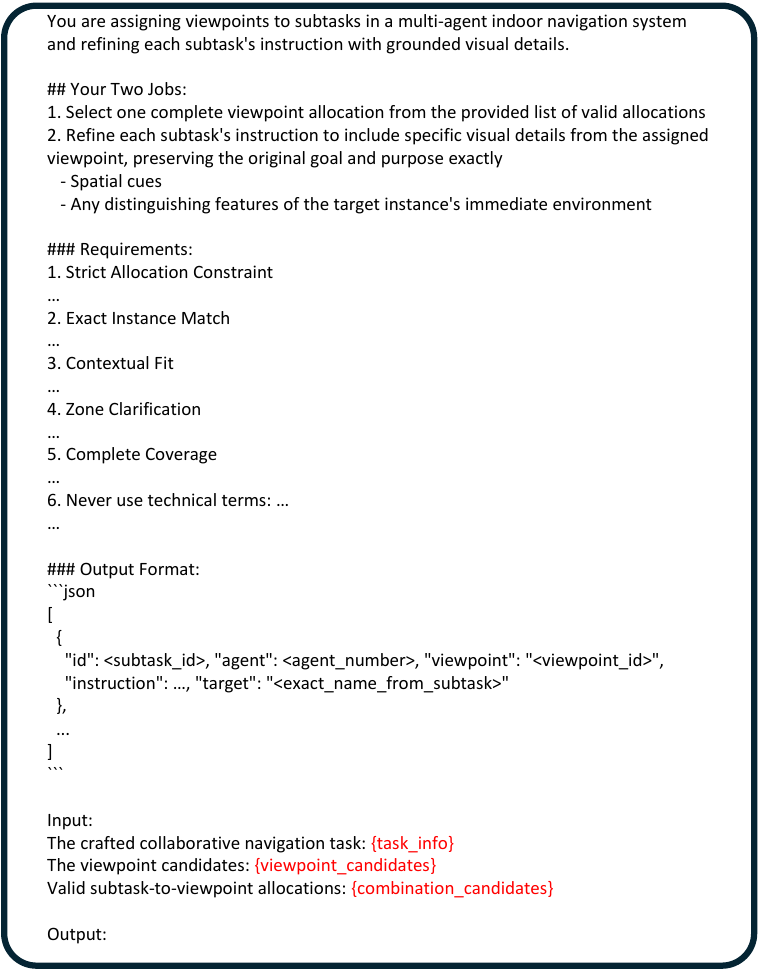}}\hfil
\subfloat[Instance grounding, viewpoint mode.]{\promptfig{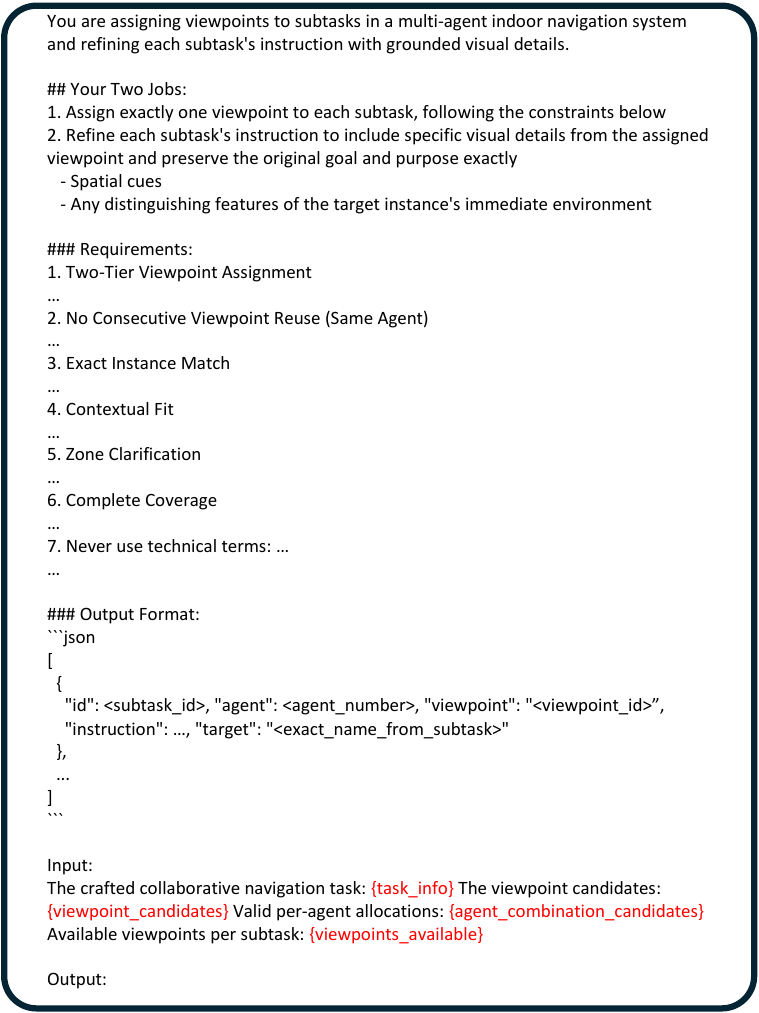}}
\\[2ex]
\subfloat[Atomic instruction generation.]{\promptfig{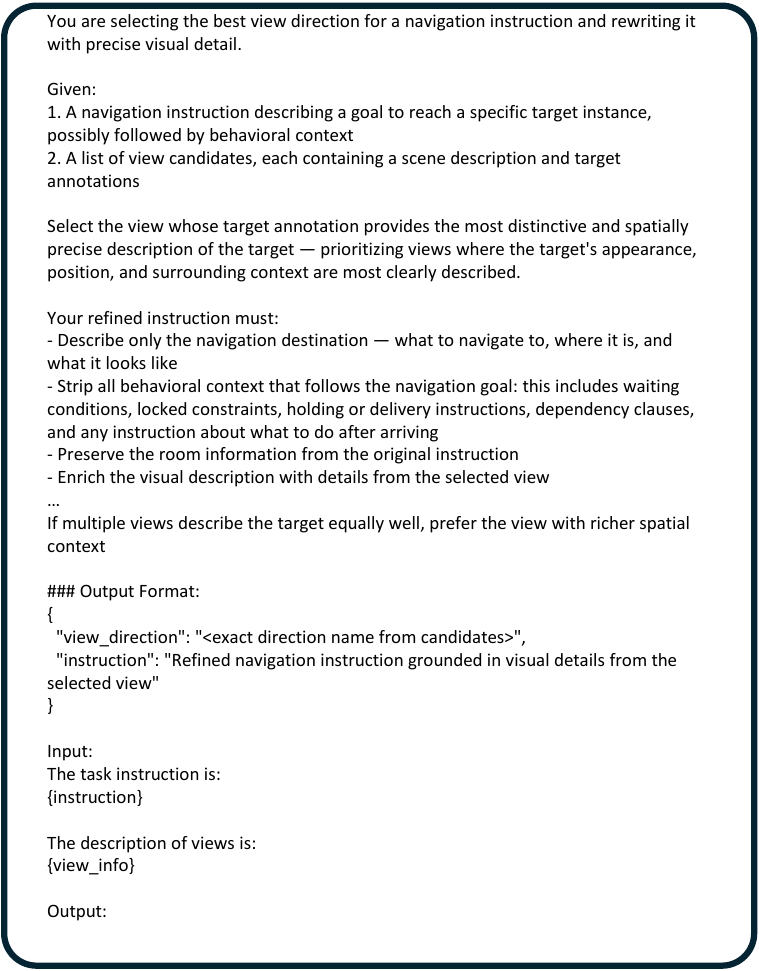}}
\caption{Prompt templates of mission refinement (\Cref{app:stage3}). \textbf{(a)} Instantiation when the joint combination space is enumerable, so that every agent is offered complete viewpoint sequences of already validated joint combinations. \textbf{(b)} Instantiation when it is not, so that the subtasks of the unconstrained agents are offered their candidate viewpoints individually. \textbf{(c)} Rewriting of a grounded instruction into a destination-only atomic instruction from the directional views of its goal viewpoint.}
\label{fig:prompt_stage3}
\end{figure}

\begin{figure}[p]
\centering
\subfloat[Perception-verified refinement.]{\promptfig{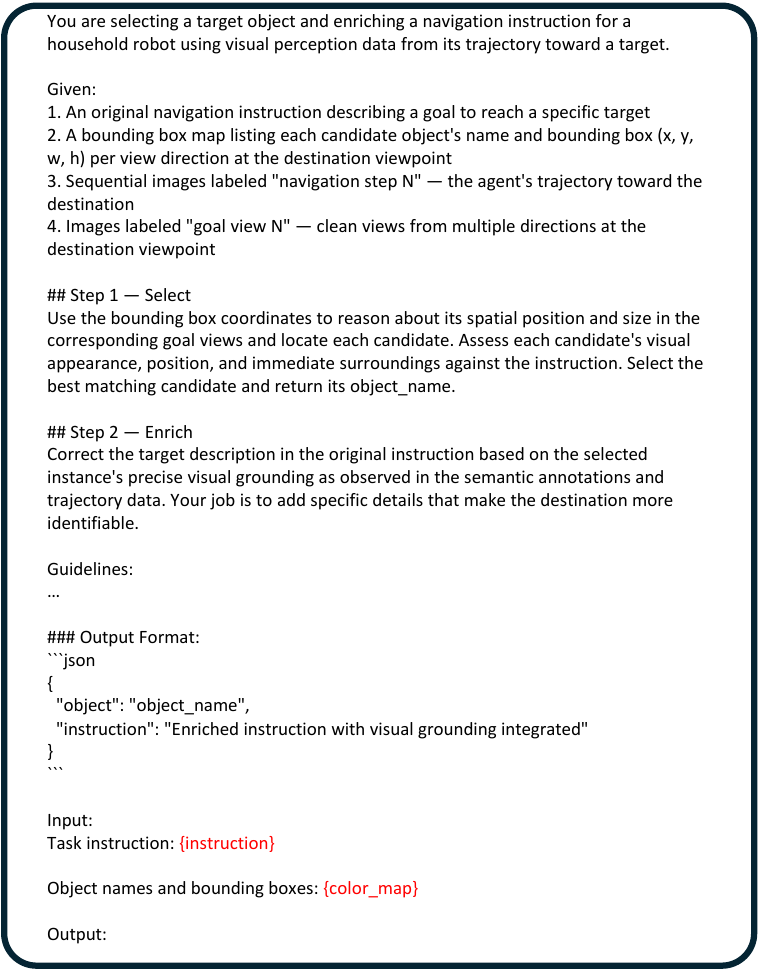}}\hfil
\subfloat[Decentralized and centralized rendering.]{\promptfig{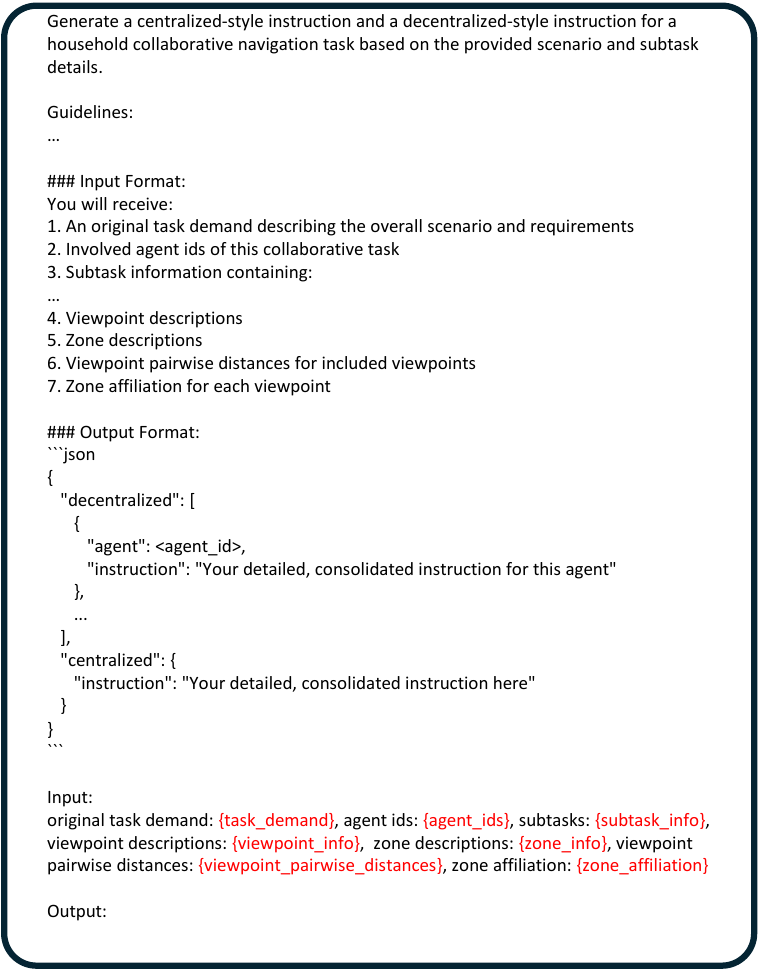}}
\\[2ex]
\subfloat[Centralized-implicit rendering.]{\promptfig{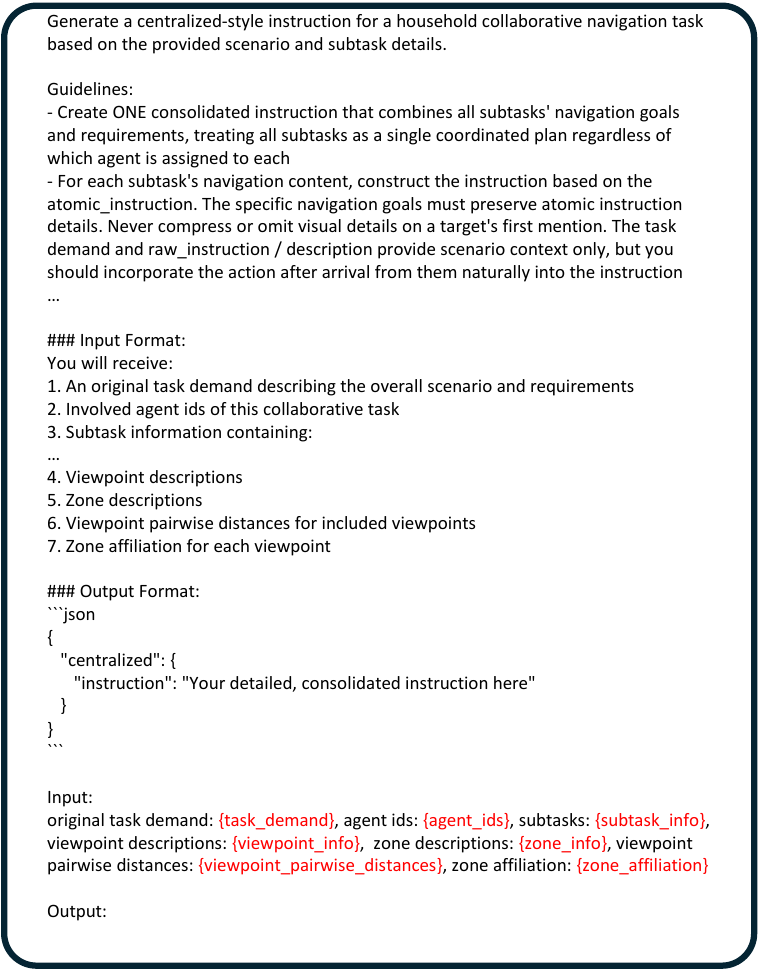}}\hfil
\subfloat[Arrival-sequence reconstruction.]{\promptfig{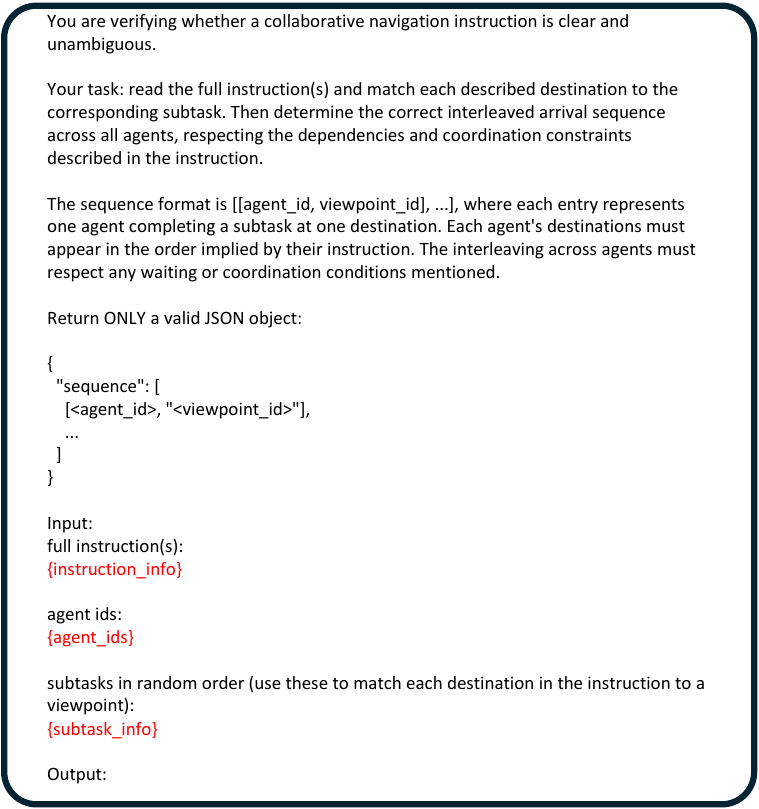}}
\caption{Prompt templates of verified instruction rendering (\Cref{app:stage4}). \textbf{(a)} Selection of the target instance among the colored candidates and rewriting of its description from what is visible at the arrival position. \textbf{(b)} Rendering of the decentralized and centralized instructions, organized by the final executors. \textbf{(c)} Rendering of the centralized-implicit instruction, which receives no executor information. \textbf{(d)} Reconstruction of the interleaved arrival sequence used by the closed-loop validity filter.}
\label{fig:prompt_stage4}
\end{figure}

\begin{figure}[p]
\centering
\subfloat[Atomic instruction extraction.]{\promptfig{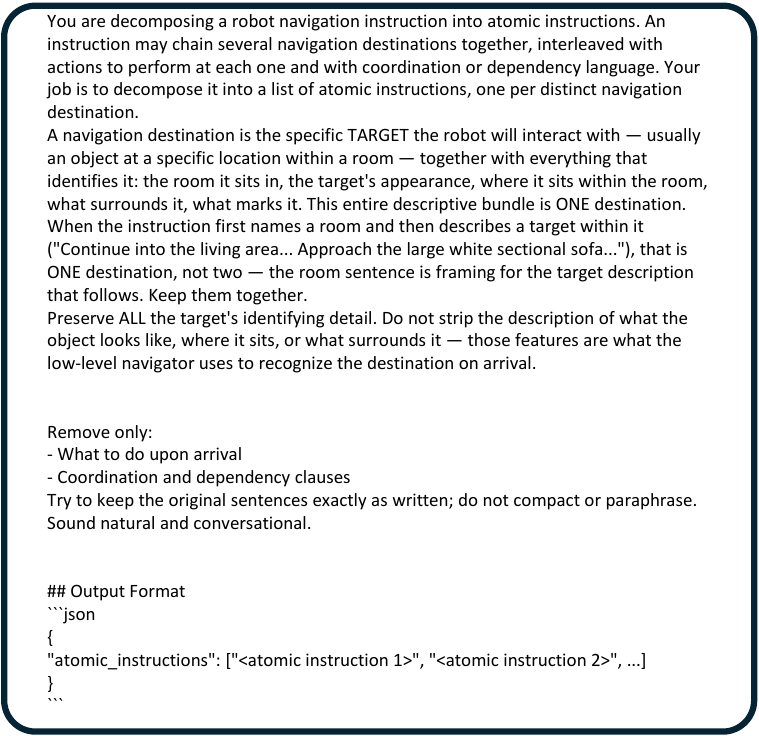}}\hfil
\subfloat[Round-wise scheduling.]{\promptfig{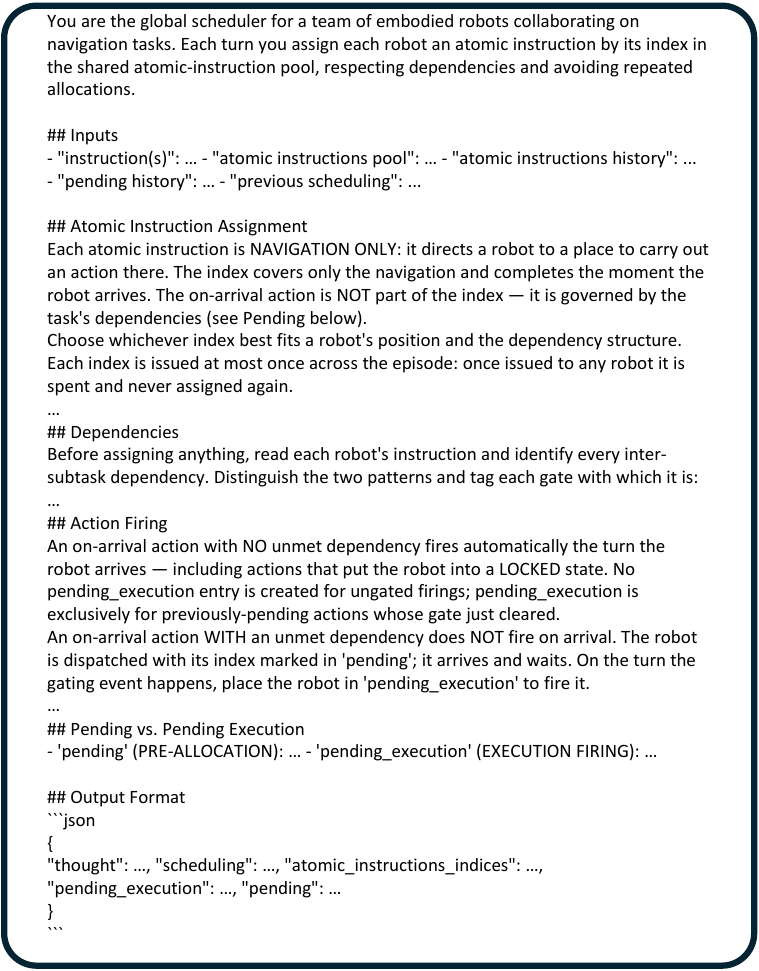}}
\\[2ex]
\subfloat[Observation-guided scheduling.]{\promptfig{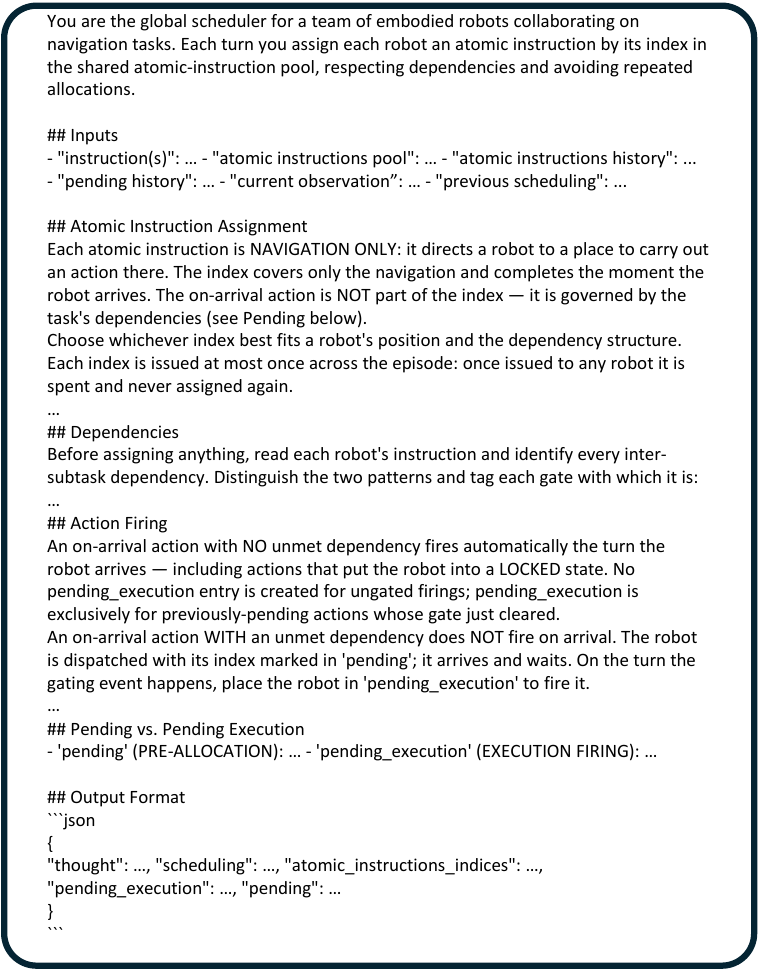}}\hfil
\subfloat[Viewpoint description.]{\promptfig{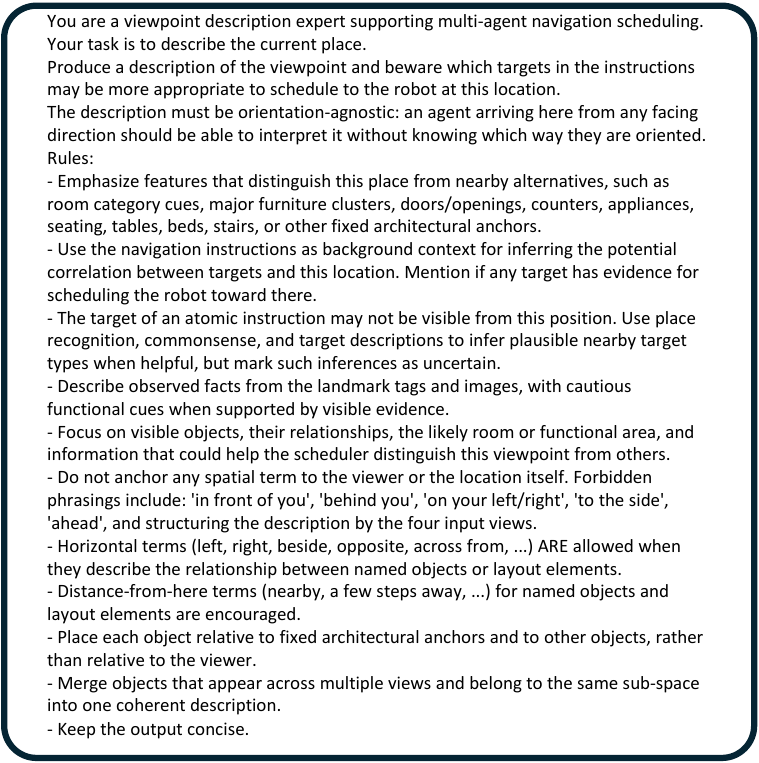}}
\caption{Prompt templates of the subtask scheduler of TRISS (\Cref{app:scheduler}). \textbf{(a)} Decomposition of the task instructions into the shared pool of atomic instructions. \textbf{(b)} Emission of the per-agent assignment, the firing list and the pre-allocated agents of one round. \textbf{(c)} The same protocol extended by the situation report of \Cref{eq:situation}. \textbf{(d)} Orientation-agnostic description of an agent's current viewpoint, which forms that report.}
\label{fig:prompt_sched}
\end{figure}

\begin{figure}[p]
\centering
\subfloat[Zero-shot VLM planner.]{\promptfig{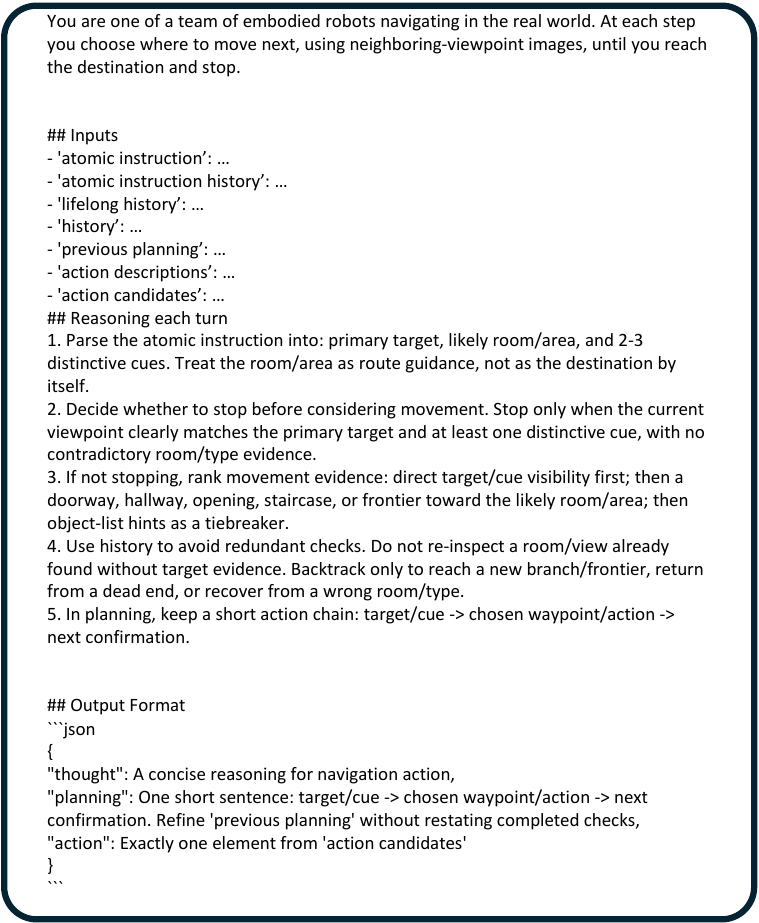}}\hfil
\subfloat[Hybrid planner.]{\promptfig{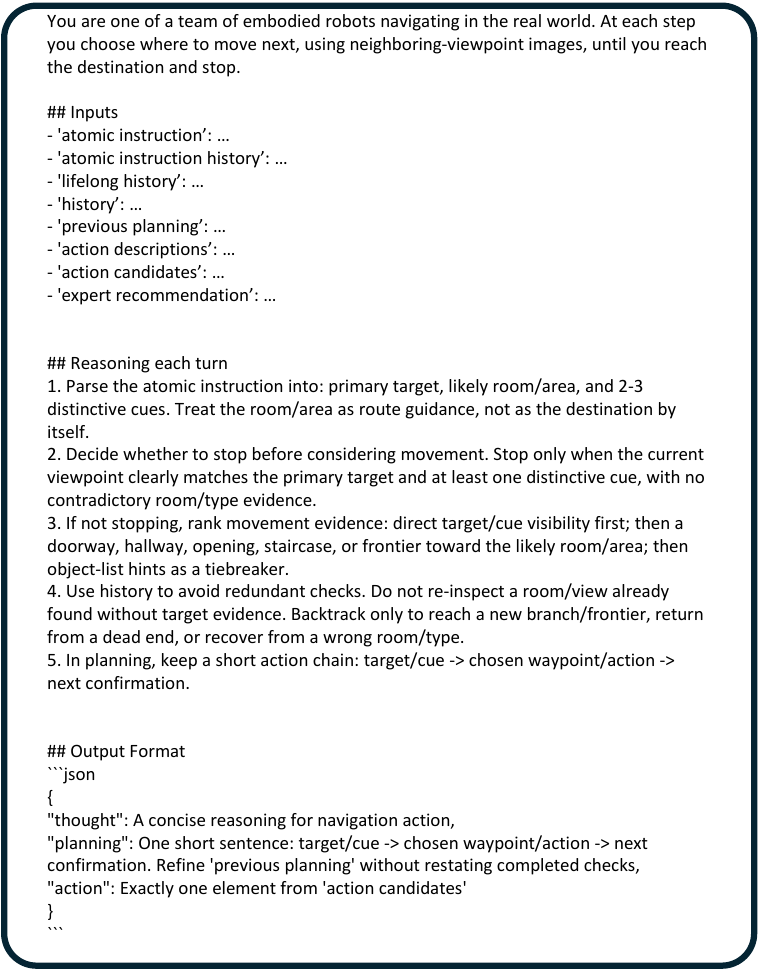}}
\caption{Prompt templates of the VLM-assisted planners (\Cref{app:hybrid}). \textbf{(a)} Zero-shot planning from the candidate views, the verbalized action options and the agent's own step history. \textbf{(b)} Hybrid planning, queried only when the trained planner is uncertain, which additionally receives its ranked shortlist as a strong prior and the targets already claimed by teammates.}
\label{fig:prompt_nav}
\end{figure}